\documentclass[runningheads]{llncs}
\usepackage[utf8]{inputenc}
\usepackage{xcolor}
\usepackage[textwidth=3cm]{todonotes} % disable
\usepackage{amssymb}
\usepackage[english]{babel}
\usepackage{graphicx}
\usepackage{wrapfig}
\def\R{\mathbb{R}}
\def\N{\mathbb{N}}
\usepackage{url}
\usepackage{subcaption}
\usepackage{bbold}
\def\one{\mathbb{1}}
\def\zero{\mathbb{0}}
\usepackage{neuralnetwork}

\usepackage{subcaption}
\usepackage{hyperref}

\def\ux{\underline{x}}
\def\ox{\overline{x}}

\begin{document}

\title{Static analysis of ReLU neural networks with tropical polyhedra} 

%[Goubault, Palumby, Sankaranarayanan and Putot]
\author{Eric Goubault\thanks{This work was partially supported by the academic Chair "Engineering of Complex Systems", Thal\`es-Dassault Aviation-Naval Group-DGA-Ecole Polytechnique-ENSTA Paris-T\'el\'ecom Paris, and AID project "Drone validation and swarms of drones"}\inst{1} \and S\'ebastien Palumby\inst{1} \and Sriram Sankaranarayanan\inst{2} \and Sylvie Putot\inst{1} \and Louis Rustenholz\inst{1}}
\institute{LIX, Ecole Polytechnique, CNRS and Institut Polytechnique de Paris, 91128 Palaiseau, France
        \email{\{name.surname\}@polytechnique.edu} \and          Engineering Center Computer Science, University of Colorado at Boulder, USA \email{srirams@colorado.edu}}

%\author{Eric Goubault}
%\affiliation{%
% \institution{LIX, Ecole Polytechnique, CNRS and Institut Polytechnique de Paris}
% \city{Palaiseau}
% \country{France}}
%\email{Eric.Goubault@polytechnique.edu}

%\author{Sebastien Palumby}
%\affiliation{%
% \institution{LIX, Ecole Polytechnique, CNRS and Institut Polytechnique de Paris}
% \city{Palaiseau}
% \country{France}}
% \email{Sebastien.Palumby@polytechnique.edu}

% \author{Sylvie Putot}
%\affiliation{%
% \institution{LIX, Ecole Polytechnique, CNRS and Institut Polytechnique de Paris}
% \city{Palaiseau}
% \country{France}}
% \email{Sylvie.Putot@polytechnique.edu}

% \author{Louis Rustenholz}
%\affiliation{%
% \institution{LIX, Ecole Polytechnique, CNRS and Institut Polytechnique de Paris}
% \city{Palaiseau}
% \country{France}}
% \email{Louis.Rustenholz@polytechnique.edu}
 
% \author{Sriram Sankaranarayanan}
%\affiliation{%
% \institution{Engineering Center Computer Science, University of Colorado at Boulder}
% \city{Palaiseau}
% \country{USA}}
 %\email{srirams@colorado.edu}
 
\authorrunning{E. Goubault et al.}
%\date{September 2020}
%\author{Double-blind review process}
%\institute{}

\bibliographystyle{plain}

\maketitle
\begin{abstract}
%Neural net verification is central to the adoption of AI techniques as key technologies for critical systems, such as perception for autonomous driving and control systems in the large. As a first step, 
%%We are tackling in this work the fundamental problem of range analysis for neural networks, which is the basis for considering more intricate properties, e.g. robustness, compliance to specifications, reachability when used in control systems.
This paper studies the problem of range analysis for feedforward neural networks, which is a basic primitive for applications such as robustness of neural networks, compliance to specifications and reachability analysis of neural-network feedback systems. Our approach focuses on ReLU (rectified linear unit) feedforward neural nets that present specific difficulties: approaches that exploit derivatives do not apply in general, the number of patterns of  neuron activations can be quite large even for small networks, and convex approximations are generally too coarse. In this paper, we employ set-based methods and abstract interpretation that have been very successful in coping with similar difficulties in classical program verification. %We concentrate here on the precision loss when using convex domains, against which disjunctive analyzes were proposed in the context of program verification.
We present an approach that abstracts ReLU feedforward neural networks using tropical polyhedra. We show that tropical polyhedra can efficiently abstract ReLU activation function, while being able to control the loss of precision due to linear computations. We show how the connection between ReLU networks and tropical rational functions can provide approaches for range analysis of ReLU neural networks. % with linear tropical functions. 
%\todo{Phrase precedente pas mal mais se la pete un peu trop?}
We report on a preliminary evaluation of our approach using a prototype implementation.
%...\todo{EG: will anonymize later}
\end{abstract}

%\keywords{}

\section{Introduction and related work}

Neural networks are  now widely used in numerous applications including speech recognition, natural language processing, image segmentation, control and planning for autonomous systems. A central question is how to verify that they are correct with respect to some specification. Beyond correctness, we are 
also interested in questions such as explainability and fairness, that can in turn be specified as formal verification problems.
%In many cases, e.g. image classification, it is already difficult to make sense of the problematic at hand, as there is no way to formally specify what we should be computing.
%The interest in explainability and validation for neural networks based systems has fostered an enormous amount of research in recent years, revisiting some older themes in formal methods. 
Recently, the problem of verifying properties of neural networks has been investigated extensively under a variety of contexts.
A natural neural network analysis problem is that of \emph{range estimation}, i.e. bounding the values of neurons on the output layer, or some function of the output neurons,
given the range of neurons on the input layer. A prototypical application of range estimation is the verification of  the ACAS Xu - the next generation collision avoidance system for autonomous aircrafts, which is implemented by a set of neural networks~\cite{Julian+Kochendorfer/2018/Deep}. Such a verification problem 
is translated into a range estimation problem over these
neural network wherein the input ranges concern a set of possible scenarios and the outputs indicate the possible set of advisories provided by the network~\cite{reluplex}. 

Another prototypical application concerns the robustness of image classification wherein we wish to analyze whether a classification label remains constant for images in a neighborhood of a given image that is often specified using ranges over a set of pixels.
Robustness is akin to numerical stability analysis, and for neural nets used as decision procedures (e.g. control of a physical apparatus), this is a form of decision consistency. 
It is also linked to the existence or non-existence of adversarial inputs, i.e. those inputs close to a well classified input data, that dramatically change the classification \cite{advers}, and may have dire consequences in the real world \cite{adversphysical}.
% e.g. Reluplex Katz, Barrett, Dill, Julian, Kochenderfer, Reluplex: An Efficient SMT Solver for Verifying Deep Neural Networks, CAV 2017, also DeepSafe etc.
%
%Finally, even though the specification of a neural net, in isolation, is unknown, it may be the case that a partial specification is avalaible, when put in context. This is most notably the case when neural nets are used as controllers, within a feedback loop, for which it is of great interest to prove stability or reachability properties of the closed loop system. 
%; Sherlock [MILP] Dutta, Chen, Jha, Sankaranarayanan, Tiwari, Sherlock - A tool for verification of neural network feedback systems, HSCC 2019, Verisig [Hybrid System] Ivanov, Weimer, Alur, Pappas, Lee, Verisig: Verifying Safety Properties of Hybrid Systems with Neural Network Controllers HSCC 2019, ReachNN Huang, Fan, Li, Chen, Zhu, ReachNN: Reachability Analysis of Neural-Network Controlled Systems, 2019 etc.

Many formal methods approaches that have been successfully used in the context of program verification seem to be successfully leveraged to the case of neural net verification: proof-theoretic approaches, SMT techniques, constraint based analyzers and abstract interpretation. %Multiple Layer Perceptrons (MLP) or feedforward neural networks can be seen as loop-free programs, with linear assignments and applications of activation functions.  It is therefore tempting to assess how to translate some of the successes of program verification to neural net verification. 
In this paper, we are interested in developing abstract interpretation \cite{AI} techniques for feedforward networks with ReLU activation functions. ReLU feedforward networks can be seen as loop-free programs with affine assignments and conditionals with affine guards, deciding whether the corresponding neuron is activated or not. For researchers in program analysis by abstract interpretation, this is a well known situation. The solutions range from designing a scalable but imprecise analyses by %abstracting the collecting semantics, generally propagating 
convexifications of the set of possible values of each neurons throughout all layers to designing a potentially exponentially complex analysis by performing a fully disjunctive analysis. In between, some heuristics have been successfully used in program analysis, that may alleviate the burden of disjunctive analysis, see e.g. \cite{partitioning}, \cite{Bourdoncle92abstractinterpretation}. 
Among classical convex abstractions, the zones \cite{Mine2}  are a nice and scalable abstraction, successfully used in fully-fledged abstract interpretation based static analyzers \cite{ASTREE}. 
In terms of disjunctive analysis,  a compact way to represent a large class of disjunctions of zones are the tropical polyhedra, used for disjunctive program analysis in e.g. \cite{SAS2008,PhDXavier}. 
Tropical polyhedra are, similarly to classical convex polyhedra, defined by sets of affine inequalities but where the sum is replaced by max operator and the multiplication is replaced by the addition. %This will be explained in full detail in Section \ref{sec:troppoly}. 

\begin{figure}
    \centering
    \begin{subfigure}[b]{0.2\textwidth}
    \centering
        % Sketch output, version 0.3 (build 7d, Thu Mar 18 10:53:46 2021)
% Output language: PGF/TikZ,LaTeX
\begin{tikzpicture}[line join=round,scale=0.7]
\draw[line width=2pt,draw=blue](-1,0)--(0,0)--(1,1);
\draw[arrows=->,line width=1pt,dotted](-2,0)--(2,0);
\draw[arrows=->,line width=1pt,dotted](0,-1)--(0,2);
\path (-1,0) node[below] {$-1$}
     (0,0) node[below] {$0$}
     (1,0) node[below] {$1$}
     (0,1) node[left] {$1$}
     (0,2) node[left] {$y$}
     (2,0) node[below] {$x$};\end{tikzpicture}% End sketch output
        \subcaption{Exact}
        \label{fig:RELU:exact}
        \end{subfigure}
        \begin{subfigure}[b]{0.3\textwidth}
        \centering
        % Sketch output, version 0.3 (build 7d, Thu Mar 18 10:53:46 2021)
% Output language: PGF/TikZ,LaTeX
\begin{tikzpicture}[line join=round,scale=0.7]
\filldraw[fill=lightgray](-1,0)--(1,0)--(1,1)--cycle;
\draw[line width=2pt,draw=blue](-1,0)--(0,0)--(1,1);
\draw[arrows=->,line width=1pt,dotted](-2,0)--(2,0);
\draw[arrows=->,line width=1pt,dotted](0,-1)--(0,2);
\path (-1,0) node[below] {$-1$}
     (0,0) node[below] {$0$}
     (1,0) node[below] {$1$}
     (0,1) node[left] {$1$}
     (0,2) node[right] {$y$}
     (2,0) node[below] {$x$};\end{tikzpicture}% End sketch output
        \subcaption{1-ReLU (DeepPoly)}
        \label{fig:RELU:poly}
        \end{subfigure}
        \begin{subfigure}[b]{0.2\textwidth}
        \centering
        % Sketch output, version 0.3 (build 7d, Thu Mar 18 10:53:46 2021)
% Output language: PGF/TikZ,LaTeX
\begin{tikzpicture}[line join=round,scale=0.7]
\filldraw[fill=lightgray](-1,0)--(0,0)--(1,1)--(0,1)--cycle;
\draw[line width=2pt,draw=blue](-1,0)--(0,0)--(1,1);
\draw[arrows=->,line width=1pt,dotted](-2,0)--(2,0);
\draw[arrows=->,line width=1pt,dotted](0,-1)--(0,2);
\path (-1,0) node[below] {$-1$}
     (0,0) node[below] {$0$}
     (1,0) node[below] {$1$}
     (0,1) node[left] {$1$}
     (0,2) node[left] {$y$}
     (2,0) node[below] {$x$};\end{tikzpicture}% End sketch output
        \subcaption{Zones}
        \label{fig:RELU:zone}
        \end{subfigure}
            \begin{subfigure}[b]{0.27\textwidth}
    \centering
        % Sketch output, version 0.3 (build 7d, Thu Mar 18 10:53:46 2021)
% Output language: PGF/TikZ,LaTeX
\begin{tikzpicture}[line join=round,scale=0.7]
\draw[line width=2pt,draw=blue](-1,0)--(0,0)--(1,1);
\draw[arrows=->,line width=1pt,dotted](-2,0)--(2,0);
\draw[arrows=->,line width=1pt,dotted](0,-1)--(0,2);
\path (-1,0) node[below] {$-1$}
     (0,0) node[below] {$0$}
     (1,0) node[below] {$1$}
     (0,1) node[left] {$1$}
     (0,2) node[left] {$y$}
     (2,0) node[below] {$x$};\end{tikzpicture}% End sketch output
        \subcaption{Tropical polyhedra}
        \label{fig:RELU:trop}
        \end{subfigure}
    \caption{Abstractions of the ReLU graph on $[-1,1]$}
    \label{fig:RELU}
\end{figure}
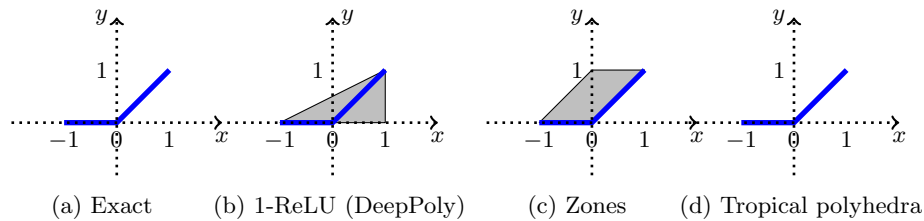

Zones are interesting for synthesizing properties such as robustness of neural networks used for classifying data. Indeed, classification relies on determining which output neuron has the greatest score, translating immediately into zone-like constraints. 
%All existing abstract interpretation based neural networks analysers are based on classical convex domains so far, such as zonotopes or polyhedra. 
%This should be of no surprise: zonotopes \cite{zonotopes} have been successfully used for proving numerical and control programs correct \cite{DeepZ}, and even for studying the robustness \cite{VMCAI2011} of numerical programs, all implemented in the FLUCTUAT static analyzer \cite{FLUCTUAT}.  
%\todo{Le "this should be of no surprise" et ce qui suit fait un peu bizarre, je crois que j'enleverais}
%
%\todo{Mettre tres informellement ce qu'est un polyedre tropical et exemple ReLU}
ReLU functions $x \mapsto max(0,x)$ are tropically linear, hence an abstraction using tropical polyhedra will be exact. A direct verification of classification specifications can be done from a tropical polyhedron by computing the enclosing zone, see \cite{SAS2008} and Section \ref{sec:background:zone}. In Figure \ref{fig:RELU}, we pictured the graph of the ReLU function $y=max(x,0)$ for $x \in [-1,1]$ (Figure \ref{fig:RELU:exact}), and its abstraction by 1-ReLU in DeepPoly \cite{DeepPoly} (Figure \ref{fig:RELU:poly}), by a zone (Figure \ref{fig:RELU:zone}), and by a tropical polyhedron (Figure \ref{fig:RELU:trop}), which is exactly the graph of the function. 
%Tropical polyhedra are convex in the max-plus algebra 
%but non convex in the classical sense, making it possible to represent in a much finer way all possible activation states.
Unfortunately, (classical) linear functions are tropically non-linear. But contrarily to program analysis where we generally discover the function to abstract inductively on the syntax, we are here given the  weights and biases for the full network, allowing us to design much better abstractions than if directly using the ones available from the program verification literature.%\footnote{In e.g. \cite{SAS2008} the emphasis is on tropical linear assignments or tropical linearization of the addition and scalar multiplication. Similarly for zones, e.g. \cite{Mine}, where the emphasis is on sums with intervals, and not general linear assignments.} 
%\todo{J'ai enlevé ta phrase au-dessus car pas clair ici a mon avis, mais idealement si on pouvait donner un exemple de ce que ca donnerait sur un petit exemple dnas la section 3?}
%(both for zones and tropical polyhedra), this is the center of this paper, Section \ref{sec:1layer}. 

It was recently proved \cite{RELU} that the class of functions computed by a feedforward neural network with ReLU activation functions is exactly the class of rational tropical maps, at least when dealing with rational weights and biases. It is thus natural to look for  guaranteed approximants of these rational tropical maps as abstractions. %, for instance by tropical linearization as in this work. 

\begin{example}[Running example]
\label{ex:running}
Consider a neural network with 2 inputs $x_1$ and $x_2$ given in [-1,1] and 2 outputs. The linear layer is defined by $h_1=x_1-x_2-1$, $h_2=x_1+x_2+1$ and followed by a ReLU layer with neurons $y_1$ and $y_2$ such that $y_1=max(0,x_1-x_2-1)$ and $y_2=max(0,x_1+x_2+1)$.
%depicted in Figure \ref{fig:simpleneuralnet} where we separated the first layer into a linear layer, with neuron $h_1$ and $h_2$ from top to bottom
%and where the biases are $b_1=-1$ and $b_2=1$, and a ReLU layer, with neurons $y_1$ and $y_2$. Thus $h_1=x_1-x_2-1$, $h_2=x_1+x_2+1$, $y_1=max(0,x_1-x_2-1)$ and $y_2=max(0,x_1+x_2+1)$. 
%We will see in Section \ref{sec:troppoly} that $y_1$ is a tropical rational function and $y_2$ is a tropical polynomial in $x_1$ and $x_2$. 

The exact range for nodes $(h_1,h_2)$ is depicted in Figure \ref{fig:simplenetinnerrange} in magenta (an octagon here), and the exact range for the output layer is shown in Figure \ref{fig:simplenetinnerrange2} in cyan: $(y_1,y_2)$ take the positive values of $(h_1,h_2)$. %\todo{le range de h1 et h2 est plutot l'union du magenta et du cyan?}. 
In Figure \ref{fig:simpleneuralnet:zone}, the set of values the linear node $h_1$ can take as a function of $x_1$, is represented in magenta. The set of values of the output neuron $y_1$ in function of $x_1$ is depicted in Figure \ref{fig:simpleneuralnet:zone2}, in cyan: when $x_1$ is negative, $h_1$ is negative as well, so $y_1=0$ (this is the horizontal cyan line on the left). When $x_1$ is positive, the set of values $y_1$ can take is the positive part of the set of values $h_1$ can take (pictured as the right cyan triangle). The line plus triangle is a tropical polyhedron, as we will see in Section \ref{sec:troppoly}. 

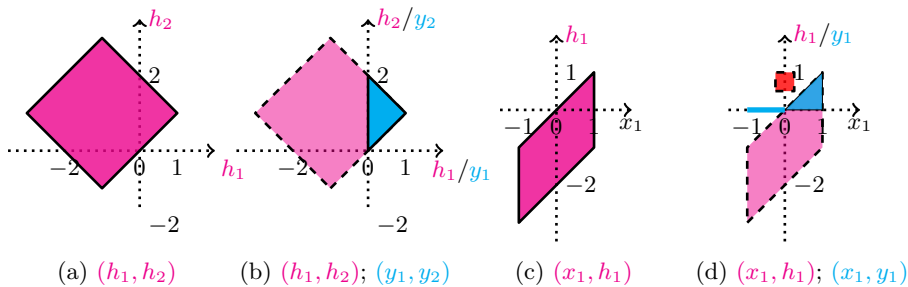
\begin{figure}
    \centering
%    \begin{subfigure}[b]{0.3\textwidth}
%    \centering
%    \begin{neuralnetwork}[height=4,layerspacing=15mm]
%\newcommand{\nodetextclear}[2]{$b_#2$}
%		\newcommand{\nodetextx}[2]{$x_#2$}
%		\newcommand{\nodetexty}[2]{$y_#2$}
%		\newcommand{\linklabelRELU}[4]{$\scriptstyle ReLU$}
%		\newcommand{\linklabelone}[4]{$\scriptstyle 1$}
%		\newcommand{\linklabelmone}[4]{$\scriptstyle -1$}
%		\inputlayer[count=2, bias=false, title=Input\\layer, text=\nodetextx]
%		\hiddenlayer[count=2, bias=false, title=Linear\\layer, text=\nodetextclear] 
%		\link[from layer=0, to layer=1, from node=1, to node=1, label=\linklabelone]
%		\link[from layer=0, to layer=1, from node=2, to node=2, label=\linklabelone]
%		\link[labelpos=near end,from layer=0, to layer=1, from node=2, to node=1, label=\linklabelmone]
%		\link[labelpos=near end,from layer=0, to layer=1, from node=1, to node=2, label=\linklabelone]
%		\outputlayer[count=2, title=Output\\ ReLU layer, text=\nodetexty] %\linklayers
%		\link[from layer=1, to layer=2, from node=1, to node=1, label=\linklabelRELU]
%		\link[from layer=1, to layer=2, from node=2, to node=2, label=\linklabelRELU]
%	\end{neuralnetwork}
%	            \subcaption{A simple neural network}
%        \label{fig:simpleneuralnet}
%        \end{subfigure}
        \begin{subfigure}[b]{0.24\textwidth}
        \centering
        % Sketch output, version 0.3 (build 7d, Thu Mar 18 10:53:46 2021)
% Output language: PGF/TikZ,LaTeX
\begin{tikzpicture}[line join=round,scale=0.5]
\draw[arrows=->,line width=1pt,dotted](-3.5,0)--(2,0);
\draw[arrows=->,line width=1pt,dotted](0,-1.5)--(0,3.5);
\filldraw[fill=magenta,line width=1pt,fill opacity=0.8](-1,-1)--(1,1)--(-1,3)--(-3,1)--cycle;
\path (-2,0) node[below] {$-2$}
     (0,0) node[below] {$0$}
     (1,0) node[below] {$1$}
     (0,2) node[right] {$2$}
     (0,3.5) node[right] {$\textcolor{magenta}{h_2}$}
     (2.5,0) node[below] {$\textcolor{magenta}{h_1}$}
     (0,-2) node[right] {$-2$};\end{tikzpicture}% End sketch output
        \subcaption{ $\color{magenta}(h_1,h_2)$}
        \label{fig:simplenetinnerrange}
        \end{subfigure}
        \begin{subfigure}[b]{0.24\textwidth}
        \centering
        % Sketch output, version 0.3 (build 7d, Thu Mar 18 10:53:46 2021)
% Output language: PGF/TikZ,LaTeX
\begin{tikzpicture}[line join=round,scale=0.5]
\draw[arrows=->,line width=1pt,dotted](-3.5,0)--(2,0);
\draw[arrows=->,line width=1pt,dotted](0,-1.5)--(0,3.5);
\filldraw[fill=magenta,line width=1pt,fill opacity=0.5,dashed](-1,-1)--(1,1)--(-1,3)--(-3,1)--cycle;
\filldraw[fill=cyan,line width=1pt,fill opacity=1](0,0)--(1,1)--(0,2)--cycle;
\path (-2,0) node[below] {$-2$}
     (0,0) node[below] {$0$}
     (1,0) node[below] {$1$}
     (0,2) node[right] {$2$}
     (0,3.5) node[right] {$\textcolor{magenta}{h_2}$/$\textcolor{cyan}{y_2}$}
     (2.5,0) node[below] {$\textcolor{magenta}{h_1}$/$\textcolor{cyan}{y_1}$}
     (0,-2) node[right] {$-2$};\end{tikzpicture}% End sketch output
        \subcaption{$\color{magenta}(h_1,h_2)$; $\color{cyan} (y_1,y_2)$}
        \label{fig:simplenetinnerrange2}
        \end{subfigure}
        \begin{subfigure}[b]{0.24\textwidth}
        \centering
        % Sketch output, version 0.3 (build 7d, Thu Mar 18 10:53:46 2021)
% Output language: PGF/TikZ,LaTeX
\begin{tikzpicture}[line join=round,scale=0.5]
\filldraw[fill=magenta,line width=1pt,fill opacity=0.8](-1,-3)--(1,-1)--(1,1)--(-1,-1)--cycle;
\draw[arrows=->,line width=1pt,dotted](-1.5,0)--(2,0);
\draw[arrows=->,line width=1pt,dotted](0,-3.5)--(0,2);
\path (-1,0) node[below] {$-1$}
     (0,0) node[below] {$0$}
     (1,0) node[below] {$1$}
     (0,1) node[right] {$1$}
     (0,2) node[right] {$\textcolor{magenta}{h_1}$}
     (2,0) node[below] {$x_1$}
     (0,-2) node[right] {$-2$};\end{tikzpicture}% End sketch output
        \subcaption{$\color{magenta}(x_1,h_1)$} %\\ \mbox{ }}
        \label{fig:simpleneuralnet:zone}
        \end{subfigure}
                \begin{subfigure}[b]{0.24\textwidth}
        \centering
        % Sketch output, version 0.3 (build 7d, Thu Mar 18 10:53:46 2021)
% Output language: PGF/TikZ,LaTeX
\begin{tikzpicture}[line join=round,scale=0.5]
\draw[arrows=->,line width=1pt,dotted](-1.5,0)--(2,0);
\draw[arrows=->,line width=1pt,dotted](0,-3.5)--(0,2);
\filldraw[fill=magenta,line width=1pt,fill opacity=0.5,dashed](-1,-3)--(1,-1)--(1,1)--(-1,-1)--cycle;
\filldraw[fill=cyan,line width=0pt,fill opacity=0.8](0,0)--(1,0)--(1,1)--cycle;
\draw[color=cyan,line width=2pt](-1,0)--(0,0);
\filldraw[fill=red,line width=1pt,fill opacity=0.8,dashed](-.25,.5)--(.25,.5)--(.25,1)--(-.25,1)--cycle;
\path (-1,0) node[below] {$-1$}
     (0,0) node[below] {$0$}
     (1,0) node[below] {$1$}
     (0,1) node[right] {$1$}
     (0,2) node[right] { $\textcolor{magenta}{h_1}$/$\textcolor{cyan}{y_1}$}
     (2,0) node[below] {$x_1$}
     (0,-2) node[right] {$-2$};\end{tikzpicture}% End sketch output
        \subcaption{$\color{magenta}(x_1,h_1)$; $\color{cyan} (x_1,y_1)$}
        \label{fig:simpleneuralnet:zone2}
        \end{subfigure}
    \caption{Exact ranges for the neural net of Example \ref{ex:running} on $[-1,1]\times [-1,1]$. ($P_2$) is the complement of the red square in Fig.~\ref{fig:simpleneuralnet:zone2}.}
    \label{fig:simpleneuralnet}
\end{figure}

%As we will see in Section \ref{sec:troppoly}, the graph $y_1$ as a function of $x_1$ is a tropical polyhedron, that is not convex in the classical sense. 
%
We want to check two properties on this simple neural network: 
\begin{enumerate}
    \item [($P_1$): ] the input is always classified as belonging to the class identified by neuron $y_2$, i.e. we always have $y_2\geq y_1$ 
    \item [($P_2$): ] in the neighborhood [-0.25,0.25] of 0 for $x_1$,  whatever $x_2$ in [-1,1], the output $y_1$ is never above threshold 0.5 (unsafe zone materialized in red in Fig. \ref{fig:simpleneuralnet:zone2})
    \end{enumerate}
($P_2$) is a robustness property.     
We see on the blue part of Figure \ref{fig:simplenetinnerrange2} (resp. \ref{fig:simpleneuralnet:zone2}) that the first (resp. second) property is true. 

As we will see in Section \ref{sec:1layer}, our tropical polyhedron abstraction is going to give the exact graph of $y_1$ as a function of $x_1$, in cyan again, Figure \ref{fig:simpleneuralnet:trop2}.

\begin{figure}
    \centering
    \begin{subfigure}[b]{0.24\textwidth}
    \centering
      % Sketch output, version 0.3 (build 7d, Thu Mar 18 10:53:46 2021)
% Output language: PGF/TikZ,LaTeX
\begin{tikzpicture}[line join=round,scale=0.5]
\draw[arrows=->,line width=1pt,dotted](-3.5,0)--(2,0);
\draw[arrows=->,line width=1pt,dotted](0,-1.5)--(0,3.5);
\filldraw[fill=magenta,line width=1pt,fill opacity=0.5,dashed](-3,-1)--(-1,-1)--(1,1)--(1,3)--(-1,3)--(-3,1)--cycle;
\filldraw(-3,-1) circle (2pt);
\filldraw(1,1) circle (2pt);
\filldraw(-1,3) circle (2pt);
\filldraw[fill=cyan,line width=1pt,fill opacity=0.8](0,0)--(1,1)--(1,3)--(0,3)--cycle;
\filldraw(0,0) circle (2pt);
\filldraw(0,3) circle (2pt);
\path (-2,0) node[below] {$-2$}
     (0,0) node[below] {$0$}
     (1,0) node[below] {$1$}
     (0,2) node[right] {$2$}
     (0,3.5) node[right] {$\textcolor{cyan}{y_2}$/$\textcolor{magenta}{h_2}$}
     (2.5,0) node[below] {$\textcolor{cyan}{y_1}$/$\textcolor{magenta}{h_1}$}
     (0,-2) node[right] {$-2$};\path (-1,3) node[left] {$B_2$}
     (1,1) node [below] {$B_1$}
     (-3,-1) node[below] {$A$};\path (0,3) node [left] {$B'_2$}
     (.5,0) node[below] {$A'$};
     
     %\node at (-0.5,-3) {\textbf{(a)}};
     \end{tikzpicture}% End sketch output
	  \subcaption{With zone, tropical polyhedra}
        \label{fig:simplenet:trop}
        \end{subfigure}
        \begin{subfigure}[b]{0.24\textwidth}
        \centering
        % Sketch output, version 0.3 (build 7d, Thu Mar 18 10:53:46 2021)
% Output language: PGF/TikZ,LaTeX
\begin{tikzpicture}[line join=round,scale=0.5]
\draw[arrows=->,line width=1pt,dotted](-1.5,0)--(2,0);
\draw[arrows=->,line width=1pt,dotted](0,-3.5)--(0,2);
\filldraw[fill=magenta,line width=1pt,fill opacity=0.5,dashed](-1,-3)--(1,-1)--(1,1)--(-1,-1)--cycle;
\filldraw[fill=cyan,line width=0pt,fill opacity=0.8](0,0)--(1,0)--(1,1)--cycle;
\draw[color=cyan,line width=2pt](-1,0)--(0,0);
\filldraw[fill=red,line width=1pt,fill opacity=0.8,dashed](-.25,.5)--(.25,.5)--(.25,1)--(-.25,1)--cycle;
\draw[color=black,line width=1pt,dashed](-1,0)--(1,1);
\path (-1,0) node[below] {$-1$}
     (0,0) node[below] {$0$}
     (1,0) node[below] {$1$}
     (0,1) node[right] {$1$}
     (0,2) node[right] {$\textcolor{magenta}{h_1}$/$\textcolor{cyan}{y_1}$}
     (2,0) node[below] {$x_1$}
     (0,-2) node[right] {$-2$};
     %\node at (0,-4)  {\textbf{(b)}};
     %\node at (0, -5){};
     \end{tikzpicture}% End sketch output
        \subcaption{$\color{magenta}(x_1,h_1)$; $\color{cyan} (x_1,y_1)$\\ \ \\}
        \label{fig:simpleneuralnet:trop2}
        \end{subfigure}
        \begin{subfigure}[b]{0.24\textwidth}
        \centering
        % Sketch output, version 0.3 (build 7d, Thu Mar 18 10:53:46 2021)
% Output language: PGF/TikZ,LaTeX
\begin{tikzpicture}[line join=round,scale=0.5]
\filldraw[fill=magenta,line width=1pt,fill opacity=0.5,dashed](-3,0)--(-2,0)--(0,2)--(0,3)--(-1,3)--(-3,1)--cycle;
\filldraw(-3,0) circle (2pt);
\filldraw(0,2) circle (2pt);
\filldraw(-1,3) circle (2pt);
\filldraw[fill=magenta,line width=1pt,fill opacity=0.5,dashed](-2,-1)--(-1,-1)--(1,1)--(1,2)--(0,2)--(-2,0)--cycle;
\filldraw(-2,-1) circle (2pt);
\filldraw(1,1) circle (2pt);
\filldraw(0,2) circle (2pt);
\draw[arrows=->,line width=1pt,dotted](-3.5,0)--(2,0);
\draw[arrows=->,line width=1pt,dotted](0,-1.5)--(0,3.5);
\filldraw[fill=cyan,line width=1pt,fill opacity=0.8](0,0)--(1,1)--(1,2)--(0,2)--cycle;
\filldraw(0,0) circle (2pt);
\filldraw(1,1) circle (2pt);
\filldraw(0,2) circle (2pt);
\path (-2,0) node[below] {$-2$}
     (0,0) node[below] {$0$}
     (1,0) node[below] {$1$}
     (0,2) node[right] {$2$}
     (0,3.5) node[right] {$\textcolor{cyan}{y_2}$/$\textcolor{magenta}{h_2}$}
     (2.5,0) node[below] {$\textcolor{cyan}{y_1}$/$\textcolor{magenta}{h_1}$}
     (0,-2) node[right] {$-2$};\path (0,2) node [left] {$B'_2$}
     (.5,0) node[below] {$A'$}
     (1,1) node [below] {$B_1$}
     ;
     
     %\node at (0, -3) {\textbf{(c)}};
     \end{tikzpicture}% End sketch output
        \subcaption{Once subdivided tropical polyhedra}
        \label{fig:simplenet:1subd}
        \end{subfigure}
        \begin{subfigure}[b]{0.24\textwidth}
        \centering
        % Sketch output, version 0.3 (build 7d, Thu Mar 18 10:53:46 2021)
% Output language: PGF/TikZ,LaTeX
\begin{tikzpicture}[line join=round,scale=0.5]
\filldraw[fill=magenta,line width=1pt,fill opacity=0.5,dashed](-3,.5)--(-2.5,.5)--(-.5,2.5)--(-.5,3)--(-1,3)--(-3,1)--cycle;
\filldraw(-3,.5) circle (2pt);
\filldraw(-.5,2.5) circle (2pt);
\filldraw(-1,3) circle (2pt);
\filldraw[fill=magenta,line width=1pt,fill opacity=0.5,dashed](-1.5,-1)--(-1,-1)--(1,1)--(1,1.5)--(.5,1.5)--(-1.5,-.5)--cycle;
\filldraw(-1.5,-1) circle (2pt);
\filldraw(1,1) circle (2pt);
\filldraw(.5,1.5) circle (2pt);
\filldraw[fill=magenta,line width=1pt,fill opacity=0.5,dashed](-2.5,0)--(-2,0)--(0,2)--(0,2.5)--(-.5,2.5)--(-2.5,.5)--cycle;
\filldraw(-2.5,0) circle (2pt);
\filldraw(0,2) circle (2pt);
\filldraw(-.5,2.5) circle (2pt);
\filldraw[fill=magenta,line width=1pt,fill opacity=0.5,dashed](-2,-.5)--(-1.5,-.5)--(.5,1.5)--(.5,2)--(0,2)--(-2,0)--cycle;
\filldraw(-2,-.5) circle (2pt);
\filldraw(.5,1.5) circle (2pt);
\filldraw(0,2) circle (2pt);
\draw[arrows=->,line width=1pt,dotted](-3.5,0)--(2,0);
\draw[arrows=->,line width=1pt,dotted](0,-1.5)--(0,3.5);
\filldraw[fill=cyan,line width=0pt,fill opacity=0.8](0,0)--(1,1)--(1,1.5)--(0,1.5)--cycle;
\filldraw[fill=cyan,line width=0pt,fill opacity=0.8](0,0)--(.5,.5)--(.5,2)--(0,2)--cycle;
\path (-2,0) node[below] {$-2$}
     (0,0) node[below] {$0$}
     (1,0) node[below] {$1$}
     (0,2) node[right] {$2$}
     (0,3.5) node[right] {$\textcolor{cyan}{y_2}$/$\textcolor{magenta}{h_2}$}
     (2.5,0) node[below] {$\textcolor{cyan}{y_1}$/$\textcolor{magenta}{h_1}$}
     (0,-2) node[right] {$-2$};
      %\node at (0, -3) {\textbf{(d)}};
     \end{tikzpicture}% End sketch output
        \subcaption{twice subdivided tropical polyhedra.}
                \label{fig:simplenet:2subd}
        \end{subfigure}
    \caption{Abstractions of a simple neural net on $[-1,1]\times [-1,1]$. Dashed lines in (b) enclose the classical convexification.}
    \label{fig:simpleneuralnet}
\end{figure}
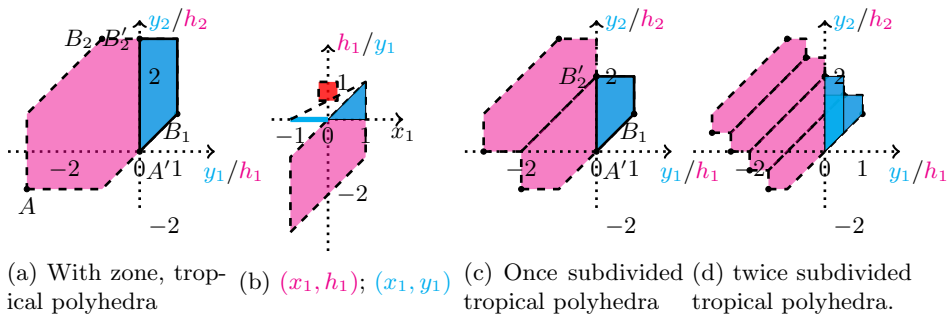

Therefore we will be able to prove robustness, %\footnote{In general, such "classification" properties are directly expressible in terms of zones. A direct verification or synthesis of these specifications can be done from our tropical polyhedric abstraction by computing the enclosing zone of it, see \cite{SAS2008} and Section \ref{sec:background:zone}.}, 
i.e. ($P_2$): the exact range for $y_1$ in cyan does not intersect the non complying states, in red. Note that all classically convex abstractions, whatever their intricacies, will need to extend the cyan zone up to the dashed line pictured in Figure \ref{fig:simpleneuralnet:trop2}, to get the full triangle, at the very least. This triangle is intersecting the red region making classically convex abstractions unable to prove ($P_2$). 

Our tropical abstraction projected on the $y_2$, $y_1$ coordinates is not exact: compare the exact range in cyan in Figure \ref{fig:simplenetinnerrange2} with the abstraction in cyan in Figure \ref{fig:simplenet:trop}. However, the cyan region in Figure \ref{fig:simplenet:trop} is above the diagonal, which is enough for proving ($P_1$). 

Still, the abstraction has an area 2.5 times larger than the exact range, due to the tropical linearization of the tropical rational function $y_1$. As with classical linearizations, a workaround is to make this linearization local, through suitable subdivisions of the input. We show in Figure \ref{fig:simplenet:1subd} the tropical polyhedric abstraction obtained by subdividing $x_1$ into two sub-intervals (namely $[-1,0]$ and $[0,1]$): the cyan part of the picture is much closer to the exact range (1.5 times the exact area). Subdividing further as in Figure \ref{fig:simplenet:2subd} naturally further improves the precision (area 1.25 times the exact one). %The area can actually be shown to converge to the exact one, with a factor of $1+1/k$, where $k>1$ is the number of (regular) subdivisions on $x_1$. 

As we will see in Section \ref{sec:troppoly}, tropical polyhedra are particular unions of zones: the tropical polyhedra in cyan of Figures \ref{fig:simplenet:trop} and \ref{fig:simplenet:1subd} are composed of just one zone, but the tropical polyhedron in cyan in Figure \ref{fig:simplenet:2subd} and the tropical polyhedron in magenta in Figure \ref{fig:simplenet:1subd} are the union of two zones. Finally, the tropical polyhedron in magenta in Figure \ref{fig:simplenet:2subd} is the union of four zones (generated by 9 extreme points, or 5 constraints, obtained by joining results from the subdivisions of the inputs). %\todo{Pas sure de comprendre le discours vu qu'il y a 4 subdivisions?} 
%The extra advantage of using tropical polyhedra instead of ad-hoc representations of unions of zones 
%is that we have a very compact and efficient representation 
%of these unions in just one (tropical) polyhedron, with $k+2$ extreme points, or equivalently, $k+1$ tropical constraints generating it, for $k$ subdivisions. %, and for which we have an exact and direct characterization, see Section \ref{sec:subd}.
\end{example}

%\todo{Ne faudrait-il pas definir rapidement polyedres tropicaux voire zones pour rendre cette introduction moins abstraite pour le non specialiste?}
%, and thus be used in more (memory and time) contrained (embedded systems) environments.

%\todo{Mention somewhere the classification problem for networks, suitably defined using DBMs on the output neurons, or just focus on the range? Should add some on the relations we find...so on the enveloping zone}

%Mention binary/ternary networks
%Similarly, for recurrent RELU-based neural nets, max policy iteration 
%\cite{policy}
%may
%prove very useful for dealing with the necessary fixed-point computation. 

%Abstract interpretation ref \cite{AI}. 

%Zones 
%ref \cite{Dill}, 
%\cite{Uppaal}, 
%\cite{Mine2}, 
%octagons ref \cite{Mine}. 

%Disjunctive analysis ref? 

\paragraph{Contributions.}

Section \ref{sec:background} introduces the necessary background notions, in particular tropical polyhedra. We then describe the following contributions: 
\begin{itemize}
    \item Section \ref{sec:1layer} introduces our abstraction of (classical) affine functions from $\R^m$ to $\R^n$ with tropical polyhedra. We fully describe internal and external representations, extending the classical abstractions of assignments in the zone abstract domain \cite{Mine2} or in the tropical polyhedra domain \cite{SAS2008}. We prove correctness and equivalence of internal and external representations, allowing the use of the double description method \cite{Tropical}. 
    \item Based on the analysis of one layer networks of Section \ref{sec:1layer}, we show in Section \ref{sec:multilayer} how to get to multi-layered networks. 
    %\item The methods of Section \ref{sec:1layer} rely on a tropical linearization of classical affine functions (which are non-linear, tropically). Of course this is only precise locally, and for larger input ranges, it is important to be able to subdivide the input domain to regain precision. To get an internal description, we can just analyze subdomains separately and take the tropical polyhedron union, which is just concatenation of internal representations on each of the subdomains. We show that we can get an external representation directly as well, that we develop in Section \ref{sec:subd}. In the general case of the abstraction of affine functions $f: \R^m \rightarrow \R^n$ this external representation is only an outer approximation of the union on subdomains, but is exact when $n=m=1$. 
    %
%    \todo{Ordre a inverser avec point precedent?}
    \item Finally, Section \ref{sec:implexp} describes our implementations in C++ and using polymake \cite{polymake} and presents some promising experiments. We discuss  the cost and advantages of using the double description or of relying for further abstraction on either internal or external representations of tropical polyhedra. 
\end{itemize}

\paragraph{Related work.}

%VNN etc.

There exist many approaches to neural networks verification. We concentrate here on methods and tools designed for at least range over-approximation of ReLU feedforward networks. %We ignore here e.g. methods for analyzing neural nets used as controllers, within a feedback loop as they are not specifically relevant to our paper. 

%\paragraph{Constraint based methods for ReLU feedforward networks: }

It is natural to consider constraint based methods for encoding the ReLU function and the combinatorics of activations in a ReLU feedforward neural net. 

Determining the range of a ReLU feedforward neural net amounts to solving min and max problems under sets of linear and ReLU constraints. This can be solved either by global optimization techniques and branch and bound mechanisms, see e.g. DeepGo \cite{ijcai18}. 
The encoding of the activation combinatorics can also be seen as mixed integer linear constraints, and MILP solver used for solving the range outer-approximation problem, see e.g. 
\cite{mipverify}, \cite{ilp}, %\cite{duality}, %\cite{fastlinlip}, 
or both branch and bound and MILP techniques, like Venus~\cite{Venus}. 
Similarly, Sherlock \cite{sherlock2,sherlock-tool} performs 
range analysis using optimization methods (MILP and a combination of local search and global branch-and-bound
approach), and considers also neural nets as controllers within a feedback loop. Finally, some of these constraint-based analyzer improve the solution search by exploiting the geometry of the activation regions, \cite{PeregriNN}.

A second category of such approaches is based on SMT methods, more specifically satisfiability modulo extensions of linear real arithmetic (encoding also ReLU). The network is encoded in this logics and solvers provide answers to queries, in particular range over-approximation and robustness, see e.g.  
{Marabou} \cite{Marabou}, extending Reluplex \cite{reluplex}, and \cite{Ehlers1}, \cite{dlv}.

%Finally, some authors have been designing verification methods specifically for binary/ternary neural nets, by encoding their semantics as first order propositional logical formulars, see e.g. NPAQ \cite{NPAQ}. 

%NPAQ takes as input a BNN and one of the properties to quantify: robustness, fairness and trojan attack success. NPAQ, then, outputs an estimate of how often the property is satisfied for the given neural network. The property is defined as a simple constraint over inputs and outputs.
%In a nutshell, NPAQ encodes the BNN and the property to a specification as a Boolean logical formula in conjunctive normal form. In the encoding we show that we can preserve the number of solutions from the neural network space of solutions to the Boolean logical formula’s solution space. The last step is to estimate the number of solutions with PAC-style guarantees using an approximate model counter.

%\href{PaRoT from FiveAI}{https://five.ai/}: based on \href{paper}{https://arxiv.org/abs/2001.02152}
%Commercial, classical, for robust training (uses zonotopes)

%\href{ARFramework}{https://github.com/formal-verification-research/ARFramework}
% not much info

% Optim/SMT etc.
% --------------

%SMT solvers have also been successfully used for proving robustness 
%properties (absence of adversarial inputs) of neural nets, see e.g. \cite{reluplex,Ehlers1}

% Proof based...
% --------------

%\paragraph{Set-based/abstract interpretation based methods: }

Range estimation for ReLU activated feedforward neural nets can also be performed using some of the abstract domains \cite{Cousot} that have been designed for program analysis, and in particular convex domains for numerical program verification. These include zonotopes~\cite{AI2,DeepZ}, especially considering that feedforward neural nets with one hidden layer and ReLU activation functions are known to be characterizable by zonotopes, see e.g. \cite{RELU}, polyhedra \cite{DeepPoly},
and other sub-polyhedric or convex abstractions like symbolic intervals \cite{ibp} used in Neurify \cite{neurify} extending Reluval~\cite{reluval} or CROWN-IBP~\cite{CROWNIBP}. 
These abstractions allow to perform range estimation, i.e. to estimate outer approximations of the values of the output neurons given a set of values for the input neurons. They also allow to deal with robustness properties around training data, by proving that the range of the neural net on a small set around a training point gives the same class of outputs. 
%is such that it would classify the same class as this training data means the neural net exhibits some form of robustness.
%Doing range estimation allows for dealing with robustness properties around training
%points, as done in e.g. AI$^2$ and DeepZ.

% Improvements on union/case analysis
% -----------------------------------
The main difficulty with these convex abstract domains is that they tend to lose too much precision on (non-convex) ReLU functions. Several methods have been proposed to cope with this phenomenon. 
The first one is to improve on the abstraction of ReLU, in particular by combining the abstraction of several ReLU functions on the same layer  
\cite{krelu}. 
Another solution that has been proposed in the literature is to combine abstraction and some level of combinatorial exploration of the possible neuron activations, in the line of disjunctive program analysis \cite{Bourdoncle92abstractinterpretation,partitioning}. 
RefineZono \cite{singh2019refinement} implements methods combining polyhedric abstract domains with MILP solvers for encoding ReLU activation and refining the abstractions, NNENUM \cite{nnenum} uses combinations of zonotopes, stars sets with case splitting methods, and Verinet \cite{Verinet} 
uses abstractions similar to the polyhedric relaxations of DeepPoly, based on symbolic-interval propagation, with adaptive refinement strategies. 

\section{Preliminaries and notations}
\label{sec:background}

\subsection{Zones}

\label{sec:background:zone}

The \textit{zone}~\cite{Mine2} abstraction represents restricted forms of affine invariants over variables, bounds on variable differences. Let a n-dimensional variable $x=(x_1,\ldots,x_n) \in \R^n$. The zone domain represents 
invariants of the form $(\bigwedge\limits_{1 \leq i,j \leq n} x_i -x_j \leq c_{i,j}) \wedge (\bigwedge\limits_{1 \leq i \leq n} a_i \leq x_i \leq b_i)$. 
A convenient representation is using \textit{difference bound matrices}, or DBM.  In order to encode interval constraints seamlessly in this matrix, a special variable $x_0$, which is assumed to be a constant set to zero, is added to $x \in \R^n$. A DBM is then a $(n+1) \times (n+1)$ square matrix $C=(c_{ij})$, with elements in $\R \cup \{+ \infty\}$, representing (concretization operator)
%\begin{itemize}
%    \item $c_{ij} \in \R$, for $i \in [1,n]$ and $j \in [1,n]$ denotes a constraint $x_i - x_j \leq c_{i,j} $, while $c_{ij}=  \{+ \infty\}$ denotes the absence of any constraint
%    \item $c_{0j} \in \R$ for $j \in [1,n]$ corresponds to a constraint $x_0 - x_j \leq c_{0,j} $ or equivalently $x_j \geq -c_{0,j}$, and $c_{i0} \in \R$, for $i \in [1,n]$ corresponds to a constraint $x_i - x_0 \leq c_{i,0} $ or equivalently $x_i \leq c_{i,0} $. %In other words, the first line and first column of the DBM encode interval bounds on variables. 
%\end{itemize}
%A DBM $C$ represents 
the following set of points in $\R^n$: 
$\gamma(C) = \{ (x_1,\ldots,x_n) \in \R^n | \; \forall i, j \in [0,n], x_i -x_j \leq c_{i,j} \wedge x_0 = 0 \}$. 
%Mention the encoding of $n$-dimensional zones on ($x_0$,\ldots,$x_n$) with $x_0$ the extra variable, always equal to zero. 

%The pointwise extension on DBM of the arithmetic order $\leq$ on $\R \cup \{+ \infty\}$ gives a partial order. However, there can exist several matrices that represent the same set of points in $\R^n$. A normal form is thus needed, which will be given by a closure operation. 
For a matrix $C$ that has non-empty concretization, the closure denoted $C^*$ will be the smallest DBM for the partial order on matrices which represents $\gamma(C)$.  Formally, a closed zone $C=(c_{ij})$ is such that:
%    \begin{align*}
  $  \forall k \in \N, \forall (i_0,\ldots,i_k) \in [0,n]^{k+1}, \;  c_{i_0,i_k}\leq c_{i_0,i_1}+\dots+c_{i_{k-1},i_k}$, $
    \forall i\in[0,j], \; c_{i,i}=0$.
%    \end{align*}
Every constraint in a closed zone saturates the set $\gamma(C)$. 

The best abstraction in the sense of abstract interpretation \cite{Cousot} of a non-empty set $S\subset\R^n$ is the zone defined by the closed DBM:
$(c)_{ij} = sup \{x_i-x_j | \; (x_1,\ldots,x_n) \in S \wedge x_0= 0 \}$.
    
%    \begin{proposition}[Saturated zones are closed]\-
%    \label{prop:optimal_zone}
%\todo{Sylvie deplacer dans section 2.1 zones, ptet ecrire zone=($\delta_{i,j}$) que la meilleure abstraction c'est ce truc en dessous, qu'elle est fermee, et la concretization est la definition d'une zone, et juste citer un article d'Antoine Mine.}
%    Let $S\subset\R^n$ a non-empty set. 
%   Let $a_0:=0$, and, 
%    for each $i,j\in[0,n]$, let
%    $$\delta_{i,j} := \sup_{x\in S}(x_i-x_j)$$
%    Then, the zone $H$ defined by $(\delta_{i,j})$ is the tightest zone containing $S$.
%    
%    Moreover, the family $(\delta_{i,j})$ verifies the closure hypothesis.
%\end{proposition}

%\begin{proof}
 %   This directly follows from the definition of $\sup$ and the fact that for all functions $f,g:\R^n\to\R$,
 %   $$\sup_{x\in S}(f(x)+g(x)) \leq \sup_{x\in S}f(x)+\sup_{x\in S}g(x).$$
%\end{proof}
    
    \begin{example}
    \label{ex:running2}
    Consider the region defined as the union of the magenta and cyan parts of 
Figure \ref{fig:simplenet:trop} in Example \ref{ex:running}. 
    It is a zone given by the inequalities:
$(-3 \leq h_1 \leq 1) \wedge (-1 \leq h_2 \leq 3) \wedge (-4 \leq h_1 - h_2 \leq 0)$, i.e. given by the following DBM:
$$\left(\begin{array}{ccc}
0 & 3 & 1 \\
1 & 0 & 0 \\
3 & 4 & 0 \\
\end{array}\right)$$
    \end{example}
    
The \textit{octagon}~\cite{Mine} abstraction is an extension of the zone abstraction, which represents constraints of the form  \[ (\bigwedge_{1 \leq i,j \leq n} \pm x_i \pm x_j \leq c_{i,j}) \wedge (\bigwedge_{1 \leq i \leq n} a_i \leq x_i \leq b_i) \]   
A set of octagonal constraints can be encoded as a difference bound matrix, similarly to the case of zones, but using a variable change to map octagonal constraints on zone constraints. For each variable $x_i$, two variables are considered in the DBM encoding, that correspond respectively to   $+ x_i$ and $- x_i$. 
Note that unary (interval) constraints, such as $x_i \leq b_i$, can be encoded directly as
$x_i + x_i \leq 2 b_i$, so that no additional variable $x_0$ is needed. 

\begin{example}
\label{ex:running3}
The figure below right shows the exact range (the rotated square) of $h_1$, $h_2$ of Example \ref{ex:running}. 
%\todo{Exemplify the positive/negative variables contruction on $h_1$, $h_2$ of Example \ref{ex:running}.}
%(we will use this again in the tropical construction later, in particular for $y_1$, $y_2$)

\begin{minipage}{.7\textwidth}
%Figure \ref{fig:simplenetinnerrange} 
% These two zones are particular zone contraints on positive/negative variables...
\noindent It is depicted in gray, as the intersection of two zones, one in cyan, $Z_2$, and one in olive, $Z_1$. $Z_1$ is the zone defined in Example \ref{ex:running2} and $Z_2$ is the zone defined on variables $(h_1,-h_2)$ as follows: 
$$ (-3 \leq h_1 \leq 1) \wedge (-1 \leq h_2 \leq 3) \wedge (-2 \leq h_1 + h_2 \leq 2)$$
\end{minipage}
\begin{minipage}{.3\textwidth}
    \centering
    % Sketch output, version 0.3 (build 7d, Thu Mar 18 10:53:46 2021)
% Output language: PGF/TikZ,LaTeX
\begin{tikzpicture}[line join=round,scale=0.6]
\draw[arrows=->,line width=1pt,dotted](-3.5,0)--(2,0);
\draw[arrows=->,line width=1pt,dotted](0,-1.5)--(0,3.5);
\filldraw[fill=olive,line width=1pt,fill opacity=0.5,dashed](-3,-1)--(-1,-1)--(1,1)--(1,3)--(-1,3)--(-3,1)--cycle;
\filldraw[fill=cyan,line width=1pt,fill opacity=0.5,dashed](-3,3)--(-3,1)--(-1,-1)--(1,-1)--(1,1)--(-1,3)--cycle;
\filldraw[fill=lightgray,line width=1pt,fill opacity=0.8](-3,1)--(-1,-1)--(1,1)--(-1,3)--cycle;
\path (-2,0) node[below] {$-2$}
     (0,0) node[below] {$0$}
     (1,0) node[below] {$1$}
     (0,2) node[right] {$2$}
     (0,3.5) node[right] {${h_2}$}
     (2.5,0) node[below] {${h_1}$}
     (0,-2) node[right] {$-2$};\end{tikzpicture}% End sketch output
    %\caption{The octagon abstraction of the linear and ReLU layers of Example \ref{ex:running}.}
    %\label{fig:simplenet:octagon}
\end{minipage}
\end{example}

\subsection{Tropical polyhedra}

\label{sec:troppoly}

%\subsubsection{Definitions and basic properties}

Tropical polyhedra are similar to ordinary convex polyhedra. Both can be defined either using affine constraints, known as the external description, or as convex hulls of extremal points and rays, known as the internal description. The major difference is the underlying algebra. Instead of using the classical ring $\R$ of coefficients, with ordinary sum and multiplications, we use the so-called max-plus semiring $\R_{max}$. This semiring is based on the set $\R_{max}=\R \cup \{-\infty\}$, equipped with the addition $x \oplus y:=max(x,y)$ and the multiplication $x \otimes y:=x+y$. 
This is almost a ring: we have neutral elements $\one:=0$ for $\otimes$, and $\zero:=-\infty$ for $\oplus$, and an inverse for $\otimes$ on $\R_{max}\backslash \{\zero\}$ but not for $\oplus$. The algebra also fits in with the usual order $\leq$ on $\R$, extended to $\R_{max}$: $x \leq y$ if and only if $x \oplus y=y$. 

%A classical hyperplane in $\R^d$ is defined as the set of vectors $x=(x_i) \in \R^d$ which satisfies an affine inequality of the form $\sum_{i=1}^d a_i x_i \leq c$. 
%We would be tempted to define hyperplanes in a similar way in the tropical world, changing sums into max and scalar multiplication by scalar translation 
%$\max\limits_{1 \leq i \leq d} \{ a_i + x_i\} \leq c$, or, with the max plus notations:
%$\mathop{\bigoplus}\limits_{1 \leq i \leq d} a_i \otimes x_i \leq c. $
%But $\R_{max}$ being only a semi-ring, $\oplus$ has no inverse, and we need to put "negative" $a_i$ on the right hand-side of the inequality, i.e. 
%\todo{pas clair pour moi. Utiliser running ex pour ca ; il y a tous les types d'enveloppes convexes dans cet exemple.}
%\begin{equation}
Tropical hyperplanes are similar to classical hyperplanes, and defined as the set of points satisfying 
$\mathop{\bigoplus}\limits_{1 \leq i \leq k} a_i \otimes x_i \oplus c \leq \mathop{\bigoplus}\limits_{1 \leq i \leq k} b_i \otimes x_i  \oplus d
%\label{eq:trophyper}
$. %\end{equation}

Now, as in the classical case, tropical polyhedra will be given (externally) as an intersection of $n$ tropical hyperplanes, i.e. will be given as the location of points in $\R_{max}^k$ satisfying $n$ inequalities of the form of above. %Equation (\ref{eq:trophyper}). 
This can be summarized using matrices $A=(a_{ij})$ and $B=(b_{ij})$, two $n\times k$ matrices with entries in $\R_{max}$, and vectors of size $k$ $C$ and $D$ as $A x\oplus C \leq Bx \oplus D$.

Still similarly to the case of ordinary convex polyhedra, tropical polyhedra can also be described internally, as generated by extremal generators (points, rays). A tropical polyhedron can then be defined as the set of vectors $x \in \R_{max}^k$ which can be written as a tropical affine combination of generators $v^i$ (the extreme points) and $r^j $ (the extreme rays) as $x=\mathop{\bigoplus}\limits_{i \in I} \lambda_i v^i \oplus \mathop{\bigoplus}\limits_{j \in J} \mu_j r^j
%$$ \noindent 
\mbox{ \ with $\mathop{\bigoplus}\limits_{i \in I} \lambda_i = \one$}$. 

\begin{example}[Running example]
\label{ex:running4}
%\todo
Consider again the zone consisting of the union of the magenta and cyan parts in 
Figure \ref{fig:simplenet:trop}. 
This is a tropical polyhedron, defined externally by: 
%$$max\left(\begin{array}{c}
%h_1 \\
%-3 \\
%h_2\\
%-1 \\
%h_2 \\
%h_1
%\end{array}\right)\leq 
%max\left(\begin{array}{c}
%1\\
%h_1 \\
%%3 \\
%h_2 \\
%h_1+4 \\
%h_2 
%\end{array}\right)
%$$
$max\left(
h_1, 
-3, 
h_2, 
-1, 
h_2, 
h_1
%\end{array}
\right)\leq 
max\left( 
1, 
h_1, 
3, 
h_2, 
h_1+4, 
h_2 
\right)
$.

It can also be defined internally by the extremal point $A$, $B_1$ and $B_2$ of respective coordinates $(-3,-1)$, $(1,1)$ and $(-1,3)$, depicted as dots in Figure \ref{fig:simplenet:trop}. This means that the points $z$ in this tropical polyhedron have coordinates $(h_1,h_2)$ with 
$(h_1,h_2)=max\left(\lambda_0+A,\lambda_1+B_1,\lambda_2+B_2\right)$
with $max(\lambda_0,\lambda_1,\lambda_2)=\one=0$, i.e. all $\lambda_i$s are negative or null, and one at least among the $\lambda_i$s is zero. 

For instance, when $\lambda_2=-\infty$, $z$ is on the tropical line linking $A$ to $B_1$: 
\begin{equation}
\label{eq:hext}
\left(
\begin{array}{c}
h_1, 
h_2
\end{array}
\right)=\left( \begin{array}{c}
max(\lambda_0-3,\lambda_1-1), 
max(\lambda_0 -1,\lambda_1+3)
\end{array}
\right)
\end{equation}
\noindent with $\lambda_0, \lambda_1\neq 0$ and either $\lambda_0=0$ or $\lambda_1=0$. Suppose $\lambda_0=0$, and suppose first that  
$\lambda_1 \leq -4$: $(h_1,h_2)=(-3,-1)$ which is point $A$. Suppose now $-4\leq \lambda_1\leq -2$, then 
$(h_1,h_2)=(-3,\lambda_1+3)$, which is the vertical line going from $A$ to point $(-3,1)$. Finally, suppose $-2\leq \lambda_1 \leq 0$, $(h_1,h_2)=(\lambda_1-1,\lambda_1+3)$ which is the diagonal going from $(-3,1)$ to $B_1$. 
Similarly, one can show that the tropical line going from $B_1$ to $B_2$ is given by fixing $\lambda_0=-\infty$ and making vary $\lambda_1$ and $\lambda_2$. If $\lambda_0< 0$ then $\lambda_1=0$ and $z$ is point $B_1$. 
%Show the 3 tropical segments in dim 2, they are $AB_1$, $AB_2$ and $B_1B_2$, and explain tropical convexity - with some computations.

Now, applying the ReLU operator, which is linear in the tropical algebra, defines a tropical polyhedron 
%which has as external description the same as the one given in Equation \ref{eq:hext}, where $h_1$
with internal description given by ReLU (in each coordinate) of extreme points $A$, $B_1$ and $B_2$, i.e. $A'=(0,0)$, $B'_1=B_1=(1,1)$ and  $B'_2=(0,3)$, see Figure \ref{fig:simplenet:trop}. Similarly, the zone which gives $h_1$ as a function of $x_1$, see Figure \ref{fig:simpleneuralnet:trop2}, can be seen as a tropical polyhedron with extreme points $(-1,-3)$, $(1,1)$ and $(1,-1)$. Applying ReLU to the second coordinate of these three extreme points gives three points $(-1,0)$, $(1,1)$ and $(1,0)$ which generate the tropical polyhedron in cyan of Figure \ref{fig:simpleneuralnet:trop2}. 

It is also easy to see that after one subdivision, Figure \ref{fig:simplenet:1subd}, the set of values for $(y_1,y_2)$ in cyan is a tropical polyhedron with three extreme points $A'$, $B'_1$ and $B_2$. After two subdivisions, Figure \ref{fig:simplenet:2subd}, the values of $y_1$ as a function of $h_1$ is a tropical polyhedron with 4 generators (depicted as dots in Figure \ref{fig:simplenet:2subd}). Note that the tropical polyhedron of Figure \ref{fig:simplenet:2subd} is the encoding of the union of two zones, one zone being the classical convex hull of points $(0,0)$, $(0,1)$, $(0.5,1.5)$, $(1,1.5)$ and $(1,1)$, and the other being the classical convex hull of points $(0,1)$, $(0,2)$, $(0.5,2)$ and $(0.5,1.5)$.  
%{Show generators and constraints explicitly. They are named or dots on the picture of Example \ref{ex:running}: Figure \ref{fig:simplenet:trop}, \ref{fig:simpleneuralnet:trop2}, \ref{fig:simplenet:1subd} and \ref{fig:simplenet:2subd}.}
\end{example}

All tropical polyhedra can thus be described both internally and externally, and algorithms, although costly, can be used to translate an external description into an internal description and vice-versa. This is at the basis of the double description method for classical polyhedra \cite{Halbwachs} and for tropical polyhedra \cite{Tropical}. Double description is indeed useful when interpreting set-theoretic unions and intersections, as in validation by abstract interpretation, see \cite{Halbwachs} again for the classical case, and e.g. \cite{SAS2008} for the tropical case: unions are easier to compute using the extreme generator representation (the union of the convex hulls of sets of points is the convex hull of the union of these sets of points) while intersections are easier to compute using the external representation (the intersection of two polyhedra given by sets of constraints is given by the concatenation of these sets of constraints).

In the sequel, we will be using explicitly the max and (ordinary) + operators in place of $\oplus$ and $\otimes$ for readability purposes.

\subsection{From zone to tropical polyhedra and vice-versa}

\label{sec:zone:troppoly}

The following proposition characterizes the construction of tropical polyhedric abstractions from zones. We show that 
a zone defined on $n$ variables can be expressed as the tropical convex hull of $n+1$ points.

\begin{proposition}[Internal tropical representation of closed zones]
\label{prop:troprepofzones}

    Let $H_{ext}\subset \R^n$ be the $n$-dimensional zone defined by the conjunction of the $(n+1)^2$ inequalities
    $\bigwedge_{0\leq i,j \leq n}(x_i-x_j \leq c_{i,j}),$
    where $\forall i,j\in[0,n],\, c_{i,j} \in \R\cup\{+\infty\}$.
    Assume that this representation is \emph{closed}, then $H_{ext}$ is equal to the tropical polyhedron $H_{int}$ defined, with internal representation, as the tropical convex hull of the following extreme points (and no extreme ray): %\todo{Insister sur le fait qu'il n'y a pas de rayon}: 
%    $$H_{int} := \Bigg\{\mu A \oplus \bigoplus_{1\leq i \leq n} \lambda_i B_i \,\Bigg|\, \mu \oplus \bigoplus_{1\leq i \leq n} \lambda_i = 0\Bigg\},$$
 %   where
    \begin{align*}
        & A =(a_i)_{1\leq i\leq n}:= (-c_{0,1}, \dots, -c_{0,n}), \\
        & B_k = (b_{ki})_{1\leq i\leq n} := (c_{k,0} - c_{k,1}, \dots, c_{k, 0} - c_{k,n}), \ \mbox{ \ $k$=1, $\ldots$, $n$}
    \end{align*}
%        with %$a_i, b_i$ chosen optimally so that $a_i \leq x_i \leq b_i$ i.e.
%        $$a_i := -\delta_{0,i},\quad b_i := \delta_{i, 0}.$$
\end{proposition}
The proof is given in Appendix \ref{proof:troprepofzones}. 

\begin{example}
The zone of Example \ref{ex:running2} is the tropical polyedron with the three extreme generators $A$, $B_1$ and $B_2$ pictured in Figure \ref{fig:simplenet:trop}, as deduced from Proposition \ref{prop:troprepofzones} above. 
\end{example}

%We now prove that the closure hypothesis is always satisfied for zones defined by saturation, as is the case when abstracting a given non-empty set $S$:

Moreover, we can easily find the best zone (and also, hypercube) that outer approximates a given tropical polyhedron, as follows \cite{SAS2008}. Suppose we have $p$ extreme generators and rays for a tropical polyhedron $\mathcal{H}$, $A_1,\ldots, A_p$, that we put in homogeneous coordinates in $\R^{n+1}$ by adding as last component 0 to the coordinates of the extreme generators, and $-\infty$ to the last component, for extreme rays, as customary for identifying polyhedra with cones, see e.g. \cite{gauberkatz2006}. 
\begin{proposition}[\cite{SAS2008}]
%\label{prop:troptozone}
Let $A$ be the matrix of generators for tropical polyhedron $\mathcal{H}$ stripped out of rows consisting only of 
$-\infty$ entries, and $A/A$ the residuated matrix which entries are 
$(A/A)_{i,j}=\min\limits_{1 \leq k \leq p} a_{i,k}-a_{j,k}$.  Then the smallest zone containing $\mathcal{H}$ is given %\cite{SAS2008}
by the inequalities:
\begin{align*}
& x_i-x_j \geq (A/A)_{i,j} & \mbox{for all $i$,$j$=$1,\ldots,n$} \\
& (A/A)_{i,n+1} \leq x_i \leq -(A/A)_{n+1,i} & \mbox{for all $i=1,\ldots,n$}    \\
\end{align*}
\end{proposition}

%\todo{To be completed}

\begin{example}
Consider the graph of the ReLU function on $[-1,1]$, pictured in Figure \ref{fig:RELU:trop}. It has as generators the two extreme points $A_1=(-1,0)$ and $A_2=(1,1)$ (the graph is the tropical segment from $A_1$ to $A_2$). Homogenizing the coordinates and putting them in a matrix $A$ (columns correspond to generators), we have 
$$A=\left(\begin{array}{cc}
-1 & 1 \\
0 & 1 \\
0 & 0
\end{array}\right)
\mbox{ and }
(A/A) = \left(\begin{array}{ccc}
0 & -1 & -1\\
0 & 0 & 0 \\
-1 & -1 & 0
\end{array}\right)
$$ 
meaning that the enclosing zone is given by %\begin{align*}
$-1  \leq x-y \leq 0, \ -1 \leq x \leq 1, \ 
0 \leq y \leq 1$, 
%\end{align*}
%\noindent 
which is the zone depicted in Figure \ref{fig:RELU:zone}.
\end{example}

\subsection{Feedforward ReLU networks}

\label{sec:RELU}

Feedforward ReLU networks that we are considering in this paper are a succession of layers of neurons, input layer first, a given number of hidden layers and then an output layer, each computing a certain affine transform followed by the application of the ReLU activation function:

\begin{definition}
A $n$-neurons {ReLU network layer} $L$ with $m$ inputs is a function $\mathbb{R}^m\rightarrow\mathbb{R}^n$ defined by, 
%\begin{itemize}
%    \item 
    a weight matrix $W\in\mathcal{M}_{n,m}(\mathbb{R})$,  
%    \item 
    a bias vector $b\in\mathbb{R}^n$, and 
%    \item 
    an activation function $ReLU~:~\mathbb{R}^n\rightarrow\mathbb{R}^n$ given by $ReLU(x_1,\ldots,x_n)=(max(x_1,0),\ldots,max(x_n,0))$
%\end{itemize}
so that for a given input $x\in\mathbb{R}^n$, its output is $L(x) = ReLU(Wx + b)$
\end{definition}

%This way, the first layer %\todo{Pas tres heureux le fait d'avoir layer pour la fonction, et layer pour les neurones?}
%$L_0: \mathbb{R}^d\rightarrow\mathbb{R}^{n_1}$ of a network maps the system's state space to a first layer of neurons. With $L_i(x_i) = {}^t(l_{i,1},...,l_{i,n_1})$, the $l_{i,j}$s are real numbers corresponding to the activation value of the $j$-th neuron in the $i$-th layer. The last layer $L_N: \mathbb{R}^{n_{N}}\rightarrow\mathbb{R}^k$ maps the last layer of neurons to the output of the network. When we compose network layers, we get a MLP.

\begin{definition}
\label{def:MLP}
A multi-layer perceptron $F_N$ is given by a list of network layers $L_0,...,L_N$, where layers $L_{i}$ ($i=0,\ldots,N-1$) are $n_{i+1}$-neurons layers with $n_i$ inputs. the action of $F_N$ on inputs is defined by composing the action of successive layers: $F_N = L_N\circ...\circ{}L_0$
\end{definition}

With the above notations, there are $N$ hidden layers of neurons in the network. %With $n_0 = d$ and $n_{N+1} = k$, 
For each layer, we have $L_i~:~\mathbb{R}^{n_i}\rightarrow\mathbb{R}^{n_{i+1}}$ and we say that $F_N$ is a $n_1\times{}n_2\times...\times{}n_{N}$ network.

%In tropical algebra, ReLU activation functions are affine functions. This means that their action is exact in tropical polyhedra. On the contrary, affine functions in the classical sense are non linear in the tropical sense, and we are going to see how to abstract their action in tropical algebra. 

%\todo{Typo plus def a inferieur a b}

\section{Abstraction of linear maps}

\label{sec:1layer}
%\label{tropicalf}

\subsection{Zone-based abstraction}

\label{sec:zonebased}

%\subsection{Abstraction of linear maps}

%\label{sec:scalarlinear}

%\todo{Explain the principle: abstraction of the linear part, then ReLU}

We consider in this section the problem of abstracting the graph $\mathcal{G}_f=\{(x,y) \mid y=f(x)\}$ of a linear map 
$f(x)=W x + b$ with $x\in [\ux_1,\ox_1]\times \ldots [\ux_m,\ox_m]$ where $W=(w_{i,j})$ is a $n \times m$ matrix and $b$ a $n$-dimensional vector, by a tropical polyhedron $\mathcal{H}_f$. We will treat the case of multilayered networks in Section \ref{sec:multilayer}. 

The difficulty is that linear maps in the classical sense are not linear maps in the tropical sense, but are rather (generalized) tropical polynomials, hence the exact image of a tropical polyhedron by a (classical) linear map is not in general a tropical polyhedron. 
We begin by computing the best zone abstracting $\mathcal{G}_f$ and then represent it by a tropical polyhedron, using the results of Section \ref{sec:zone:troppoly}. 
We then show in Section \ref{sec:abstraction:octagon} that we can improve results using an octagon abstraction. 

%In order to discuss the tropical abstraction, and in particular, optimality, we need to describe the relationship between zones and tropical polyhedra. 

%\subsection{Zone and tropical polyhedric abstraction}
%\label{sec:zones_tropical_representation}

%\todo{Est-ce que ca ne serait pas plus logique de mettre la proposition ci-dessous  dans la Section 2, quand on introduit zones et polyedres tropicaux? (et peut-etre changer le titre de la section)}

%\todo{Expliquer ici un peu le cheminement logique, c'est toujours mieux ;-)}

The tightest zone containing the image of a cube going through a linear layer can be computed as follows: 

\begin{proposition}[Optimal approximation of a linear layer by a zone]
    \label{prop:optimal_linear_zone}

    Let $n,m\in\N$ and $f:\R^m\to\R^n$  an affine transformation defined, for all $x\in\R^m$ and $i\in[1,n]$, by
    $\big(f(x)\big)_i = \sum_{j=1}^m w_{i,j} x_j + b_i.$
Let $K\subset\R^m$ be an hypercube defined as $K=\prod_{1 \leq j \leq m} [\ux_j, \ox_j]$, with $\ux_j,\ox_j\in\R$.
    Then, the tightest zone $\mathcal{H}_f$ of $\R^m \times \R^n$ containing
    $S := \big\{\big(x,f(x)\big)\,\Big|\,x\in K \big\}$ 
    is the set of all $(x,y)\in\R^m\times\R^n$ satisfying
    \begin{align*}
     &\Big(\bigwedge_{1\leq j\leq m} \ux_j\leq x_j \leq \ox_j\Big)
      \wedge \Big(\bigwedge_{1\leq i\leq n} m_i\leq y_i \leq M_i\Big)
      \wedge \Big(\bigwedge_{1\leq i_1, i_2 \leq n} y_{i_1} - y_{i_2} \leq \Delta_{i_1,i_2}\Big) \\
       \wedge &\Big(\bigwedge_{1\leq i \leq n,1 \leq j \leq m} m_i - \ox_j + \delta_{i,j} \leq y_i - x_j   \leq M_i - \ux_j - \delta_{i,j} \Big),
    \end{align*}
    where, for all $i, i_1, i_2 \in [1,n]$ and $j \in [1,m]$:
    \begin{align*}
        m_i &= \sum_{w_{i,j}<0}w_{i,j}\ox_j + \sum_{w_{i,j}>0}w_{i,j}\ux_j + b_i, \\
        M_i &= \sum_{w_{i,j}<0}w_{i,j}\ux_j + \sum_{w_{i,j}>0}w_{i,j}\ox_j + b_i, \\
        \Delta_{i_1,i_2} &=  
                    \sum_{w_{i_1,j}<w_{i_2,j}} (w_{i_1,j}-w_{i_2,j})\ux_j
                  + \sum_{w_{i_1,j}>w_{i_2,j}} (w_{i_1,j}-w_{i_2,j})\ox_j + (b_{i_1} - b_{i_2}), \ \\
        \delta_{i,j} &=
        \begin{cases}
          0, & \text{if }\ w_{i,j} \leq 0 \\
          w_{i,j}(\ox_j-\ux_j), & \text{if }\ 0 \leq w_{i,j} \leq 1 \\
          (\ox_j-\ux_j), & \text{if }\ 1 \leq w_{i,j}
        \end{cases}
    \end{align*}
\end{proposition}

\begin{figure}
\begin{center}
\includegraphics[scale=0.35]{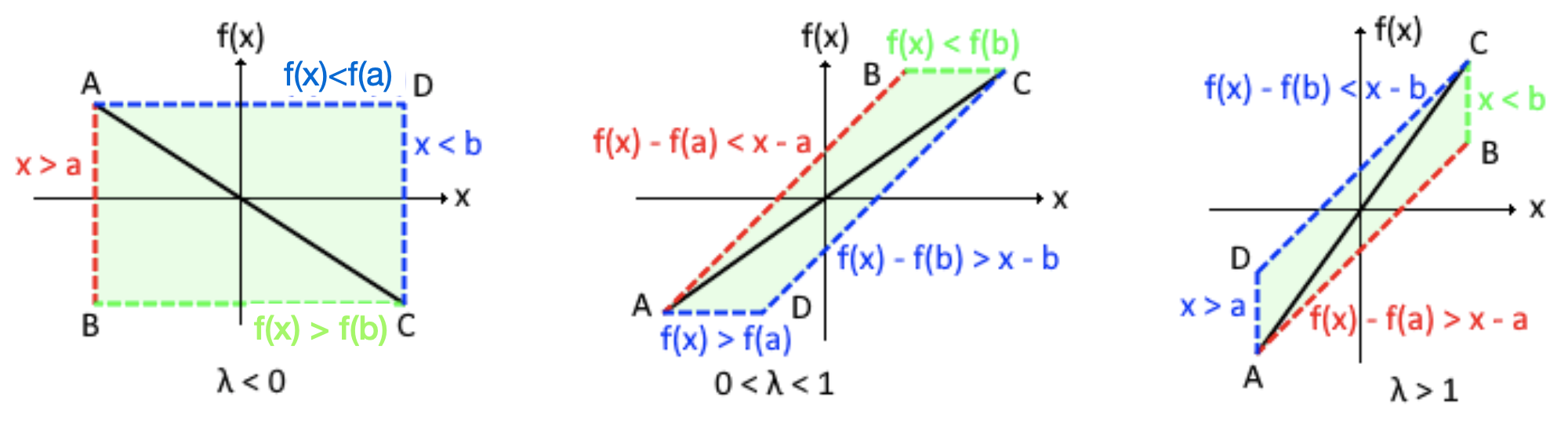}
\caption{The 3 cases for approximating the graph of an affine scalar function by a tropical polyhedron, on domain $[a,b]$.}
\label{fig1}
\end{center}
\end{figure}

%\begin{example}
%(1D)
%\todo{Montrer le calcul de l'enveloppe convexe de A, B, C etc.}
Figure \ref{fig1} shows the three different types of zones that over-approximate the range of a scalar function $f$, with $f(x)=\lambda x+b$, on an interval. When $\lambda < 0$, the best that can be done is to abstract the graph of $f$ by a square, we cannot encode any dependency between $f(x)$ and $x$: this corresponds to the case 
$\delta_{i,j}=0$ in Proposition \ref{prop:optimal_linear_zone}. The two other cases for the definition of $\delta_{i,j}$ are the two remaining cases of Figure \ref{fig1}: 
when $\lambda$ is between 0 and 1, this is the picture in the middle, and when $\lambda$ is greater than 1, this is the picture at the right hand side.  As we have seen in Proposition \ref{prop:troprepofzones} and as we will see more in detail below in Theorem \ref{thm:RmtoRn}, these zones can be encoded as tropical polyhedra.
Only the points $A$, $B$ and $C$ are extreme points: 
%It is an easy exercise to see that the points that are indicated in these pictures are extreme points. For instance, 
$D$ is not an extreme point of the polyhedron as it is on the tropical segment $[AC]$ (the blue, green and red dashed lines each represent a tropical segment).

%We can use these points to define the polyhedron internally, instead of externally with inequalities.

%\begin{center}
%\begin{tabular}{|r|c|c|c|}
%\hline
%& $A$ & $B$ & $C$ \\
%\hline
%If $\lambda \leq 0$ & & $(a, f(b))$ & \\
%If $0 \leq \lambda \leq 1$ & $(a, f(a))$ & $(a + f(b) - f(a), f(b))$ & $(b, f(b))$ \\
%If $\lambda \geq 1$ & & $(b, f(a) + b - a)$ & \\
%\hline
%\end{tabular}
%\end{center}

%\end{example}

\begin{figure}
\begin{center}
\includegraphics[scale=0.15]{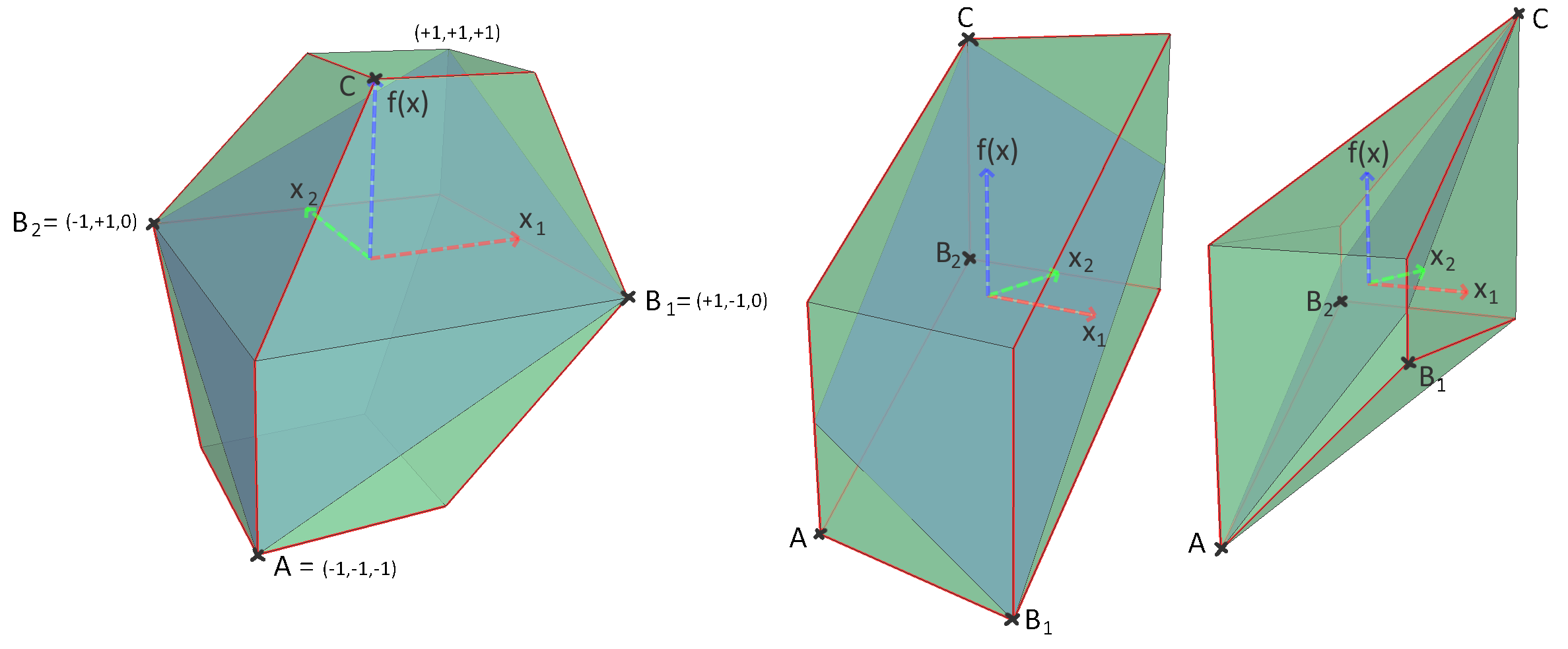}
\caption{Over-approximation for $\lambda_1 = \lambda_2 = 0.5$ (left), $\lambda_1 = -0.5$ and $\lambda_2 = 1.5$ (middle), and $\lambda_1 = \lambda_2 = 1.2$ (right).}
\label{fig4}
\end{center}
\end{figure}

%\begin{example}
%{Example: $f:\R^2\rightarrow\R$}
For $f:\R^2\rightarrow\R$, there are 6 cases, depending on the values of $\lambda_1$ and $\lambda_2$. In all cases, these zones can be represented as tropical polyhedra using only 4 extreme points and 4 inequalities (instead of 8 and 6 in the classical case), as we will see in Theorem \ref{thm:RmtoRn}. 
Fig. %\ref{fig2}, \ref{fig3}, 
\ref{fig4} represents the resulting polyhedron for different values of $\lambda_1$ and $\lambda_2$. 
%The inequalities and the corresponding faces on the cube are also shown. 
%Note that we only need 4 extreme points and inequalities to define a cube as in Figure \ref{fig4}, instead of 8 and 6 in the classical case.
Each figure shows the extreme points $A$, $B_1$, $B_2$ and $C$, the faces of the polyhedron (in green), the tropical segments inside the polyhedron (in red), and the actual graph of $f(x)$ (in blue).
%\begin{figure}
%\begin{minipage}{6.5cm}
%\begin{center}
%\includegraphics[scale=0.15]{R2-05-05.png}
%\caption{Over-approximation for $\lambda_1 = \lambda_2 = -0.5$. }
%\label{fig2}
%\end{center}
%\end{minipage}
%\begin{minipage}{6.5cm}
%\begin{center}
%\includegraphics[scale=0.16]{R2-05+05.png}
%\caption{Over-approximation for $\lambda_1 = -0.5$ and $\lambda_2 = 0.5$.}
%%\label{fig3}
%\end{center}
%\end{minipage}
%\end{figure}
We have the corresponding external description in Theorem \ref{thm:RmtoRn} below, which proof is given in Appendix \ref{proof:RmtoRn}.
\begin{theorem}
\label{thm:RmtoRn}
The best zone abstraction $\mathcal{H}_f$ of of the graph $\mathcal{G}_f=\{(x_1,\ldots,x_m,$ $y_1,\ldots,y_n) \mid \ux_j \leq x_j \leq \ox_j, \ y_i=f_i(x_1,\ldots,x_m) \} \subseteq \R^{+n}$ of the linear function $f: \R^m \rightarrow \R^n$ defined in Proposition \ref{prop:optimal_linear_zone} 
%$\mathcal{H}_f$ is a sound\todo{Plutot zone optimale representee par un polyedre tropical} abstraction of the graph $\mathcal{G}_f=\{(x_1,\ldots,x_m,y_1,\ldots,y_n) \mid \ux_j \leq x_j \leq \ox_j, \ y_i=f_i(x_1,\ldots,x_m) \} \subseteq \R^{+n}$ of the linear function $f: \R^m \rightarrow \R^n$. It 
can be seen as the tropical polyhedron defined externally with $m + n + 1$ inequalities, for all $i \in [1,n]$ and $j \in [1,m]$:

%& xternal description & \\ \\
%& $\forall i \in [1,m]$ & $\forall j \in [1,n]$ \\
%\begin{multline}
\begin{eqnarray}
\hspace*{-0.3cm} \max (
x_1 - \ox_1,\ldots,x_m - \ox_m,y_1 - M_1,\ldots,y_n - M_n
)
& \leq & 0 \label{eq:RmtoRn1} %\end{equation}
\\
%\end{multline}
%\begin{multline} %\\
%\begin{equation}
\max (
0, y_1 - M_1 + \delta_{1,j},\ldots,y_n - M_n + \delta_{n,j}
)  & \leq & x_j - \ux_j \label{eq:RmtoRn2} \\ %\end{equation}
%\end{eqnarray}
%\begin{equation}
%\begin{multline}
\hspace*{-0.3cm} \max( 
0, x_1 - \ox_1 + \delta_{i,1},\ldots,x_n - \ox_n + \delta_{i,n},y_1 - d_{i,1},\ldots,y_n - d_{i,n}
) & \leq & y_i - m_i
\label{eq:RmtoRn3}
%\end{multline}
\end{eqnarray}
where $d_{j_1,j_2}$ denotes
    the quantity $\Delta_{j_1,j_2} + m_{j_2}$ for $i_1$ and $i_2$ in $[1,n]$.
%\todo{A verifier qu'il n'y a pas le meme souci de non symetrie du deltaij entre les 2 cotes de l'inegalite. EG: c'est bon (c'est la version d'origine de Sebastien, sans la typo de Louis}
%\begin{tabular}{l l}
%& External description & \\ \\
%& $\forall i \in [1,m]$ & $\forall j \in [1,n]$ \\
%\begin{equation}
% \max \mbox{\begin{pmatrix}
%x_1 - b_1 \\ \vdots \\ x_m - b_m \\ y_1 - M_1 \\ \vdots \\ y_n - M_n
%\end{pmatrix}} \leq 0 \label{eq:RmtoRn1}\end{equation}
%&\begin{equation}
%\max \begin{pmatrix}
%0 \\ y_1 - M_1 + \delta_{i,1} \\ \vdots \\ y_n - M_n + \delta_{i,n}
%\end{pmatrix} \leq x_i - a_i \label{eq:RmtoRn2} \mbox{ } \end{equation}
%\end{tabular}\\
%\begin{equation}\max \begin{pmatrix}
%0 \\ x_1 - b_1 + \delta_{1,j} \\ \vdots \\ x_n - b_n + \delta_{n,j} \\ y_1 - d_{1,j} \\ \vdots \\ y_n - d_{n,j}
%\end{pmatrix} \leq y_j - m_j
%\label{eq:RmtoRn3}
%\end{equation}
%\end{center}
\end{theorem}    
We have the matching internal representation in Theorem \ref{thm:intRmtoRn}, which proof is given in Appendix \ref{sec:proof:thm:intRmtoRn}.
\begin{theorem}
\label{thm:intRmtoRn}
$\mathcal{H}_f$ can also be described, internally, as the tropical convex hull of $m + n + 1$ extreme points: 
%& Internal description \\ \\
\begin{eqnarray*}
A & = & (\ux_1, \ldots, \ux_m, m_1, \ldots, m_n) \\
B_1 & = & (
\ox_1, \ux_2, \ldots, \ux_m, m_1 + \delta_{1,1}, \ldots, m_n + \delta_{n,1}) \ldots \\
%\hdots  \\
B_m & = & (
\ux_1, \ldots, \ux_{m-1}, \ox_m, m_1 + \delta_{1,m}, \ldots, m_n + \delta_{n,m}) \\
C_1 & = & (
\ux_1 + \delta_{1,1}, \ldots, \ux_m + \delta_{1,m}, M_1, c_{1,2}, \ldots, c_{1,n}) \hdots \\
%\hdots \\
C_n & = & (
\ux_1 + \delta_{n,1}, \ldots, \ux_m + \delta_{n,m}, c_{n,1}, \ldots, c_{n,n-1}, M_n)
\end{eqnarray*}

%\begin{center}
%\begin{tabular}{c c c c}
%%& Internal description \\ \\
%$A = \begin{pmatrix}
%a_1 \\ \vdots \\ a_m \\ m_1 \\ \vdots \\ m_n
%\end{pmatrix}$ &
%$B_1 = \begin{pmatrix}
%b_1 \\ a_2 \\ \vdots \\ a_m \\ m_1 + \delta_{1,1} \\ \vdots \\ m_n + \delta_{1,n}
%\end{pmatrix}$ &
%$\hdots$ &
%$B_m = \begin{pmatrix}
%a_1 \\ \vdots \\ a_{m-1} \\ b_m \\ m_1 + \delta_{m,1} \\ \vdots \\ m_n + \delta_{m,n}
%\end{pmatrix}$ 
%\end{tabular}
%\begin{tabular}{c c c }
%$C_1 = \begin{pmatrix}
%a_1 + \delta_{1,1} \\ \vdots \\ a_m + \delta_{m,1} \\ M_1 \\ c_{1,2} \\ \vdots \\ c_{1,n}
%\end{pmatrix}$ &
%$\hdots$ &
%$C_n = \begin{pmatrix}
%a_1 + \delta_{1,n} \\ \vdots \\ a_m + \delta_{m,n} \\ c_{n,1} \\ \vdots \\ c_{n,n-1} \\ M_n
%\end{pmatrix}$
%\end{tabular}
%\end{center}
%\noindent where $c_{i_1,i_2} = M_{i_1} - d_{i_1,i_2} + m_{i_2}$ for $i_1$ and $i_2$ in $[1,n]$. 
\noindent where $c_{i_1,i_2} = M_{i_1} - {\Delta}_{i_1,i_2}$ for $i_1$ and $i_2$ in $[1,n]$. 
\end{theorem}

%    Keeping the notations of proposition \ref{prop:optimal_linear_zone}, we have: 
    
%\begin{proposition}[Tropical approximation of a linear layer]
%    $S$ is contained in the tropical convex polyhedron generated by the $1+n+m$ points $A, B_1, \dots, B_n, C_1, \dots, C_m$, where \\
    
%        $A\, = (0, a_1, \dots, a_n, m_1, \dots, m_m)$
%    \todo{A verifier, j'ai change ses "*" en j ou k}
%    \begin{align*}
%        B_j &= (0, x, y) \text{ with } x_j = b_j,& & x_{i} = a_i, \mbox{ $i \neq j$} & & y_k =  m_j + \delta_{j,k}\\
%        C_j &= (0, x, y) \text{ with } y_j = M_j, & & y_{k} = M_j - \Delta_{j,k} \mbox{ $k \neq j$} & & x_i = a_i + \delta_{i, j}
%    \end{align*}
    
%\end{proposition}

%\todo{Les exemples sont a retravailler un peu, c'est un peu sec comme ca! (ca a ete repris de l'autre version ;-)}

\begin{example}[Running example]
\label{ex:runningzone}
Let us detail the computations for Example \ref{ex:running}: 
$h_1=x_1-x_2-1$, $h_2=x_1+x_2+1$. 
%$y_1=max(0,x_1-x_2-1)$ and $y_2=max(0,x_1+x_2+1)$.
We have respectively, $\delta_{1,1}=2$, $\delta_{1,2}=0$, $\delta_{2,1}=2$, $\delta_{2,2}=2$, $\Delta_{1,1}=0$, $\Delta_{1,2}=0$, $\Delta_{2,1}=4$,  $\Delta_{2,2}=0$, $d_{1,1}=-3$, $d_{1,2}=-1$, $d_{2,1}=1$, $d_{2,2}=-1$, $m_1=-3$, $m_2=-1$, $M_1=1$ and $M_2=3$. Hence the external description for the tropical polyhedron relating values of $x_1$, $x_2$, $h_1$ and $h_2$ are:
%\begin{eqnarray}
$ \max (
x_1 -1,x_2 -1,h_1 - 1,h_2 - 3
)
\leq  0 \label{ex:running:ineq1}, 
\
%\begin{equation}
\max (
0, h_1+1,h_2 -1
)  \leq x_1 +1 \label{ex:running:ineq2}, \
\max (
0, h_1 - 1,h_2 - 1
)  \leq x_2 +1 \label{ex:running:ineq3}, \
%\end{equation}
%\begin{equation}
\max( 
0, x_1 +1,x_2 - 1, h_1 +3,h_2-1
)  \leq  h_1 +3 \label{ex:running:ineq4}, \
\max( 
0, x_1 +1,x_2 +1,h_1 +1,h_2 +1
) \leq h_2 +1 \label{ex:running:ineq5}$
%\end{eqnarray}
%\noindent 
which encode all zones inequalities:
$ %$\begin{array}{rcl}
-1\leq  x_1  \leq 1, \
-1\leq  x_2  \leq 1, \
-3 \leq  h_1  \leq 1, \
-1 \leq  h_2  \leq 3, \
-2\leq  h_1-x_1  \leq 0, \
-4 \leq  h_1-x_2  \leq 2, \
0 \leq  h_2-x_1  \leq 2, \
0 \leq  h_2-x_2  \leq 2, \
-4 \leq  h_1-h_2  \leq 0
$. Note that the zone abstraction of \cite{Mine2} would be equivalent to an interval abstraction and would not infer the relations between $h_1$, $h_2$, $x_1$ and $x_2$. 
%\end{array}$$ %\todo{to be checked}
%The first three tropical inequalities (\ref{ex:running:ineq1}), (\ref{ex:running:ineq2}) and (\ref{ex:running:ineq3} encode the zone for $x_1$, $x_2$, $h_1$ and $h_2$. 
%\todo{Make the explicit computation and relate to Figures \ref{fig:simplenet:zone}, \ref{fig:simplenetinnerrange}.}
Now the internal representation of the corresponding zone is %\todo{To be completed}:
%\begin{eqnarray*}
$A  =  (-1, -1, -3, -1), \
B_1  =  (
1, -1, -1, 1), \
B_2  =  (-1, 1, -3, 1), \
C_1  = 
(-1, -1, 1, 1), \
C_2  =  (
-1, 1, -1, 3)$. 
%\end{eqnarray*}
The projections of these 5 extreme points on $(h_1,h_2)$ give
the points $(-3,-1)$, $(-1,1)$, $(-3,1)$, $(1,1)$, $(-1,3)$, among which $(-3,1)$ and $(-1,1)$ are in the tropical convex hull of $A=(-3,-1)$, $B_1=(1,1)$ and $B_2=(-1,3)$ represented in Figure \ref{fig:simplenet:trop}. Indeed $(-3,1)$ is on the tropical line $(AB_2)$ and $(-1,1)$ whereas $(-1,1)$ is on the tropical line $(AB_1)$ as a tropical linear combination of $-2+B_1$ and $-2+B_2$: 
$(-1,1)=max(-2+(1,1),-2+(-1,3))$. 
%\todo{Add what \cite{Mine2} would give on such "assignments"?} 
%$\delta_{1,1}=2$, $\delta_{1,2}=0$, $\delta_{2,1}=2$, $\delta_{2,2}=2$, $\Delta_{1,1}=0$, $\Delta_{1,2}=0$, $\Delta_{2,1}=4$,  $\Delta_{2,2}=0$, $d_{1,1}=-3$, $d_{1,2}=-1$, $d_{2,1}=1$, $d_{2,2}=-1$, $m_1=-3$, $m_2=-1$, $M_1=1$, $c_{i_1,i_2} = M_{i_1} - d_{i_1,i_2}
%+ m_{i_2}$ for $i_1$ and $i_2$ in $[1,n]$, $c_{1,1}=1+3-3=1$, $c_{1,2}=1+1-1=1$, $c_{2,1}=1-1-3=-3$, $c_{2,2}=3+1-1=3$
\end{example}

\begin{example}
\label{ex:RmtoRn1}
%In order to illustrate the constructions of this section, we will consider 
Consider now function $f: \R^2 \rightarrow \R^2$ with $f(x_1,x_2) = (0.9 x_1 + 1.1 x_2, y_2 = 1.1 x_1 - 0.9 x_2)$ on $(x_1,x_2)\in [-1,1]$. 
We have in particular $M_1=2$, $M_2=2$, $m_1=-2$ and
$m_2=-2$. We compute $\delta_{1,1}=1.8$, $\delta_{1,2}=2$, $\delta_{2,1}=2$ and $\delta_{2,2}=0$ and we have indeed $y_1+2 \geq x_1-1+\delta_{1,1}=x_1+0.8$, 
$y_2 + 2\geq x_1-1+2=x_1+1$, 
$y_1+2 \geq x_2-1+\delta_{2,1}=x_1+1$, $y_2 + 2\geq x_2-1$ 
and $y_1-2\leq x_1+1-1.8=x_1-0.8$, 
$y_2-2\leq x_1+1-2=x_1-1$, 
$y_1-2\leq x_2+1-2=x_2-1$, $y_2-2\leq x_2+1$. Overall: 
$$\begin{array}{rrcll}
x_1 -1.2& \leq & y_1 & \leq & x_1+1.2 \\
x_2-1 & \leq & y_1 & \leq & x_2+1 \\
x_1-1 & \leq & y_2 & \leq & x_1+1 \\
x_2-3 & \leq & y_2 & \leq & x_2+3 \\
\end{array}$$
We also find
$d_{1,1}=-2$, $d_{1,2}=0.2$, $d_{2,1}=0.2$ and $d_{2,2}=-2$. Hence 
$y_1-d_{1,2} \leq y_2-m_2$, i.e. $y_1-0.2\leq y_2+2$ that is $y_1-y_2\leq 2.2$. Similarly, we find
$y_2-y_1 \leq 2+0.2$ hence $-2.2 \leq y_1-y_2 \leq 2.2$. 
%We get back to Example \ref{ex:RmtoRn}. 

The equations we found 
can be written as the following linear tropical constraints as in Theorem \ref{thm:RmtoRn}:
$$\begin{array}{c c c}
max \left(\begin{array}{c}
x_1-1 \\ x_2 -1 \\ y_1-2 \\ y_2 -2 
\end{array}\right) \leq 0 & 
max \left(\begin{array}{c}
0 \\
y_1-0.2 \\
y_2
\end{array}\right) \leq x_1+1 & 
max \left(\begin{array}{c}
0 \\
y_1 \\
y_2-2
\end{array}\right) \leq x_2+1 
\end{array}
$$
$$
\begin{array}{c c}
max \left(\begin{array}{c}
0 \\
x_1+0.8 \\
x_2+1 \\
\color{gray}y_1+2 \\
y_2-0.2
\end{array}\right) \leq y_1+2 & 
max \left(\begin{array}{c}
0 \\
x_1+1 \\
x_2-1 \\
y_1-0.2 \\
\color{gray}y_2+2
\end{array}\right) \leq y_2+2 \\
\end{array}$$
We now depict both the image of $f$  as a blue rotated central square, and its over-approximation by the convex tropical polyhedron calculated as in Theorem \ref{thm:RmtoRn} in green, in the plane $(y_1,y_2)$.
%We depict the image of $f$ on $[-1, 1]\times [-1, 1]$ 
%in Figure \ref{fig5}.
\begin{figure}
\begin{center}
\includegraphics[scale=0.25]{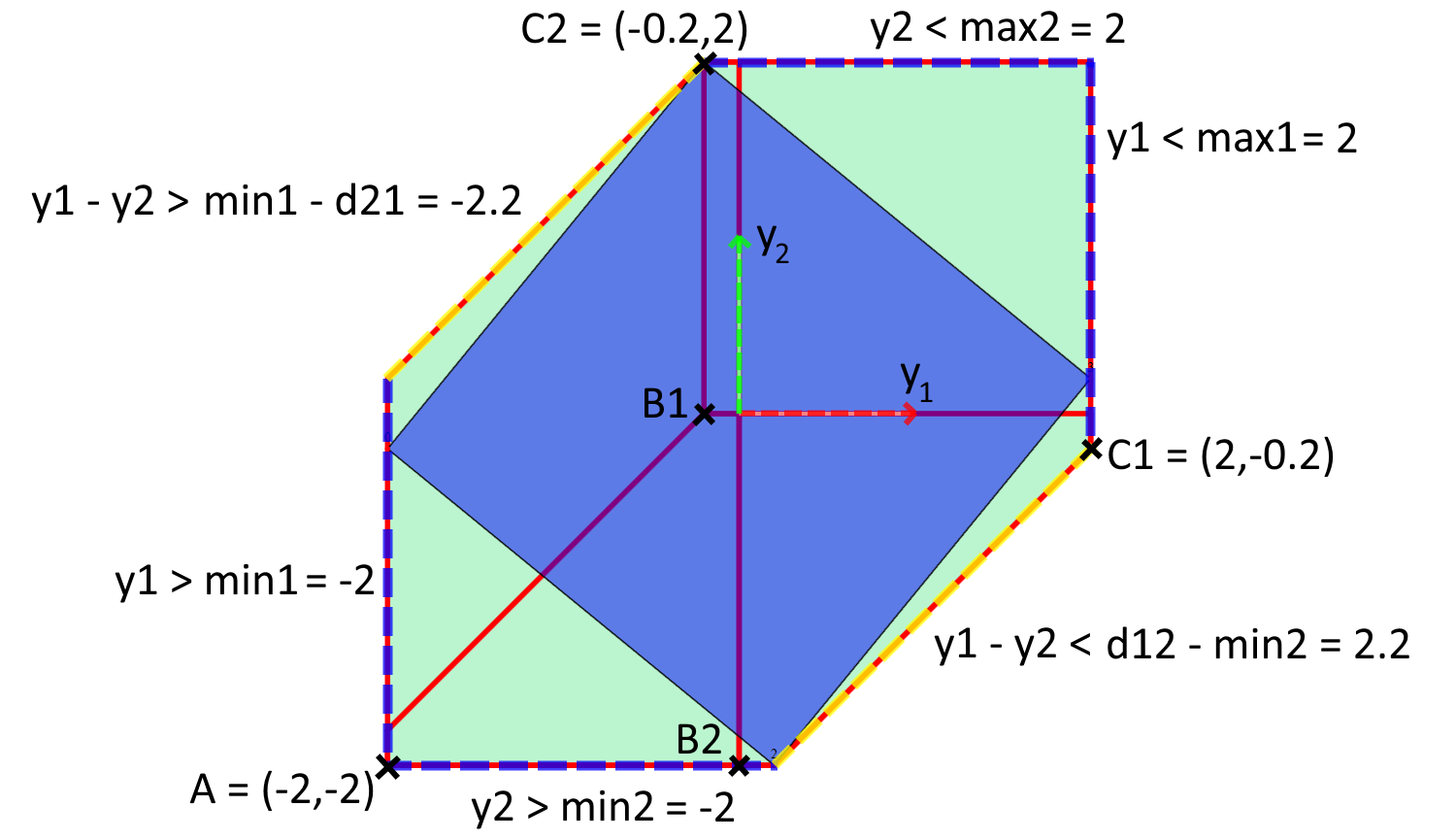}
\caption{Over-approximation for $(y_1,y_2)=f(x_1,x_2) = (0.9 x_1 + 1.1 x_2, y_2 = 1.1 x_1 - 0.9 x_2)$.}
\label{fig5}
\end{center}
\end{figure} %\todo{Put the right equations on the figure. - C'est fait - Sébastien}
As $c_{1,1}=2$, $c_{1,2}=-0.2$, $c_{2,1}=-0.2$ and $c_{2,2}=2$, the extremal points are, in the $(x_1,x_2,y_1,y_2)$ coordinates: 
\begin{center}
\begin{tabular}{c c c}
$A = \begin{pmatrix}
-1 \\ -1 \\ -2 \\ -2
\end{pmatrix}$ &
$B_1 = \begin{pmatrix}
1 \\ -1 \\ -0.2 \\ 0
\end{pmatrix}$ &
$B_2 = \begin{pmatrix}
-1 \\ 1 \\ 0 \\ -2
\end{pmatrix}$ 
\end{tabular} 
\begin{tabular}{c c }
$C_1 = \begin{pmatrix}
0.8 \\ 1 \\ 2 \\ -0.2
\end{pmatrix}$ &
$C_2 = \begin{pmatrix}
1 \\ -1 \\ -0.2 \\ 2
\end{pmatrix}$
\end{tabular}
\end{center}
% Be careful, the projections of the extremal points are not extremal points of the projection
\end{example}

%\todo{EG: je me suis arrete la 12/12/2020}

%HOPHOP

\subsection{Octagon abstractions and $(\max, +, -)$ algebra}

\label{sec:abstraction:octagon}

%\todo{Utiliser kReLU6 avec l'intersection de 2 zones pour un octagones dans notre running example}

%\todo{Pas tres joli comme titre}

%[TODO: 1. Explanation/definition of the domain through abstraction / concretization]

%\todo{Expliquer un peu plus ce qu'on fait la - mais c'est lie a la section 2 ou on introduit les octagones a partir des zones}

 As in Section \ref{sec:zonebased}, we consider the abstraction of the image of an hypercube $K$ of $\R^m$ by  an affine transformation $f:\R^m\to\R^n$  defined, for all $x\in\R^m$ and $i\in[1,n]$, by
    $\big(f(x)\big)_i = \sum_{j=1}^m w_{i,j} x_j + b_i$. 
But we consider here the abstraction of this image by an octagon, we will thus add some constraints on sums of variables to the abstraction computed in Section \ref{sec:zonebased}. 
    
\begin{proposition}[Optimal approximation of a linear layer by an octagon]
%\todo{Corriger la typo sur les signes sur delta et gamma}
    \label{prop:optimal_linear_octagon}
    Let $K\subset\R^m$ be an hypercube defined as $K=\prod_j [\ux_j, \ox_j]$, with $\ux_j,\ox_j\in\R$.
    The tightest octagon of $\R^m \times \R^n$ containing
    $$S := \Big\{\big(x,f(x)\big)\,\Big|\,x\in K\Big\}$$
    is the set of all $(x,y)\in\R^m\times\R^n$ satisfying
    \begin{align*}
     &\Bigg(\bigwedge_{1\leq j\leq m} \ux_j\leq x_j \leq \ox_j\Bigg)
      \wedge \Bigg(\bigwedge_{1\leq i\leq n} m_i\leq y_i \leq M_i\Bigg)
      \wedge \Bigg(\bigwedge_{1\leq i_1, i_2 \leq m} y_{i_1} - y_{i_2} \leq \Delta_{i_1,i_2}\Bigg) \\
      \wedge & \Bigg(\bigwedge_{1\leq i_1, i_2 \leq n} L_{i_1,i_2} \leq y_{i_1} + y_{i_2} \leq \Gamma_{i_1,i_2}\Bigg) \\
      \wedge &\Bigg(\bigwedge_{1\leq i \leq n,1 \leq j \leq m} m_i - \ox_j + \delta_{i,j} \leq y_i - x_j   \leq M_i - \ux_j - \delta_{i,j} \Bigg)\\
  %    \wedge &\Bigg(\bigwedge_{i,j} m_i - \ox_j \leq y_i - x_j + \delta_{i,j} \leq M_i - \ux_j\Bigg)\\
      \wedge &\Bigg(\bigwedge_{i,j} m_i + \ux_j + \gamma_{i,j} \leq y_i + x_j  \leq M_i + \ox_j - \gamma_{i,j} \Bigg)\\
    \end{align*}
    where $m_i, M_i, \delta_{i,j}, \Delta_{i_1,i_2}$ are defined as in Proposition \ref{prop:optimal_linear_zone}, and
    \begin{align*}
        \Gamma_{i_1,i_2} &:=  
                    \sum_{w_{i_1,j}+w_{i_2,j}<0} \ux_j(w_{i_1,j}+w_{i_2,j})
                  + \sum_{w_{i_1,j}+w_{i_2,j}>0} \ox_j(w_{i_1,j}+w_{i_2,j})\\
        L_{i_1,i_2} &:=  
                    \sum_{w_{i_1,j}+w_{i_2,j}<0} \ox_j(w_{i_1,j}+w_{i_2,j})
                  + \sum_{w_{i_1,j}+w_{i_2,j}>0} \ux_j(w_{i_1,j}+w_{i_2,j})\\
        \gamma_{i,j} &:=
        \begin{cases}
          0, & \text{if}\ 0 \leq w_{i,j}\\
          -w_{i,j}(\ox_j-\ux_j), & \text{if}\ -1 \leq w_{i,j} \leq 0 \\
          (\ox_j-\ux_j), & \text{if}\ w_{i,j} \leq -1
        \end{cases}
    \end{align*}
%    Those new quantities enjoy the following properties.
%    \begin{align*}
%        \gamma_{i,j} &= \min_{(x,y)\in S}\big((y_i-m_i)+(x_j-\ux_j)\big) \\
%                     &= - \max_{(x,y)\in S}\big((y_i-M_i)+(x_j-\ox_j)\big) \\
%        \Gamma_{i_1,i_2} &= \max_{(x,y)\in S} y_{i_1} + y_{i_2} \\
%        L_{i_1,i_2} &= \min_{(x,y)\in S} y_{i_1} + y_{i_2} \\
%    \end{align*}
\end{proposition}

The proof is given in Appendix \ref{proof:optimal_linear_octagon}.

With the notations of Proposition \ref{prop:optimal_linear_octagon}, we have
\begin{proposition}
\label{prop:tropoctagon}
Let $M$ be the (classically) linear manifold in $\R^m\times\R^n\times\R^m\times\R^n$ defined by $(x^+,y^+,x^-,y^-)\in M$ if and only if $x^++x^-=0$ and $y^++y^-=0$. 
    The octagon $S$ defined in Proposition \ref{prop:optimal_linear_octagon} %\todo{Plutot  Prop \ref{prop:octagon_to_zone}} 
    is equal to the intersection of $M$ with 
    %(the concretization of) 
    the tropical convex polyhedron generated by the $1+2n+2m$ points $A, B^+_1, \dots, B^+_m, B^-_1, \dots, B^-_m, C^+_1, \dots, C^+_n, C^-_1, \dots, C^-_n$, where
\[   A\, = (\ux_1, \dots, \ux_m, m_1, \dots, m_n, 
                  -\ox_1, \dots, -\ox_m, -M_1, \dots, -M_n)\]
    \begin{align*}
        B^+_k &= (0, x^+, y^+, x^-, y^-) \text{ with} 
              && x^+_k = \ox_k,
              && x^+_{j\neq k} = \ux_j,
              && y^+_i =  m_i + \delta_{i,k}\\
              & & & x_k^- = -\ox_k, 
              & & x^-_{j\neq k} = -\ox_j,
              & & y^-_i = -M_i + \gamma_{i,k}\\
        B^-_k &= (0, x^+, y^+, x^-, y^-) \text{ with} 
              & & x^-_k = -\ux_k,
              & & x^-_{j\neq k} = -\ox_j,
              & & y^-_i = -M_i + \delta_{i,k}\\
              & & & x_k^+ = \ux_k, 
              & & x^+_{j\neq k} = \ux_j,
              & & y^+_i = m_i + \gamma_{i,k}\\
        C^+_l &= (0, x^+, y^+, x^-, y^-) \text{ with} 
              & & y^+_l = M_l,
              & & y^+_{i\neq l} = M_l - \Delta_{l,i},
              & & x^+_j = \ux_j + \delta_{l, j}\\
              & & & y_l^- = -M_l, 
              & & y^-_{i\neq l} = M_l - \Gamma_{l,i},
              & & x^-_j = -\ox_j + \gamma_{l,j}\\
        C^-_l &= (0, x^+, y^+, x^-, y^-) \text{ with} 
              & & y^-_l = -m_l,
              & & y^-_{i\neq l} = -m_l - \Delta_{i,l},
              & & x^-_j =  -\ox_j + \delta_{l,j}\\
              & & & y_l^+ = m_l, 
              & & y^+_{i\neq l} = -m_l + L_{l,i},
              & & x^+_j = \ux_j + \gamma_{l,j}
    \end{align*}
\end{proposition}

The proof is given in Appendix \ref{proof:tropoctagon}. 

\begin{example}[Running example]
%\todo{Complete}
For the example network of Example \ref{ex:running}, the formulas of Proposition \ref{prop:optimal_linear_octagon} give the following constraints: 
\begin{align*}	
& -1 \leq x_1 \leq 1 \\
&	0 \leq x_1 - h_1 \leq 2 \\
&	-4 \leq x_1 + h_1 \leq 2 \\
&	-2 \leq x_1 - h_2 \leq 0 \\
&	-2 \leq x_1 + h_2 \leq 4 \\
%&	-2 \leq x_2 - x_1 \leq 2 \\
%&	-2 \leq x_2 + x_1 \leq 2 \\
&	-1 \leq x_2 \leq 1 \\
&	-2 \leq x_2 - h_1 \leq 4 \\
&	-2 \leq x_2 + h_1 \leq 0 \\
&	-2 \leq x_2 - h_2 \leq 0 \\
&	-2 \leq x_2 + h_2 \leq 4 \\
%&	0 \leq 1 - x_1 \leq 2 \\
%&	0 \leq 1 + x_1 \leq 2 \\
%&	0 \leq 1 - x_2 \leq 2 \\
%&	0 \leq 1 + x_2 \leq 2 \\
%&	1 \leq 1 \leq 1 \\ 
%&	0 \leq 1 - y_1 \leq 4 \\
%&	-2 \leq 1 + y_1 \leq 2 \\
%&	-2 \leq 1 - y_2 \leq 2 \\
%&	0 \leq 1 + y_2 \leq 4 \\
&	-3 \leq h_1 \leq 1 \\
&	0 \leq h_2 - h_1 \leq 4 \\
&	-2 \leq h_2 + h_1 \leq 2 \\
&	-1 \leq h_2 \leq 3 \\
\end{align*}
%\todo{On est s\^ur de ce $h_2-h_1\leq 4$ pas plutot 3? OK c'est bon, c'est 3 apres RELU}
And the internal description is given by Proposition \ref{prop:tropoctagon}, with the following extreme points, where coordinates are ordered as $(x_1^+, x_2^+, h_1^+, h_2^+, x_1^-, x_2^-, h_1^-, h_2^-)$:
\begin{align*}
&	(-1, -1, -3, -1, -1, -1, -1, -3) \\
&	(1, -1, -1, 1, -1, -1, -1, -3) \\
&	(-1, 1, -3, 1, -1, -1, 1, -3) \\
&	(-1, -1, -3, -1, -1, -1, -1, -3) \\ 
&	(1, -1, 1, 1, -1, 1, -1, -1)  \\
&	(1, 1, -1, 3, -1, -1, 1, -3) \\
&	(-1, -1, -3, -1, 1, -1, 1, -1) \\
&	(-1, -1, -1, -1, -1, 1, -1, -1) \\
&	(-1, -1, -3, -1, -1, -1, -1, -3) \\
&	(-1, 1, -3, 1, 1, -1, 3, -1) \\
&	(-1, -1, -1, -1, 1, 1, 1, 1) \\
	\end{align*}
%	\begin{example}
%\todo{To be done : RELU on this below}
%\begin{align}	
%%& -1 \leq x_1 \leq 1 \\
%&	0 \leq x_1 - h_1 \leq 2 \\
%&	-4 \leq x_1 + h_1 \leq 2 \\
%&	-2 \leq x_1 - h_2 \leq 0 \\
%&	-2 \leq x_1 + h_2 \leq 4 \\
%&	-2 \leq x_2 - x_1 \leq 2 \\
%&	-2 \leq x_2 + x_1 \leq 2 \\
%&	-1 \leq x_2 \leq 1 \\
%&	-2 \leq x_2 - h_1 \leq 4 \\
%&	-2 \leq x_2 + h_1 \leq 0 \\
%&	-2 \leq x_2 - h_2 \leq 0 \\
%&	-2 \leq x_2 + h_2 \leq 4 \\
%&	0 \leq 1 - x_1 \leq 2 \\
%&	0 \leq 1 + x_1 \leq 2 \\
%&	0 \leq 1 - x_2 \leq 2 \\
%&	0 \leq 1 + x_2 \leq 2 \\
%&	1 \leq 1 \leq 1 \\ 
%&	0 \leq 1 - y_1 \leq 4 \\
%&	-2 \leq 1 + y_1 \leq 2 \\
%&	-2 \leq 1 - y_2 \leq 2 \\
%&	0 \leq 1 + y_2 \leq 4 \\
%&	-3 \leq h_1 \leq 1 \\
%&	0 \leq h_2 - h_1 \leq 4 \\
%&	-2 \leq h_2 + h_1 \leq 2 \\
%&	-1 \leq h_2 \leq 3 \\
%\end{align}
%\todo{On est s\^ur de ce $h_2-h_1\leq 4$ pas plutot 3? OK c'est bon, c'est 3 apres RELU}
From the extremal points for the octagon abstraction above, we get the extremal points for $(h_1^+,h_2^+)$, discarding the non extremal ones: 
%he internal description is given by the following extreme points (coordinates are ordered as $(x_1+, x_2+, h_1+, h_2+, x_1-, x_2-, h_1-, h_2-)$):
%\begin{align*}
$(-3, -1)$, $(1, 1)$ and $(-1, 3)$. 
	and for $(h_1^+,h_2^-)$: $(-3,-3)$, $(1,-1)$ and $(-1,1)$ giving the zone in cyan of Figure \ref{ex:running3}. 
%	\begin{align*}
%&	(-1, -3) \\
%&	(-1, -3) \\
%&	(1, -3) \\
%&	(-1, -3) \\ 
%&	(-1, -1)  \\
%%&	(1, -3) \\
%&	(1, -1) \\
%&	(-1, -1) \\
%&	(-1, -3) \\
%&	(3, -1) \\
%&	(1, 1) \\%
%	\end{align*}
%\end{example}
\end{example}

\section{Validation of multi-layered neural networks}

%\subsection{Abstraction of neural nets}

\label{sec:multilayer}

\subsubsection{General algorithm}

\label{La c'est vraiment dans l'idee qu'on a la double description generee en section 3, on dira apres la section 5 que si on a juste la repres interne, on peut faire un layer, avec subdivision, sinon si on a juste la repres externe, on peut faire plusieurs layers (on peut faire l'intersection), mais avec des approximations de la subdivision (pas implemente?)}

%We chose to use the external description for tropical polyhedrons as it is easier to calculate the intersection of two polyhedron with the external description than with the internal one, and because having inequalities between input nodes and output nodes is more useful than having only extreme tropical points.
The method developed in Section \ref{sec:1layer} is the cornerstone of our algorithm for analysing  neural networks. 
%for over-approximating classical linear functions with tropical polyhedron can be used for analysing neural networks.
A ReLU neural net consists of a chain of two kinds of computations, one which applies a classical linear transformation to their inputs, and another one which applies a ReLU function. %, which is a tropical linear transformation, as in Figure \ref{fig:RELU:trop} (usually both kinds are merged and the output of each node is the ReLU of the previous classical linear transformation).
We have seen that the affine map transformation can be over-approximated using tropical polyhedra. ReLU being a tropical affine function, the ReLU transform is exact in tropical polyhedra. 
%Both of these transformations can be over-approximated, or represented exactly (as is the case with the computation of ReLU in tropical polyhedra, since ReLU is a tropical affine function), with tropical polyhedra, so 
It is thus possible to use tropical polyhedra to represent reachable states for every node in the network, at least for one layer ReLU networks.

\begin{example}
We carry on with Example \ref{ex:running} and 
complete the final computations of Example \ref{ex:runningzone}.
%: 
%$h_1=x_1-x_2-1$, $h_2=x_1+x_2+1$. 
%$y_1=max(0,x_1-x_2-1)$ and $y_2=max(0,x_1+x_2+1)$.
%We have respectively, $\delta_{1,1}=2$, $\delta_{1,2}=0$, $\delta_{2,1}=2$, $\delta_{2,2}=2$, $\Delta_{1,1}=0$, $\Delta_{1,2}=0$, $\Delta_{2,1}=4$,  $\Delta_{2,2}=0$, $d_{1,1}=-3$, $d_{1,2}=-1$, $d_{2,1}=1$, $d_{2,2}=-1$, $m_1=-3$, $m_2=-1$, $M_1=1$ and $M_2=3$. Hence the external description for the tropical polyhedron relating values of $x_1$, $x_2$, $h_1$ and $h_2$ are:
The external representation is given by the tropical linear inequalities of Example \ref{ex:runningzone} together with inequalities $max(0,h_1)\leq y_1 \leq max(0,h_1)$ and $max(0,h_2) \leq y_2 \leq max(0,h_2)$.
%\begin{eqnarray}
% \max (
%x_1 -1,x_2 -1,h_1 - 1,h_2 - 3
%)
%& \leq & 0 \label{ex:running:ineq1}
%\\
%\begin{equation}
%\max (
%0, h_1+1,h_2 -1
%)  & \leq & x_1 +1 \label{ex:running:ineq2}\\
%\max (
%0, h_1 - 1,h_2 - 1
%)  & \leq & x_2 +1 \label{ex:running:ineq3} \\
%\end{equation}
%\begin{equation}
%\max( 
%0, x_1 +1,x_2 - 1, h_1 +3,h_2-1
%) & \leq & h_1 +3 \label{ex:running:ineq4}\\
%\max( 
%0, x_1 +1,x_2 +1,h_1 +1,h_2 +1
%) & \leq & h_2 +1 \label{ex:running:ineq5}
%\end{eqnarray}
%\noindent which encode all zones inequalities:
%$$\begin{array}{rcl}
%-1\leq & x_1 & \leq 1\\
%-1\leq & x_2 & \leq 1\\
%-3 \leq & h_1 & \leq 1\\
%-1 \leq & h_2 & \leq 3\\
%-2\leq & h_1-x_1 & \leq 0 \\
%-4 \leq & h_1-x_2 & \leq 2 \\
%0 \leq & h_2-x_1 & \leq 2 \\
%0 \leq & h_2-x_2 & \leq 2 \\
%-4 \leq & h_1-h_2 & \leq 0\\
%\end{array}$$ %\todo{to be checked}
%The first three tropical inequalities (\ref{ex:running:ineq1}), (\ref{ex:running:ineq2}) and (\ref{ex:running:ineq3} encode the zone for $x_1$, $x_2$, $h_1$ and $h_2$. 
%\todo{Make the explicit computation and relate to Figures \ref{fig:simplenet:zone}, \ref{fig:simplenetinnerrange}.}
Now the corresponding tropical polyhedron is generated by the linear tropical operator ReLU on each of the extremal points $A$, $B_1$, $B_2$, $C_1$ and $C_2$ and gives the two extra (last) coordinates in the axes $(x_1,x_2,h_1,h_2,y_1,y_2)$,  %\todo{To be completed}:
%\begin{eqnarray*}
$A'  =  (-1, -1, -3, -1,0,0), \
B'_1  =  (
1, -1, -1, 1,0,1), \
B'_2  =  (-1, 1, -3, 1,0,1), \
C'_1  = 
(-1, -1, 1, 1,1,1), \ 
C'_2  =  (
-1, 1, -1, 3,0,3)$. 
%\end{eqnarray*}
The projections of theses 5 extreme points on $(h_1,y_2)$ give
the points $(0,0)$, $(0,1)$, $(1,1)$, $(0,3)$ among which $(0,1)$ is in the convex hull of 
$A'=(0,0)$, $B'_2=B_2=(1,1)$ and $B'_1=(0,3)$ represented in Figure \ref{fig:simplenet:trop}. 
%Sylvie
\end{example}

%\begin{figure}
 %   \centering
 %   \includegraphics[scale=0.7]{Network.png}
  %  \caption{Classical and tropical transformations in a neural network}
   % \label{network}
%\end{figure}

The polyhedron given by the method of Section \ref{sec:1layer} only gives relations between 2 layers (the input and the first hidden layer). In order to get a polyhedron that represents the whole network when combining with e.g. another layer, we need to embed the first polyhedron from a space that represents only 2 layers to a higher space that represents the complete network, with one dimension per node. We will then need to intersect the polyhedra generated by each pair of layers to get the final result. Finally, as we are only interested in the input-output abstraction of the whole network, we can reduce computing costs by removing the dimensions corresponding to middle layers once those are calculated.

To this end, we use the following notations. 
Let $\mathcal{L} \subset \{L_0, \hdots, L_N\}$ be a set of layers, layer $i$ containing $n_{i+1}$ neurons as in Definition \ref{def:MLP}. Let $n$ be the sum of all $n_{i+1}$, with $i$ such that $L_i \in \mathcal{L}$  %=\sum\limits_{i=0 \mbox{, $i$ s.t. $L_i \in \mathcal{L}}$}^{N}} n_{i+1}$ 
and $\mathcal{S}_\mathcal{L} \equiv \R_{max}^{n}$ %\todo{unable to parse this notation?} 
be the tropical space in which we are going to interpret the values of the neurons on layers in $\mathcal{L}$, with each dimension of $\mathcal{S}_\mathcal{L}$ corresponding to a node of a layer of $\mathcal{L}$.

For $\mathcal{L}_1, \mathcal{L}_2 \subset \{L_0, \hdots, L_N\}$, for $\mathcal{H} \subset \mathcal{S}_{\mathcal{L}_1}$ a tropical polyhedron, we denote by $Proj(\mathcal{H}, \mathcal{L}_2) \subset \mathcal{S}_{\mathcal{L}_2}$
%\todo{Proj quand vers dim inferieure sinon embedding} 
the projection of $\mathcal{H}$ onto $\mathcal{S}_{\mathcal{L}_2}$ when $\mathcal{S}_{\mathcal{L}_2} \subseteq \mathcal{S}_{\mathcal{L}_1}$ and let $Emb(\mathcal{H}, \mathcal{L}_2) \subset \mathcal{S}_{\mathcal{L}_2}$ be the embedding of $\mathcal{H}$ into $\mathcal{S}_{\mathcal{L}_2}$ when $\mathcal{S}_{\mathcal{L}_1} \subseteq \mathcal{S}_{\mathcal{L}_2}$.

The main steps of our algorithm for over-approximating the values of neurons in a multi-layer ReLU network are the following:
\begin{itemize}
    
    \item We start with an initial tropical polyhedron $\mathcal{H}_0 \subset \mathcal{S}_{\{L_0\}}$ that represents the interval ranges of the input layer $L_0$.
    \item For each additional layer $L_{i+1}$:
    \begin{itemize}
    \item Calculate an enclosing hypercube $C_i$ for the nodes of layer $L_i$, given the current abstraction $\mathcal{H}_i \subset \mathcal{S}_{\mathcal{L}_i}$ (Section \ref{sec:zone:troppoly}).
    \item Calculate the polyhedron $\mathcal{P}_{i+1}$ representing relationships between layer $L_i$ and the new layer $L_{i+1}$, for nodes of layer $L_i$ taking values in $C_i$, as described in Section \ref{sec:1layer}: Theorem \ref{thm:RmtoRn} for the external description, %(Theorem \ref{prop:optimal_linear_octagon} for the enhanced version based on octagons),
        and Theorem \ref{thm:intRmtoRn} for the internal description. %(Theorem \ref{prop:tropoctagon} for the enhanced version based on octagons) %\todo{OK Sebastien?}
        \item Let $\mathcal{L}'_{i+1} = \mathcal{L}_i \cup \{L_{i+1}\}$. Calculate $\mathcal{P}'_{i+1} = Emb(\mathcal{P}_{i+1}, \mathcal{L}'_{i+1})$ (see below).
        %\item embed polyhedron ${\cal P}_{i+1}$ as a polyhedron ${\cal P}'_{i+1}$ in the space of all nodes of the network, as described in below. % and \ref{sec:embedint}
        \item Intersect $\mathcal{P}'_{i+1}$ with the projection (using the internal description, see below) of the previous abstraction $\mathcal{H}_i$ to get $\mathcal{H}'_{i+1} = Emb(\mathcal{H}_i, \mathcal{L}'_{i+1}) \cap \mathcal{P}'_{i+1}$ (using the external description).
        \item Choose $\mathcal{L}_{i+1} \supset \{L_{i+1}\}$, and calculate $\mathcal{H}_{i+1} = Proj(\mathcal{H}'_{i+1}, \mathcal{L}_{i+1})$.  Usually, we would use $\mathcal{L}_{i+1} = \{L_0, L_{i+1}\}$ if we only want relations between the input and output layers, or $\mathcal{L}_{i+1} = \{L_0, \hdots, L_{i+1}\}$ if we want relations between every layer.
        %\item (Optional, improves performance) Remove the inner layers from the polyhedric representation of ${\cal H}_{i+1}$, by projection. 
%        \item Calculate the ranges for the nodes of the next layer.
    \end{itemize}
\end{itemize}

We need now to describe the projection and embedding functions $Proj$ and $Emb$.
Let $\mathcal{L}_2 \subset \mathcal{L}_1 \subset \{L_0, \hdots, L_N\}$ be two sets of layers. Let $\mathcal{H}$ be a polyhedron on $\mathcal{S}_{\mathcal{L}_1}$. We have $\mathcal{H}' = Proj(\mathcal{H}, \mathcal{L}_2) = \{(x_i)_{L_i \in \mathcal{L}_2}, (x_i)_{L_i \in \mathcal{L}_1} \in \mathcal{H}\}$, i.e. for each point in $\mathcal{H}$, we only keep the dimensions corresponding to layers in $\mathcal{L}_2$, and discard the other dimensions. Projecting is easy with the internal description of polyhedron, as we can project the extreme points of $\mathcal{H}$ to get generators of $\mathcal{H}'$. However, we do not have a simple algorithm to project the external description of a polyhedron.

Let $\mathcal{L}_1 \subset \mathcal{L}_2 \subset \{L_0, \hdots, L_N\}$ be two sets of layers, and $\Delta$ be
%$= \sum\limits_{i=0,\ldots,n \mbox{ s.t. } L_i \in \mathcal{L}_2 \setminus \mathcal{L}_1} n_{i+1}$
the sum of $n_{i+1}$, the number of neurons of layer $L_i$, for $i$ such that $L_i \in \mathcal{L}_2 \setminus \mathcal{L}_1$. Let $\mathcal{H}$ be a polyhedron on $\mathcal{S}_{\mathcal{L}_1}$. We note that $\mathcal{S}_2 \equiv \mathcal{S}_1 \times \R_{max}^{\Delta}$, and thus $\mathcal{H}' = Emb(\mathcal{H}, \mathcal{L}_2) \equiv \mathcal{H} \times \R_{max}^{\Delta}$, i.e. we add dimensions corresponding to each node in $\mathcal{L}_2$ which are not in $\mathcal{L}_1$, and let points in $\mathcal{H}'$ take any value of $\R_{max}$ on these dimensions. 
Embedding is based on simple matrices concatenations in the external description, see Appendix \ref{sec:embedext} for more details. Embedding using the internal description is more involved and is explained after exemplifying things on a simple example. 
%The details on embedding the internal or external description of a polyhedron are explained respectively in Sections \ref{sec:embed}. Before giving such details, we illustrate how this works on our small running example. 

%We therefore need to explain how we embed a tropical polyhedron in a higher-dimensional space. %Intersections can be taken care of using the external representation. 
\begin{example} %[Running example]
\label{ex:running:2layers}
We consider the 1-layer neural net of Example \ref{ex:running}, and add a second layer. %, to get the neural net depicted in Figure \ref{fig:running2} (biases are $b_1=-1$, $b_2=1$, $c_1=-1$ and $c_2=1$). The new linear layer is 
The new linear layer is defined by $u_1=y_2-y_1-1$, $u_2=y_1-y_2+1$ and 
the output neurons are $z_1=max(0,u_1)=max(0,y_2-y_1-1)$ and $z_2=max(0,u_2)=max(0,y_1-y_2+1)$.

The enclosing cube for the tropical polyhedron ${\mathcal H}$ containing the values of neurons of the first layer $L_1$: $y_1$, $y_2$ of Example \ref{ex:running} is $[0,1]\times [0,3]$. The analysis of the second layer $L_2$, supposing its input belongs to $[0,1]\times [0,3]$ gives the constraint (an extract of the external representation of the resulting tropical polyhedron ${\mathcal H}'$)  
%\begin{align}
$-3 \leq u_1-y_1 \leq 2 \label{eq:2ndlayer:u1y1}, \ 
-2 \leq u_1-y_2 \leq -1 \label{eq:2ndlayer:u1y2}, \
-2 \leq u_2-y_1 \leq 1 \label{eq:2ndlayer:u2y1}, \
-5 \leq u_2-y_2 \leq 2 \label{eq:2ndlayer:u2y2}, \
z_1 = max(0,u_1), \
z_2 = max(0,u_2)$. 
%\end{align}
The intersection of the embedding $Emb({\mathcal H}',\{L_0,L_1,L_2\})$ with the embedding $Emb({\mathcal H}, \{L_0,L_1,L_2\})$ consists, as we saw above, in concatenating the tropical constraints, in the common space of variables. This implies in particular that we add the constraint 
%\begin{equation}
$    -3\leq y_1-y_2 \leq 0 \label{eq:1stlayer:y1y2}$ 
%    \end{equation}
    %\noindent 
    to the above equations. %(\ref{eq:2ndlayer:u1y1}), (\ref{eq:2ndlayer:u1y2}), (\ref{eq:2ndlayer:u2y1}) and (\ref{eq:2ndlayer:u2y2}). 
    The intersection is actually a zone intersection, where we have to normalize the corresponding DBM. A manual calculation shows that this will make use of the equalities 
%\begin{align}
$u_2-y_2 = (u_2-y_1)+(y_1-y_2) \label{eq:2ndlayer:refineu2y2}, \
u_1-y_1 = (u_1-y_2)+(y_2-y_1) \label{eq:2ndlayer:refineu1y1}$.
%\end{align}
%\noindent 
By combining equations, 
%(\ref{eq:2ndlayer:u2y1}) with (\ref{eq:1stlayer:y1y2}), and Equation (\ref{eq:2ndlayer:u1y2}) with (\ref{eq:1stlayer:y1y2}), 
we get the refined bounds (refined lower bound for the first equation, refined upper bound for the second equation)
%\begin{align}
$-2 \leq u_1-y_1 \leq 2 \label{eq:2ndlayer:u1y1}, \
 -5 \leq u_2-y_2 \leq 1 \label{eq:2ndlayer:u2y2}$.
%& z_1 = max(0,u_1) \\
%& z_2 = max(0,u_2) \\
%\end{align}
\end{example}

%\subsubsection{Embedding functions}

%\label{sec:embed}
%\paragraph{Embedding tropical polyhedra: external description}

\paragraph{Embedding a tropical polyhedron: internal description}

\label{sec:embedint}

In this paragraph, we embed a polyhedron into a higher dimensional space, using the internal description. 
%This will allow us use combine the polyhedra representing relations between each pair of layers in order to determine all relations in the network.

%As in the previous section, consider the case of the embedding of tropical polyhedron ${\cal P}_i\subset \{L_i,L_{i+1}\}$ into $S_{\{L_i,L_{i+1},L_{i+2}\}}$. 
Suppose $\mathcal{H}$ is a tropical polyhedron in $\R^n$ (such as $\mathcal{P}_i$ in the previous section) %and let
%$P = (p_i)_{1\leq i \leq m}$ be the extrema of $\mathcal{H}$. Suppose 
that we want to embed $\mathcal{H}$ into a larger space, with an extra coordinate, which we consider bounded here within $[a,b]$. So we need to determine a presentation of the tropical polyedron $\mathcal{H}'=\mathcal{H}\times [a,b]$.  
%The objective of this paragraph is to find the extrema and rays of $\mathcal{H}' = \mathcal{H} \times [-\infty,\infty]$,  which is a tropical polyhedron in $\R_{max}^{n+1}$. %\todo{Sebastien: explain why this is linked to the problem explained using the external description, above}. 
%The extrema of $\mathcal{H} \times \R_{max}$ can be found as the limit case of the extrema for $\mathcal{H}' = \mathcal{H} \times [a,b]$, with $a$ and $b$ going toward $-\infty$ and $\infty$. 

%For the more general case of $\mathcal{H}'' = \mathcal{H} \times \R_{max}^k$, we can get the extreme points of $\mathcal{H}''$ by simply repeating the algorithm $k$ times.

Supposing we have $m$ extreme points $p_i$ for representing $\mathcal{H}$, a naive method consists in noticing that the family $(p_i, a), (p_i, b)$ is a generator of $\mathcal{H}'$ and removing non-extreme points from that list. But that would exhibit poor performance, as we get $m\times 2^k$ extreme points for $\mathcal{H}''$. We can in fact do better:

%\begin{theorem}

%\label{thm:I}
%\end{theorem}

%\begin{proof}

%\end{proof}

\begin{theorem}
\label{thm:embedint}
The extreme points of $\mathcal{H}'$ are $\{(p_i, a), 1 \leq i \leq m\} \cup \{(p_i, b), i \in I\}$, 
where $I$ is a subset of indexes of generators of $\mathcal{H}$, $I \subset [1,m]$, such that:
    \begin{eqnarray}
&    \forall i \in I, \forall j \in [1,m] \setminus \{i\}, p_i \oplus p_j \neq p_i \\
&\forall i \in [1,m] \setminus I, \exists j \in [1,m] \setminus \{i\} \mbox{ s.t. } p_i \oplus p_j = p_i
\label{eq:embed}
    \end{eqnarray}
\end{theorem}

%Let $p_{\min} = (\min\limits_{i=1}^m(p_{i,j}))_{1 \leq j \leq n})$, then we have:
%we have for all $i \in [1,m]$, $p_{\min} \leq p_i$.

%\begin{theorem}
%The extreme points of $H'$ are as follows. Either $p_{\min} \in P$ and $(p_i, a)_{1 \leq i \leq m}, (p_{\min}, b)$ are the extreme points of $H'$. Or else, let $I \subset [1,m]$ such that:
%    \begin{eqnarray}
%&    \forall i \in I, \forall j \in [1,m] \setminus \{i\}, p_i \oplus p_j \neq p_i \\
%&\forall i \in [1,m], \exists j \in [1,m] \setminus \{i\} \mbox{ s.t. } p_i \oplus p_j = p_i
%\label{eq:embed}
%    \end{eqnarray}
%    \noindent and the extreme points of $H'$ are $(p_i, a)_{1 \leq i \leq m}, (p_i, b)_{i \in I}$.
%\end{theorem}
    
The proof is given in Appendix \ref{proof:embedint}. 
Passing to the limit, this shows that the extreme points of $\mathcal{H}\times \R$ are $(p_i,-\infty)$, $i=1,\ldots,m$ and the extreme rays are $(p_i,0)$, $i\in I$ for the smallest %\todo{EG: j'en suis la}
%\todo{To be checked} 
$I$ verifying Equation \ref{thm:I}. 
In the current implementation, we do not use extreme rays and embed $\mathcal{H}$ into larger state spaces by using large enough values for $a$ and $b$.

\subsubsection{Checking properties on ReLU neural nets}

%\todo{EG: currently here}

Given an affine guard $$h(x,y)=\sum\limits_{i=1}^{m} h_i x_i +\sum\limits_{j=1}^n h'_j y_j+ c$$ where $x_i$, resp. $y_j$ are the input, resp. output neurons, %\todo{plus clair si exprime comme fonction des x et y ou x sont les entrees et y les sorties?}, 
we want to determine whether, for all input values in $[-1,1]$, we have  $h(x)\geq 0$ (this can encode properties $(P_1)$ and $(P_2)$ of Example \ref{ex:running}). 

There are two ways to check such properties. The first one, that we have implemented, is as follows. We abstract the input output relation that the network under analysis encodes, using a tropical polyhedron ${\mathcal H}$ as described in Section \ref{sec:multilayer}. From this, we derive the smallest zone $Z$ containing ${\mathcal H}$ as in Section \ref{sec:zone:troppoly}. % Proposition \ref{prop:troptozone}. 
Finally, we solve the linear programming problem $m=\min\limits_{x,y \in Z} h(x,y)$ using any classical algorithm (we used \texttt{glpk} in our prototype). This is enough for checking $(P_1)$ in Example \ref{ex:running} since $m \geq 0$ proves our property true, but not $(P_2)$. 
The second way can be useful to check $(P_2)$: here we have no choice but try to solve $m=\min\limits_{x, y \in {\mathcal H}} h(x,y)$ which is not a convex optimization problem, in any sense (tropical nor classical). This could be encoded as MILP problem instead and is left for future work. 
%Using the external description for ${\mathcal H}$, we see constraints of our optimization problem is of the form $\max(b,a_1+x_1,\ldots,a_{m+n}+x_{m+n})\leq \max(c,d_1 + x_1,\ldots,d_{m+n}+x_{m+n})$. These constraints can be written as a conjunction of disjunction of inequalities $a_i+x_i \leq d_j+x_j$ plus $b \leq d_j+x_j$ and $a_i+x_i \leq c$. It is a standard trick, using big M relaxation, to encode a disjunction of such (classically) affine inequalities as a mixed integer linear inequality. 

%\todo{And the second way ? (to check such specifications)}

%\todo{En fait je me demande si ce paragraphe est vraiment utile?}

%1 sous-section sur la garde lineaire (et le probleme lineaire/MILP engendre, voir avec Louis si possible)

%1 sous-section sur la zone englobante d'un polyedre tropical, et l'application aux proprietes de classification/robustesse

\section{Improvements of the analysis}

\label{sec:subd}

%\subsection{Subdivisions}

We will refine here the tropical abstractions we have defined in Section \ref{sec:1layer}
by "subdividing" the support of the functions we are approximating. Since we are linearizing tropical rational functions, we expect to be closer and closer to the actual graph of the function, by doing enough subdivisions. %, we discuss this a little more in Section \ref{convexity} though). 

%\subsection{Improving the abstraction of affine scalar function, from $\R$ to $\R$}
We begin by showing how we can improve the abstraction of affine scalar functions $f$, in the particular case where $m=n=1$ i.e. when $f$ goes from $\R$ to $\R$, before we treat the much more involved general case in Theorem \ref{thm:subdgen}. 
Thus, we suppose first that we want to abstract the graph $\mathcal{G}_f$ of $f(x) = \lambda x + m$, $x \in [a,b]$, by a tropical polyhedron. 

We can get a more precise result than what we got in Section \ref{sec:1layer}, i.e. a smaller polyhedron that still contains all $\{ (x,f(x)) | x \in [a, b]\}$ by splitting the interval $[a,b]$ in $N$ sub-intervals $[c_k,c_{k+1}]$, with $c_0 = a$ and $c_N = b$ and calculating an over-approximation of $f$ on each sub-interval, and returning the tropical union of all these polyhedra. In fact, we can give again explicit external and internal representations of these unions of tropical polyhedra as follows. 

\begin{theorem}
\label{thm:subd}
A sound abstraction %\todo{To be checked: not sure how to define "smallest abstraction" in that case} 
as a tropical polyhedron $\mathcal{P}$ of the graph $\mathcal{G}_f$ of $f$ over $[a, b]$ given by subdividing the domain in $N$ sub-intervals has the following external representation of $N+2$ tropical inequalities, 
%\todo{EG: je suis en train de travailler ca 21/12/2020}:
%The polyhedron can also be given by its external description, in which case each subdivision adds a new inequality, for a total of $N + 2$ inequalities
$\mathcal{P}$ is defined by %\todo{Sebastien: shouldn't we add constraints on $f(x)$ wrt to $f(b)$ for each of the three cases? otherwise for $N=1$ this will not be equivalent to Theorem 1?}
%\todo{yes: de 1 a N-1 et les 2 autres contraintes} 
%$x \leq b$, $a \leq x$ 
%the following $N+2$ constraints, 
depending on the value of $\lambda$:
\begin{itemize}
    \item If $\lambda \leq 0$, we add the $N-1$ constraints $0 \leq \max(x - c_k, y - f(c_k))$ for 
    $k=1,\ldots,N-1$, to the ones of Theorem \ref{thm:RmtoRn}, i.e. $a \leq x$, $f(b)\leq y$ and $max(x-b,y-f(a))\leq 0$
    \item If $0 \leq \lambda \leq 1$, we add the $N-1$ constraints $y - f(c_k) \leq \max(0, x - c_k)$ for 
    $k=1,\ldots,N-1$, to the ones of Theorem \ref{thm:RmtoRn}, i.e.
    $y-f(a)\leq x-a$, $y-f(b)\leq 0$ and $max(x-b+f(b),f(a))\leq y$
    \item If $\lambda \geq 1$, we add the $N-1$ constraints $x - c_k \leq \max(0, y - f(c_k))$
    for 
    $k=1,\ldots,N-1$, to the ones of Theorem \ref{thm:RmtoRn}, i.e.
    $\max(y - f(b) + b, a) \leq x$, $x \leq b$, 
$x - a \leq f(x) - f(a)$. 
\end{itemize}
$\mathcal{P}$ can also be internally represented as the tropical convex hull of at most $N + 2$ extreme points $A$, $B$ and $C_i$, $i \in [1, N]$ with 
%\begin{center}
%\begin{tabular}{|l|c|c|c|}
%\hline
%\begin{itemize}
%\item 
$A=(a,f(a))$, 
$B=(b,f(b))$, 
%& $A$ & $B$ & $C_i$, $i \in [1,N]$ \\
%\item 
and $C$ is $(c_{i-1}, f(c_i))$ if $\lambda \leq 0$,
$(c_{i-1} + f(c_i) - f(c_{i-1}),  f(c_i))$ if $0 \leq \lambda \leq 1$ 
and $(c_i, f(c_{i-1}) + c_i - c_{i-1})$ if $\lambda \geq 1$.
%\end{itemize}
\end{theorem}

%\end{center}
\begin{proof}
Take any $k$ in $[0, N-1]$. 
Consider first the case $\lambda \leq 0$. 
Now if $x \leq c_k$, then $f(x) \geq f(c_k)$, and if $x \geq c_k$, then $f(x) \leq f(c_k)$.
    Therefore $0 \leq \max(x - c_k, f(x) - f(c_k))$.
    
    Now, suppose $0 \leq \lambda \leq 1$. 
    If $x \leq c_k$, then $f(x) \leq f(c_k)$, otherwise 
    if $x \geq c_k$, then $f(x) - f(c_k) \leq x - c_k$, and we conclude that 
   $f(x) - f(c_k) \leq \max(0, x - c_k)$.
    
    Finally, we consider the case $\lambda \geq 1$. 
    If $x \leq c_k$, then $f(x) \leq f(c_k)$, whereas 
    if $x \geq c_k$, then $f(x) - f(c_k) \geq x - c_k$. This means that
    $x - c_k \leq \max(0, f(x) - f(c_k))$.
\end{proof}

\begin{figure}
\begin{center}
\includegraphics[scale=0.15]{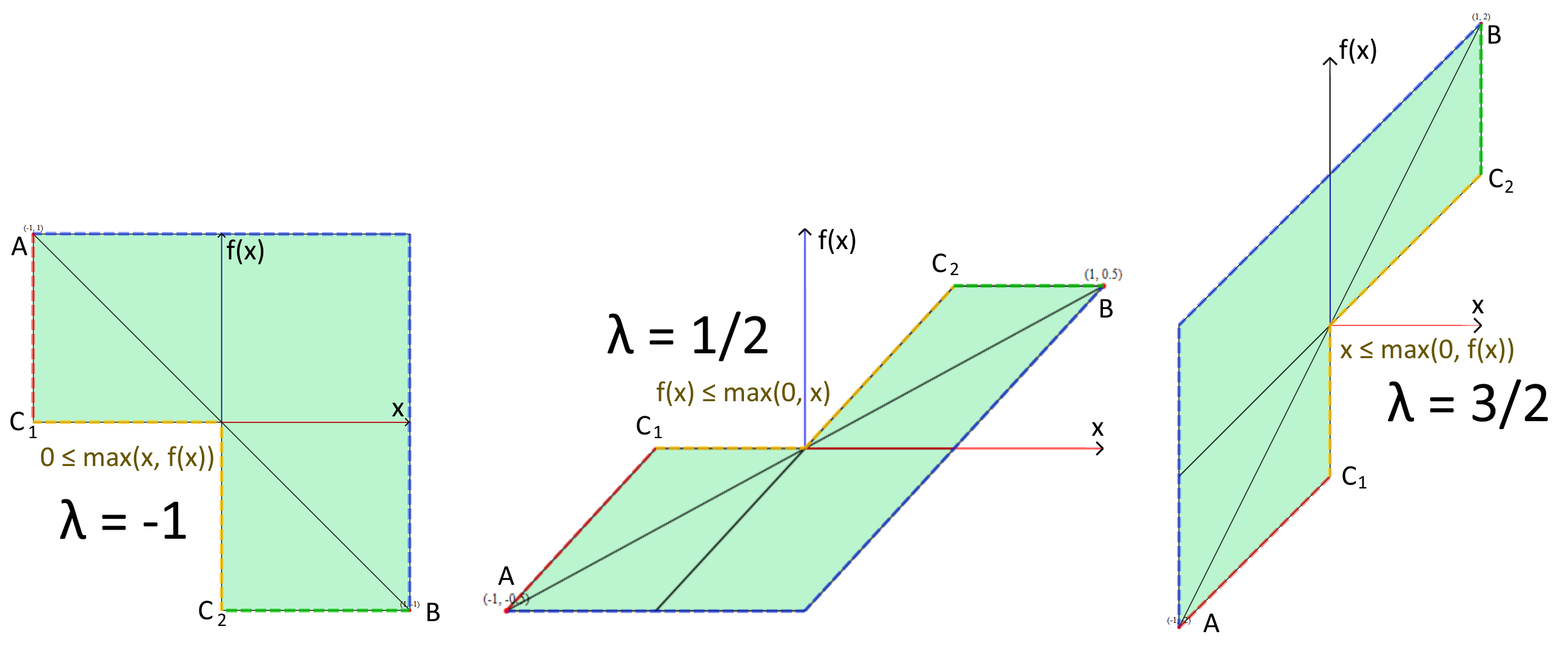}
\caption{Possible subdivisions ($N=2$) for abstracting $f:\R\rightarrow\R$}
\label{div2}
\end{center}
\end{figure}

Compare Figure \ref{div2} with Figure \ref{fig1} to see the improvement due to subdivision. Still, the tropical inequalities that we added using subdivisions give more precision on only "one side" of the polyhedron. This will not be the case with the abstraction based on the octagon abstraction of Section \ref{sec:abstraction:octagon}.   %In fact, we can show that the tropical polyhedron of Theorem \ref{thm:subd} converges towards the tropical convex hull of the graph of the function, which will "lower" convexify the abstraction, and which is not in general a tropical polyhedron unless the function itself is tropically linear (see Appendix \ref{convexity} for more details).

%\subsubsection{
Now we are considering the general case where $f$ is an affine function from $\R^m$ to $\R^n$. 
We consider here subdivisions $c^0_i=a_i,\ldots c^N_i=b_i$ of intervals $[a_i, b_i]$, $i=1,\ldots,m$ in $N$ subintervals, and want to generalize the previous result to this higher dimensional case.  
Contrarily to the previous case ($f: \R \rightarrow \R$), we have no explicit external nor internal representation for the tropical polyhedron, union of 
%We can subdivide the input domain $[a_1,b_1] \times \ldots [a_m,b_m]$ in any number of ways, compute 
the tropical polyhedra, abstraction of the graph of $f$ over each subdomain using Theorem \ref{thm:RmtoRn}. 
We could still take their unions, as a tropical polyhedron, using the double description method \cite{Tropical}. In the sequel, we describe a sound abstraction of this union, as a tropical polyhedron, that is generally sufficient for our purpose and does not need the computational complexity of the double description method. 
%Instead of taking unions of small tropical polyhedra accounting for the abstraction of the graph of a function over a subdomain, as before, to find a tighter tropical polyhedron enclosing the graph of a function, we take the view here that we can add additional tropical inequalities to the enclosing tropical polyhedron of Section \ref{tropicalf}.  

%The relations between individual $x_i$ and individual $y_j$ correspond to those in the previous case.\newline

\begin{theorem}
\label{thm:subdgen}
A sound %\todo{Can you check the formulation Sebastien?} 
and tighter over-approximation of $\mathcal{G}_f$ than the one of Theorem \ref{thm:RmtoRn} is given externally by the tropical constraints of Theorem \ref{thm:RmtoRn} plus the following constraints, for any subdivision $c^0_i=a_i,\ldots c^N_i=b_i$ of intervals $[a_i, b_i]$, $i=1,\ldots,m$ in $N$ subintervals:
\begin{itemize}
\item 
If $\lambda_{i,j} \leq 0$, then we add the constraint $0 \leq \max(x_i -c^k_i, y_j - m_j + \lambda_{i,j} (b_i - c^k_i))$. Otherwise     
if $0 \leq \lambda_{i,j} \leq 1$, we add the constraint $y_j - M_j + \lambda_{i,j} (b_i - c^k_i) \leq \max(0, x_i - c^k_i)$.
And finally, we add $x_i - c^k_i \leq \max(0, y_j - m_j - \lambda_{i,j} (c^k_i - a_i))$
if $\lambda_{i,j} \geq 1$. 
\item 
  Let $I_{j-} \subset [1,m]$, maximal, such as for all $i \in I_{j-}$, $\lambda_{i,j} \leq 0$.
    We write $\sigma_{j-} = \sum\limits_{i \in I_{j-}} \lambda_{i,j}(b_i - c^k_i)$. Then, we add the constraint $0 \leq \max(y_j - m_j + \sigma_{j-}, \max\limits_{i \in I_{j-}}(x_i - c^k_i))$.
    \item 
    Let $I_{j0} \subset [1,m]$, maximal, such as for all $i \in I_{j0}$, $0 \leq \lambda_{i,j} \leq 1$ and $\sum\limits_{i \in I_{j0}} \lambda_{i,j} \leq 1$ and let $\sigma_{j0} = \sum\limits_{i \in I_{j0}} \lambda_{i,j}(b_i - c^k_i)$. Then we add the constraint 
    $y_j - M_j + \sigma_{j0} \leq \max(0, \max\limits_{i \in I_{j0}}(x_i - c^k_i))$.
\item Finally, 
for any $J$ subset of $[1,n]$, let $\sigma_{i,J} = \sum\limits_{j \in J} \lambda_{i,j}$ and 
$m_J=\sigma_{0,J} + \sum\limits_{\sigma_{i,J} < 0} \sigma_{i,J} b_i + \sum\limits_{\sigma_{i,J} > 0} \sigma_{i,J} a_i$.
For $j \in J$, let $u_j \in [m_j, M_j]$ such as $\sum\limits_{j \in J} u_j = m_J$: 
we add the constraint 
$0 \leq \max_{j \in J}(y_j - u_j)$ %\todo{How many constraints added then, quite a lot?}.
\end{itemize}
\end{theorem}

\begin{figure}
\begin{center}
\includegraphics[scale=0.25]{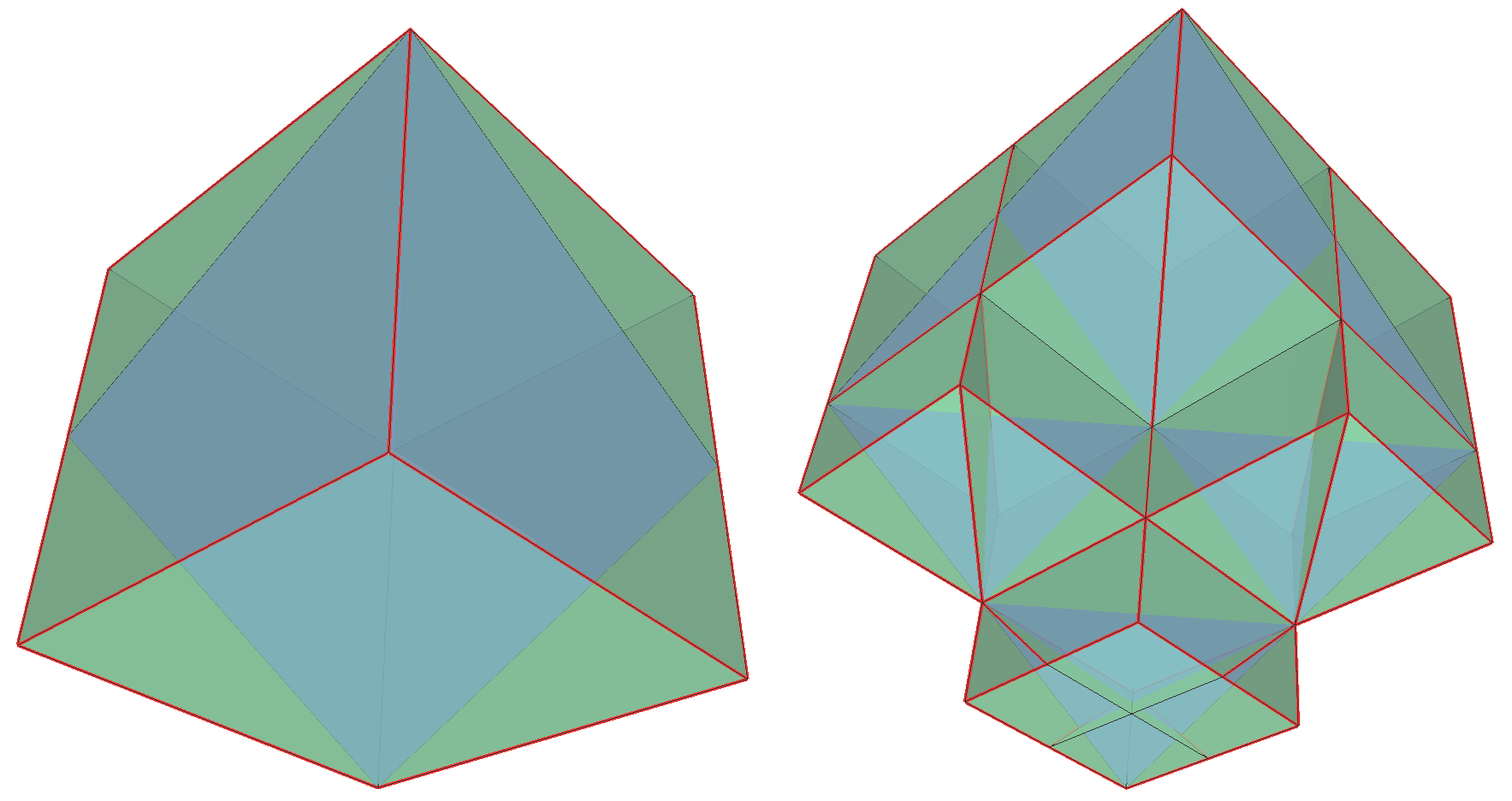}
\caption{Over-approximation in $\R^2\rightarrow\R$ with $\lambda_1 = \lambda_2 = -0.5$, with no subdivision (left), and with three extra inequalities (right): $0 \leq max(x_1, x_2, y)$, $0 \leq max(x_1, y + 0.5)$ and $0 \leq max(x_2, y + 0.5)$}
\label{fig7}
\end{center}
\end{figure}

\begin{figure}
\begin{center}
\includegraphics[scale=0.25]{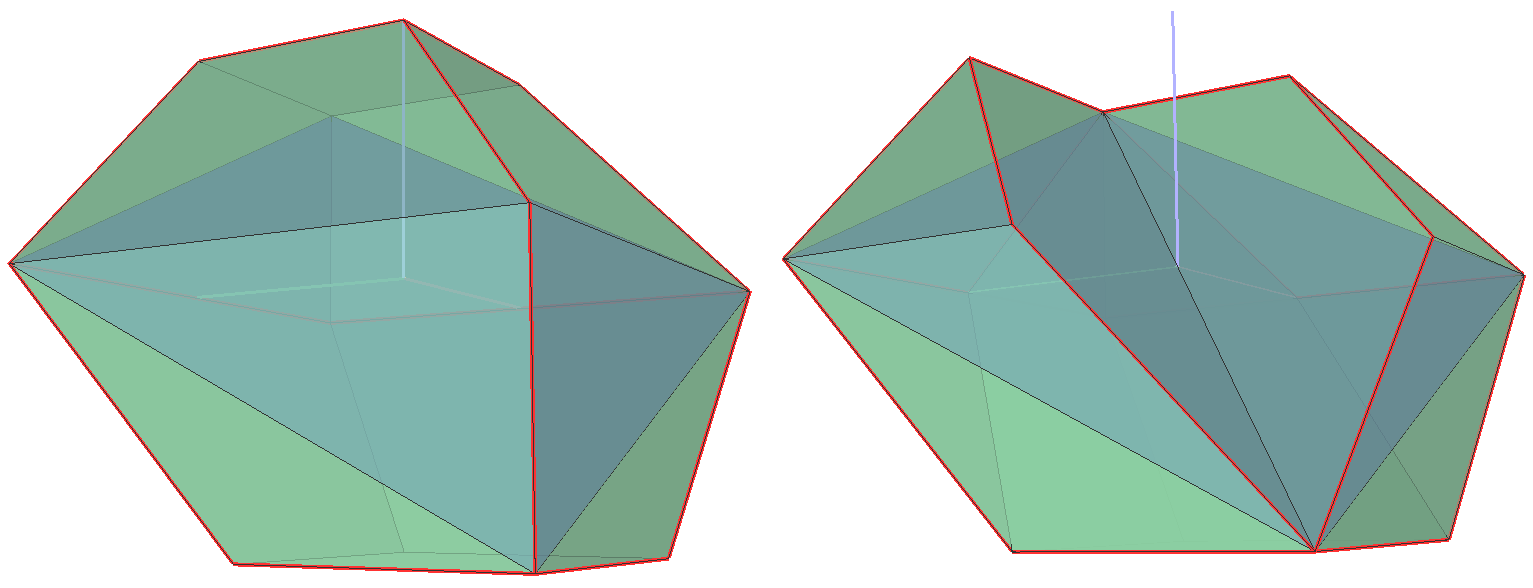}
\caption{Over-approximation in $\R^2\rightarrow\R$ with $\lambda_1 = \lambda_2 = +0.5$, with no subdivision (left), and with one extra inequalities (right): $y \leq max(x_1, x_2)$}
\label{fig8}
\end{center}
\end{figure}

We exemplify this in Figures \ref{fig7} and \ref{fig8}. %\todo{Sebastien: would be nice to write down the 3 extra inequalities of these examples.}. 
The proof is given in Appendix \ref{sec:proofsubd}.

\section{Implementation, experiments and benchmarks}
\label{sec:implexp}

\subsubsection{Internal, external and double description methods}

Overall, we have developed methods for propagating an outer-approximation of the values that the different layers of neurons can take, within a MLP with ReLU activation. Let us discuss the pros and cons of using the internal description, external description and double description methods:
\begin{itemize}
    \item The double description method allows for possibly using subdivisions, propagating values in multiple layers and projecting them onto a subset of interesting neurons (e.g. input and output layers), as well as computing an enclosing zone, for synthesizing classification properties. We have implemented this in a prototype using Polymake \cite{polymake}, whose results we briefly discuss below. 
    \item The internal description allows for analyzing one layer networks, %\todo{Mention each time the proposition/theorem that should be used there}, 
    using subdivisions, %(see Section \ref{sec:subd})\label{Rajouter un mot tres tot (union sur sous-domaines, comme une remarque}, 
    project onto an interesting subset of neurons, as well as computing an enclosing zone (Section  \ref{sec:zone:troppoly}). We have implemented this method in C++ in a standalone prototype, \texttt{nntrop}, that takes as input a Sherlock file \cite{sherlock-tool} describing the one hidden layer neural net to analyze plus a linear formula to be checked, and returns the tropical abstraction of the values that neurons can take, %have been performed % for the full pipeline (reading a sherlock file, computing the tropical abstraction of the neural network (input neurons are supposed to be always between -1 and 1), 
    its over-approximation by a zone, and whether the linear specification is satisfied or not. % over-approximating by an octagon, and then solving an LP problem), as well as for the pipeline without LP.
    \item The external description allows for analyzing multiple layer networks (see Section \ref{sec:multilayer}). %, with possibly a conservative approximation of the effect of subdivisions. % (Theorem \ref{thm:subdgen}). 
\end{itemize}

The double description method is much more expensive since the translation between the internal and external representations may be quite complex. %\todo{Complexity bounds?} 

%\subsection{Implementation of experiments}

%\subsection{Double description}

%We implemented the algorithm described previously to perform tests on different networks. We use the Polymake framework's tropical package as it provides structures and methods to represent and manipulate vectors, matrices and tropical polyhedra.
%We can use either the internal or the external description of polyhedra, however the methods that manipulate those may be more efficient on one description than one the other. Most importantly, calculating the enclosing hypercube of a polyhedron is much faster with its internal description than with its external one. Furthermore, polymake provides a method to intersect a polyhedron given by its internal description with one given by its external description, and we can intersect two polyhedra given their external description by concatenating their matrices. However there is no method to intersect two polyhedra given their internal description. Lastly, polymake does not provide methods to project a polyhedron onto a lower-dimensional space, or to embed it in a higher-dimensional space, so we have to implement the methods that were described in this paper. Therefore we cannot project a polyhedron onto a lower space given only its external description.

%We implemented the algorithm using both descriptions, and they give the same results. However the program runs much faster when using the internal description as we can project to remove inner layers while calculating, and thus reduce costs.

\subsubsection{Experiments and benchmarks}

%nntrop: when several hidden layers: enveloping hypercube, no intersection
% ex: time ./nntrop -B -i 1 --input-file running2.sherlock

We briefly compare the computation times between internal description only and double description in Table \ref{table1}. For each example, we indicate in the columns \texttt{\# inp.} the number of input neurons, \texttt{\# out.} the number of output neurons, \texttt{\# hid.} the number of hidden layers, \texttt{\# neur.} is the total number of neurons (input, output and hidden), \texttt{t. intern} is the time spent for computing the internal representation and \texttt{t. double} for the double description of the tropical polyhedron abstracting the corresponding neural net. Experiments are performed on a simple computer with ArchLinux and a Intel(R) Core(TM) i5-7200U CPU @ 2.50GHz.

We of course see the influence of a potential exponential complexity for going back and forth between internal and external descriptions, but also the fact that we relied on a perl (interpreted) implementation of tropical polyhedra (the one of polymake \cite{polymake}, with exact rational arithmetics), which is much slower than the C++ implementation we wrote for the internal description method (although the internal description method does work in a twice as big space because it considers the octagon instead of just zone abstraction). 
%polymake: time includes recompilation of extensions within polymake, perl interpreted language and rationals used for representing numbers
% ex: time polymake Inequalities.pl < tora_modified_controller_1.sherlock

%Total number of neurons is the number of internal neurons (not output, not input, should count those: dim of the pb)

%Should begin with this and discuss the first examples

\begin{table}
\caption{Execution times (internal and double description) on sample networks.}
\begin{center}
\begin{tabular}{|l|c|c|c|c|c|c|}
\hline
Example & \# inp. & \# out. & \# hid. & \# neur. & t. intern. (s) & t. double (s) \\
\hline
\texttt{running} & 2 & 2 & 0 & 4 & 0.006 & 1.83 \\
\hline
\texttt{running2} & 2 & 2 & 1 & 6 & 0.011 & 4.34 \\
\hline
\texttt{multi} & 2 & 8 & 1 & 13 & 0.005 & 3.9 \\
\hline
\texttt{krelu} & 2 & 2 & 0 & 4 & 0.011 & 1.94 \\
\hline
tora\_modified\_controller & 4 & 1 & 1 & 6 & 0.005 & 14.57 \\
\hline
tora\_modified\_controller\_1 & 4 & 1 & 1 & 105 & 0.75 & 815.12 \\
\hline 
quadcopter\_trial\_controller\_3 & 18 & 1 & 1 & 49 & 0.009 & 102.54 \\
\hline 
quadcopter\_trial\_controller\_1 & 18 & 1 & 1 & 69 & 0.2 & 469.77 \\
\hline 
quad\_modified\_controller & 18 & 1 & 1 & 20 & 0.005 & 14 \\
\hline
car\_nn\_controller\_2 & 4 & 2 & 1 & 506 & 104.75 & -- \\
\hline
car\_nn\_controller\_1 & 4 & 2 & 1 &  506 & 88.8 %104.57 
& -- \\ 
\hline
ex & 2 & 1 & 5 & 59 & 0.195 & 1682.28 \\
\hline
\end{tabular}
\label{table1}
\end{center}
\end{table}
%\todo{Exactement les memes resultats avec car\_nn\_controller\_2 et car\_nn\_controller\_1 ?}
In 
Table \ref{table1}, \texttt{running} is the network of Example \ref{ex:running}, and \texttt{running2} is the extension with an extra layer of Example \ref{ex:running:2layers}, discussed in great length in these examples. Example \texttt{krelu} is the running example from \cite{krelu} that we discuss at the end of this section, and \texttt{tora\_modified\_controller}, 
\texttt{tora\_modified\_controller\_1}, 
\texttt{quadcopter\_trial\_controller\_3}, 
\texttt{quadcopter\_trial\_controller\_1}, 
\texttt{quad\_mo\-dified\_controller}, 
\texttt{car\_nn\_controller\_2}, 
\texttt{car\_nn\_controller\_1} and \texttt{ex}
are examples from the distribution of Sherlock \cite{sherlock-tool}. \texttt{ex} is a multi-layer example for which the algorithm using only the internal representation does not compute the intersection of tropical polyhedra between layers (involving the external representation), contrarily to the double description prototype. 
We now discuss some of these examples below. 

Network \texttt{multi} is a simple 2-layer, 13 neurons example with inputs $x_1, \ x_2$, outputs $y_1,y_2,\ldots,y_8$ and 

$
\begin{bmatrix} h_1 \\ h_2 \\ h_3 \end{bmatrix} =
ReLU \begin{pmatrix}
\begin{bmatrix}
1 & 1 \\ 1 & -1 \\ -1 & -1
\end{bmatrix}
\begin{bmatrix} x_1 \\ x_2 \end{bmatrix}
\end{pmatrix}
$,
    $\begin{bmatrix} y_1 \\ y_2 \\ y_3 \\ y_4 \\ y_5 \\ y_6 \\ y_7 \\ y_8 \end{bmatrix} =
max \left(\begin{pmatrix}
\begin{bmatrix}
1 & 1 & 1 \\ 1 & 1 & -1 \\ 1 & -1 & 1 \\ 1 & -1 & -1 \\ -1 & 1 & 1 \\ -1 & 1 & -1 \\ -1 & -1 & 1 \\ -1 & -1 & -1
\end{bmatrix}
\begin{bmatrix} h_1 \\ h_2 \\ h_3 \end{bmatrix}
\end{pmatrix},0\right)$. 
%
%We exemplify our algorithm based on zones only, on the network of 
%Figure \ref{fig:neuralnet:seb}. %\todo{Sebastien: Explain more (in particular the results)}:
%The first layer is described in \ref{fig:neuralnet:seb}, with zero biais, and the second layer is given by the formula in Figure \ref{fig:neuralnet:seb:eq}. 
%From $|x_1 + x_1| + |x_1 - x_2| \leq 2$, we can easily find the following relations:
%\begin{itemize}
%    \item $h_1 - h_3 = x_1 + x_2 \in [0, 2]$.
%    \item $h_1 + h_3 = |x_1 + x_2| \in [0, 2]$.
%    \item $h_1 + h_2 + h_3 \in [0, 2]$
%    
%    \item $y_1 = |x_1 + x_2| + h_2 \in [0, 2]$.
%    \item $y_2 = ReLU(x_1 + x_2 + h_2) \in [0, 2]$.
%    \item $y_3 = ReLU(|x_1 + x_2| - h_2) \in [0, 2]$.
%    \item $y_4 = ReLU(x_1 + x_2 - h_2) \in [0, 2]$.
%    \item $y_5 = ReLU(-x_1 - x_2 + h_2) \in [0, 2]$.
%    \item $y_6 = ReLU(h_2 - |x_1 + x_2|) \in [0, 2]$.
%    \item $y_5 = ReLU(-x_1 - x_2 - h_2) \in [0, 2]$.
%    \item $y_8 = 0$.
%\end{itemize}
Our zone based abstraction returns the following ranges: 
%\todo{Sebastien: pick up some of this info between the actual range/relations and the abstraction and explain - list some interesting zone relations below}:
$y_1 \in [0, 6]$, $y_2 \in [0, 4]$, $y_3 \in [0, 4]$, $y_4 \in [0, 2]$, $y_5 \in [0, 4]$, $y_6 \in [0, 2]$,  $y_5 \in [0, 2]$ and $y_8 = 0$, whereas the exact ranges for $y_1$ to $y_7$ is $[0,2]$. Our algorithm is thus exact for $y_4$, $y_6$, $y_7$ and $y_8$ but not $y_1$, $y_2$, $y_3$ nor $y_5$. This is due to the fact that the zone-based tropical abstraction does represent faithfully the differences of neuron values, but not sums in particular. For instance, $y_2=max(0,2x_1)$ which cannot be represented exactly by our method. 
%\todo{Subdivisions? Octagons, by hand?}

%\end{itemize}

%As expected, the algorithm gives precise results for differences, as these can be represented tropically, but poor precision for sums (when all coefficients have the same sign) as those cannot be represented with a tropical polyhedron.
 
Network \texttt{krelu} is {a 2 layer 4 neurons example from \cite{krelu}}. %, pictured 
We get the correct bounds on the outputs: $0 \leq z_1, z_2 \leq 2$, as well as relations between the inputs and the outputs: $z_j \leq x_i + 1$. However, we do not have significant relations between $z_1$ and $z_2$, as those are not tropically linear.
%\todo{Compare the results with what is obtained with 1-RELU and 2-RELU in \cite{krelu}}\newline
%The complete matrix also contains information on the internal layers, i.e. the variable $y_1$ and $y_2$ before the ReLU function.
We refer to the results obtained with 1-ReLU and 2-ReLU in \cite{krelu}:  % are shown on figure \ref{k-ReLU}.
%\todo{Important to show the calculation with 2-ReLU, there are some octagon ideas behind it I think.}
%As we see, 
they both get better relations between $z_1$ and $z_2$, in particular $z_1+z_2 \leq 2$ which is not representable in a tropical manner (except by using an octagon based abstraction, which is outside the scope of this paper). However 1-ReLU does not keep track of relations between the inputs and the outputs, and has sub-optimal relations between the outputs, as it cannot represent the non linear ReLU function exactly. 2-ReLU, on the other hand gets both the relation between the output variables, and between the inputs and outputs correct, but is more computationally expensive.

\begin{wrapfigure}{r}{0.3\textwidth}
%\begin{figure}
%begin{center}
%\begin{subfigure}[b]{0.37\textwidth}
%\begin{center}
%\includegraphics[scale=0.37]{bench_2D_without_LP.png}
%\caption{ x-axis is the total number of neurons, y-axis is time.}
%\label{fig12}
%\end{center}
%\end{subfigure}
%\begin{subfigure}[b]{0.55\textwidth}
%\begin{figure}
\centering
\includegraphics[scale=0.15]{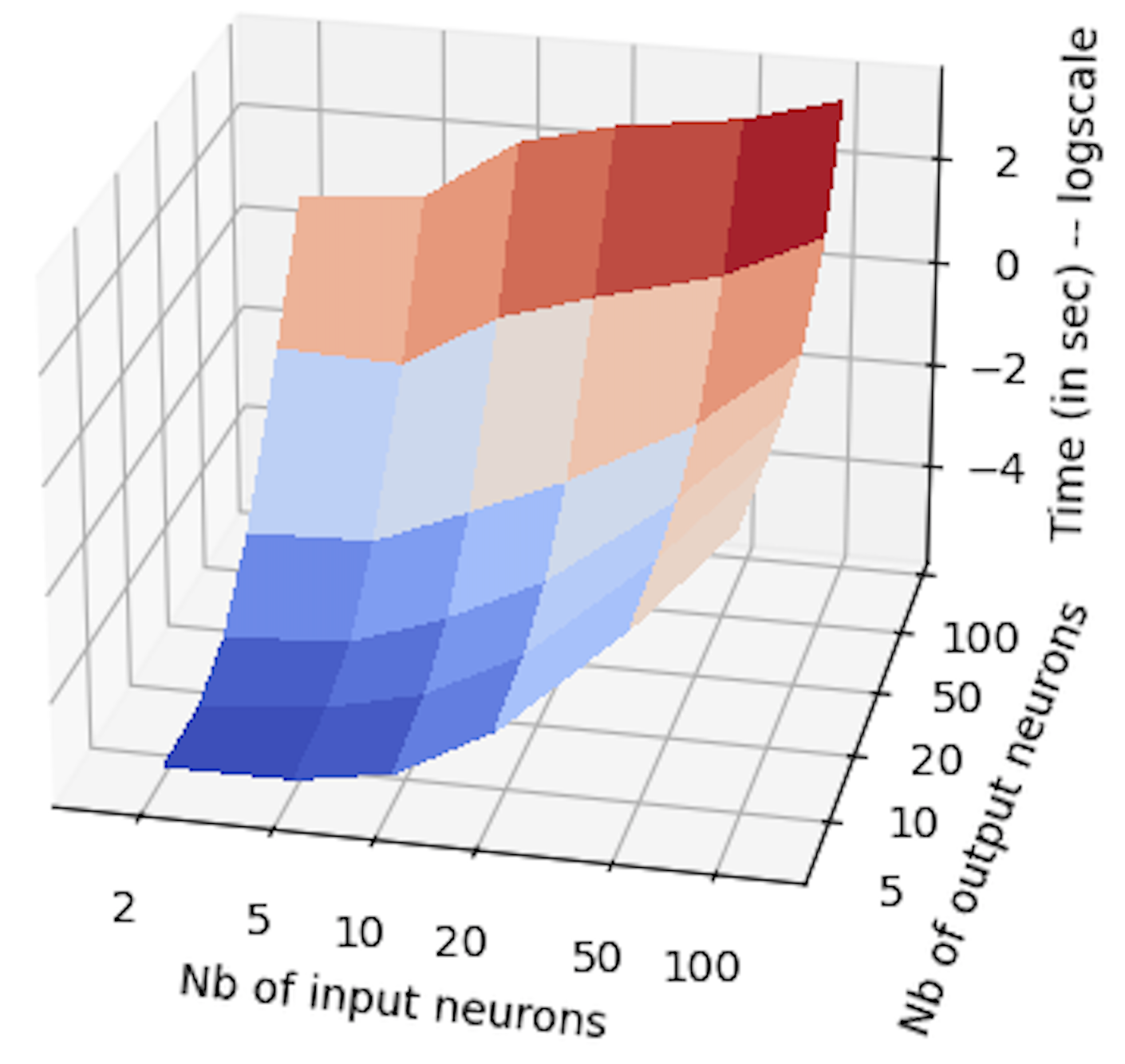}
%\caption{The x-axis is number of input neurons, y-axis is the number of output neurons, and z-axis is time.}
\label{fig16}
%\begin{center}
%\includegraphics[scale=0.5]{bench_log3D_with_LP.png}
%\caption{Full pipeline in log scale. x-axis is number of input neurons, y-axis is number of output neurons, z-axis is time.}
%\label{fig15}
%\end{center}
\end{wrapfigure}
In order to assess the efficiency of the internal description methods, we have run a number of experiments, 
%Experiments have been made 
with various number of inputs and ouputs for neural nets with one hidden layer only. The linear layers are generated randomly, with weights  between -2 and 2.
For 100 inputs and 100 neurons in the hidden layer, the full pipeline  %(reading the Sherlock file, analyze the network, and 
(checking the linear specification in particular) 
%given as a linear formula) 
took about 35 seconds, among which the tropical polyhedron analysis took 6 seconds. Timings are shown in the %Fig. 
figure on the right (demonstrating the expected complexity, cubical in the number of neurons), %\ref{fig11} %, 
%\ref{fig12} %, \ref{fig13}, \ref{fig14}, \ref{fig15} 
%and 
%\ref{fig16}
where the x-axis is number of input neurons, y-axis is the number of output neurons, and z-axis is time. 

\section{Conclusion and future work}

We have explored the use of tropical polyhedra as a way to circumvent the combinatorial complexity of neural networks with ReLU activation function. 
%These precise and time-efficient abstractions of affine maps in the tropical polyhedra domain are the first of their kind, and should be applicable to verification of timed systems and programs as well %\todo{Ca se mord un peu la queue, on est partis justement des polyedres tropicaux dans le monde des programmes - ou alors il faut decrire un peu plus ce que l'abstraction apporte de nouevau pour les programmes? Peut-etre plutot au moment de la conclusion?}. 
The first experiments we made show that our approximations are tractable when we are able to use either the internal or the external representations for tropical polyhedra, and not both at the same time. This is akin to the results obtained in the classical polyhedron approach, where most of the time, only a sub polyhedral domain is implemented, needing only one of the two kinds of representations. It is interesting to notice that a recent paper explores the use of octohedral constraints, a three-dimensional counterpart of our octagonal representations, in the search of more tractable yet efficient abstraction for ReLU neural nets \cite{muller2021precise}. 
%are indeed quite incomparable in general with more classical convex methods, but very often very precise\todo{A bit weak and awkward for now, waiting for more examples etc. to see what should be the right angle here.}.
This work is a first step towards a hierarchy of approximations for ReLU MLPs. We have been approximating the tropical rational functions that these neural nets compute by tropical affine functions, and the natural continuation of this work is to go for higher-order approximants, in the tropical world. 
%Future work: algorithmics/implementation (get rid of double description and abstraction just with extremal points; discussed a bit here though...
%Compare on this with \cite{muller2021precise}!
%\todo{Important comparison!!}
We also believe that the tropical approach to abstracting ReLU neural networks would be particularly well suited to verification of ternary nets \cite{ternary1}. %- neural networks with weights constrained to +1, 0 and -1. %- that correspond to rational quotients of rational affine functions \todo{A verifier et cite geometry of tropical networks, C'est decrit ou ca?}, often\todo{Pas si often?} close enough to their tropical linearisations. 
These ternary nets have gained importance, in particular in embedded systems: simpler weights mean smaller memory needs and faster evaluation, and it has been observed \cite{ternary2} that they can provide similar performance  to general networks. 
%+ octogons, tropical polynomials, tropical grobner bases, simple polynomial representations as in classical program analysis...

%\todo{A mettre plutot en conclusion, par rapport a Xavier et Antoine meme sur les zones} 

%\section{Examples}

\bibliography{bibliographie}

\newpage

\appendix

\section{Proof of Proposition \ref{prop:troprepofzones}}
\label{proof:troprepofzones}

\begin{proof}\,
%\begin{itemize}[label=$ $]
    %\item 
    We begin by proving that {$H_{int} \subset H_{ext}$}. 
    For this, we only need to prove that the generators are in $H_{ext}$.
    
%    \begin{itemize}[label=$\bullet$]
%        \item 
For all $i$ and $j$ such that $0\leq i,j\leq n$, we know that  $c_{0,i}+c_{i,j} \geq c_{0,j}$ as the zone representation we started with is closed. Thus $a_i - a_j = -c_{0,i} + c_{0,j} \leq c_{i,j}$ and $A\in H_{ext}$.
        
        %\item 
Similarly,  for all $k \in [1,n]$, we have $b_{k,i}-b_{k,j} =  (c_{k,0} - c_{k,i})-(c_{k,0} - c_{k,j}) = -c_{k,i} + c_{k,j} \leq c_{i,j}$ as the initial zone is closed, hence $B_k  \in H_{ext}$.
        
%        Let now $*\in [0,n]$\todo{Je n'ai pas corrig\'e mais ce n'est pas tr\`es classique en math un indice qui n'est pas une lettre - ptet changer?}, $(x_0,\dots,x_n)=B_*$, and  $i\in[0,n]$.
%        We have $x_*-x_i = b_* - b_* + \delta_{*,i} \leq \delta_{*,i}$.
 %       Moreover, $x_i-x_* = -\delta_{*,i}$ where $0\leq \delta_{*,i}+\delta_{i,*}$, thus $x_i-x_*\leq\delta_{i,*}$.
 %       Therefore, $B_*\in H_{ext}$.
    
    We then prove that {$H_{ext} \subset H_{int}$}.
    Let $x=(x_1,\dots,x_n)\in H_{ext}$.
 %   Set $P'=(0\otimes A) \otimes \bigotimes_{1\leq i \leq n} ((x_i-b_i)\otimes B_i) \in H_{int}$. 
     We define $x'=(x'_i)_{1\leq i \leq n}$ by $$  x'_i  = \max(a_i, \max_k(x_k-c_{k,i})).$$
Noting that  we can rewrite $$  x'_i = \max(a_i,\max_k(x_k-c_{k, 0}+c_{k, 0}-c_{k,i})), $$ we have $x'=\max(a,\max_k(\lambda_k+b_k)),$ with $\lambda_k \leq 0$, thus $x' \in H_{int}$.
 %   \begin{itemize}[label = $\bullet$]
        %\item 
        
        Moeover, $x'_j\geq x_j-c_{j,j} = x_j$ ($c_{j,j}$ being equal to zero for a closed zone). Finally, $a_j\leq x_j$ and for each $i\in[1,n]$, $x_i-c_{i,j}\leq x_j$ by definition of $x\in H_{ext}$, thus $x'_j \leq x_j$.
%    \end{itemize}
    We conclude $x' = x$, and $H_{ext} \subset H_{int}$.
%\end{itemize}
\end{proof}

\section{Proof of Proposition \ref{prop:optimal_linear_zone}}

\label{proof:optimal_linear_zone}

\begin{proof}
The tightest zone is obtained as the conjunction of the bounds $\ux_j\leq x_j \leq \ox_j$ on input $x$, given as hypercube $K$, the bounds on the $y_i$ and $y_{i_1}-y_{i_2}$ obtained by a direct computation of bounds of the affine transform of the input hypercube $K$, and finally the bounds on the differences $y_i - x_j$ which computation is detailed below.
Noting as in Proposition  \ref{prop:optimal_linear_zone} \[ m_i = \min_{(x,y)\in S} y_i = \sum_{w_{i,j}<0}w_{i,j}\ox_j + \sum_{w_{i,j}>0}w_{i,j}\ux_j + b_i  \]  and \[ M_i = \max_{(x,y)\in S} y_i = \sum_{w_{i,j}<0}w_{i,j}\ux_j + \sum_{w_{i,j}>0}w_{i,j}\ox_j + b_i, \] we can rewrite 
$y_i-x_j = (M_i-\ux_j) + (y_i-M_i)-(x_j-\ux_j) $ and $y_i-x_j = (m_i-\ox_j) + (y_i-m_i)-(x_j-\ox_j) $ 
and consider now the following cases:
\begin{itemize}
        \item if $w_{i,j} \leq 0$: in that case, $ (y_i-M_i)-(x_j-\ux_j)  \leq max_{x} ((w_{ij} x_j - w_{ij} \ux_j) -(x_j-\ux_j)) = max_{x}  (w_{ij}-1) (x_j-\ux_j) = 0$ (which is $-\delta_{i,j}$) and the bound on $ (y_i-M_i)-(x_j-\ux_j)$ is reached for $x_j=\ux_j$; fo the  other bound, we have
        $ (y_i-m_i)-(x_j-\ox_j) \geq \min_{x} ((w_{ij} x_j - w_{ij} \ox_j) -(x_j-\ox_j)) = \min_{x} (w_{ij} -1) (x_j-\ox_j)=0$ (which is $\delta_{i,j}$) reached for $x_j=\ox_j$.
%        \todo{Il me semble que c'est plus simple de dire: (on a besoin de l'inegalite plus forte que quand on doit comparer $w_{i,j}$ avec 1)}
%        in that case, $y_i-M_i \leq 0$ and $x_j-\ux_j\geq 0$ so $y_i-x_j=(M_i-\ux_j) + (y_i-M_i)-(x_j-\ux_j)\leq M_i -\ux_i=M_i-\ux_i-\delta_{i,j}$ since $\delta_{i,j}=0$ here
%        \todo{Et il faut encore demontrer que la borne est tight}
        \item if $w_{i,j} \geq 0$: in that case, $ (y_i-M_i)-(x_j-\ux_j)  \leq max_{x} ((w_{ij} x_j - w_{ij} \ox_j)-(x_j-\ux_j))$ %\todo{Ptet une ligne de plus pour ecrire le calcul $y_i-M_i\leq w_{i,j}x_j-w_{i,j}\underline{x}_j$?} 
        and $ (y_i-m_i)-(x_j-\ox_j) \geq \min_{x} ((w_{ij} x_j - w_{ij} \ux_j) -(x_j-\ox_j))$; we need to distinguish 2 sub-cases: 
        \begin{itemize}
        \item  if $0 \leq w_{i,j} \leq 1$: then $max_{x} ((w_{ij} x_j - w_{ij} \ox_j)-(x_j-\ux_j))= w_{ij} (\ux_j-\ox_j)$ (which is $-\delta_{i,j}$) and is reached for $x_j=\ux_j$ and $\min_{x} ((w_{ij} x_j - w_{ij} \ux_j) -(x_j-\ox_j)) = w_{ij}(\ox_j-\ux_j)$ (which is $\delta_{i,j}$) and is reached for $x_j=\ox_j$
         \item if $w_{i,j} \geq 1$: then $max_{x} ((w_{ij} x_j - w_{ij} \ox_j)-(x_j-\ux_j))= (\ux_j-\ox_j)$ (which is $-\delta_{i,j}$) and is reached for $x_j=\ox_j$ and $\min_{x} ((w_{ij} x_j - w_{ij} \ux_j) -(x_j-\ox_j)) = (\ox_j-\ux_j)$ (which is $\delta_{i,j}$) and is reached for $x_j=\ux_j$.
         \end{itemize}
 \end{itemize}
In all cases, these bounds correspond to the inequalities $m_i - \ox_j + \delta_{i,j} \leq y_i - x_j   \leq M_i - \ux_j - \delta_{i,j}$ of Proposition  \ref{prop:optimal_linear_zone}, with $\delta_{i,j}=\min_{(x,y)\in S}\big((y_i-m_i)-(x_j-\ox_j)\big) 
= -\max_{(x,y)\in S}\big((y_i-M_i)-(x_j-\ux_j)\big).$
%\todo{A finir}
\end{proof}

\section{Proof of Proposition \ref{prop:optimal_linear_octagon}}

\label{proof:optimal_linear_octagon}

\begin{proof}
This is a direct extension of Proposition \ref{prop:optimal_linear_zone}. The bounds on $x_j$, $y_i$, $y_{i_1}-y_{i_2}$ and $y_{i}-x_{j}$ are computed similarly. The bounds on the sums $y_{i_1} +y_{i_2}$ are easy to obtain. Let us concentrate on the computation bounds on $y_{i}+x_{j}$. Similarly as in the proof of Proposition\ref{prop:optimal_linear_zone}, 
we first note that we can rewrite 
$y_i + x_j = (M_i + \ox_j) + (y_i-M_i) + (x_j-\ox_j) $ and $y_i + x_j = (m_i + \ux_j) + (y_i-m_i) + (x_j-\ux_j) $ 
and consider the following cases:
\begin{itemize}
        \item if $w_{i,j} \leq 0$: in that case, $ (y_i-M_i)+(x_j-\ox_j)  \leq max_{x} ((w_{ij} x_j - w_{ij} \ux_j) + (x_j-\ox_j))$  and  $(y_i-m_i)+(x_j-\ux_j) \geq \min_{x} ((w_{ij} x_j - w_{ij} \ox_j) + (x_j-\ux_j)) $; we need to distinguish 2 sub-cases:
        \begin{itemize}
        \item  if $w_{i,j} \leq -1$: then $max_{x} ((w_{ij} x_j - w_{ij} \ux_j) + (x_j-\ox_j)) = (\ux_j-\ox_j)$ (which is $-\gamma_{ij}$)  reached for $x_j= \ux_j$ and $\min_{x} ((w_{ij} x_j - w_{ij} \ox_j) + (x_j-\ux_j)) = (\ox_j-\ux_j)$ (which is $\gamma_{ij}$) reached for $x_j= \ox_j$
        \item  if $-1 \leq w_{i,j} \leq 0$: then $max_{x} ((w_{ij} x_j - w_{ij} \ux_j) + (x_j-\ox_j)) = w_{ij} (\ox_j-\ux_j)$ (which is $-\gamma_{ij}$)  reached for $x_j= \ox_j$ and and $\min_{x} ((w_{ij} x_j - w_{ij} \ox_j) + (x_j-\ux_j)) = w_{ij} (\ux_j-\ox_j)$ (which is $\gamma_{ij}$) reached for $x_j= \ux_j$
        \end{itemize}   
        \item if $w_{i,j} \geq 0$: in that case,$ (y_i-M_i)+(x_j-\ox_j)  \leq max_{x}((w_{ij} x_j - w_{ij} \ox_j) +(x_j-\ox_j))= max_{x} (w_{ij}+1)(x_j-\ox_j)=0$ and $(y_i-m_i)+(x_j-\ux_j) \geq \min_{x} ((w_{ij} x_j - w_{ij} \ux_j)+(x_j-\ux_j))=0$ (which is $\gamma_{ij}$).
\end{itemize}     
In all cases, these bounds correspond to the inequalities $m_i + \ux_j + \gamma_{i,j} \leq y_i + x_j   \leq M_i + \ox_j - \gamma_{i,j}$ of Proposition  \ref{prop:optimal_linear_zone}, with $\gamma_{i,j}=\min_{(x,y)\in S}\big((w_{ij} x_j - w_{ij} \ox_j) + (x_j-\ux_j)\big) 
=-\max_{(x,y)\in S}\big((y_i-M_i) + (x_j-\ox_j) \big).$
%octagons are simply concretization of zones defined on the extended space\todo{Dire ca mieux quand on a ecrit les octagones a partir des zones en section 2}.
\end{proof}

\section{Proof of Proposition \ref{prop:tropoctagon}}

\label{proof:tropoctagon}

\begin{proof}
Let $M$ be the linear manifold defined above. 
    The octagon defined in Proposition \ref{prop:optimal_linear_octagon} is the concretization of the zone defined as the set of all $(x^+,y^+,x^-,y^-)\in \R^m\times\R^n\times\R^m\times\R^n$ satisfying the inequalities below, intersected with $M$: 
    \begin{align*}
     &\Bigg(\bigwedge_{1\leq j\leq m} \ux_j\leq x_j^+ \leq \ox_j\Bigg)
      \wedge \Bigg(\bigwedge_{1\leq i\leq n} m_i\leq y_i^+ \leq M_i\Bigg)
      \wedge \Bigg(\bigwedge_{1\leq j\leq m} -\ox_j\leq x_j^- \leq -\ux_j\Bigg) \\
      & \wedge \Bigg(\bigwedge_{1\leq i\leq n} -M_i\leq y_i^- \leq -m_i\Bigg)
      \wedge \Bigg(\bigwedge_{i,j} m_i - \ox_j + \delta_{i,j} \leq y_i^+ - x_j^+  \leq M_i - \ux_j - \delta_{i,j} \Bigg) \\
      & \wedge \Bigg(\bigwedge_{i,j} \ux_j - M_i + \delta_{i,j} \leq y_i^- - x_j^- \leq \ox_j - m_i - \delta_{i,j}  \Bigg)\\
      & \wedge \Bigg(\bigwedge_{i,j} m_i + \ux_j \leq y_i^+ - x_j^- + \gamma_{i,j}  \leq M_i + \ox_j - \gamma_{i,j} \Bigg) \\
      & \wedge \Bigg(\bigwedge_{i,j} -M_i - \ox_j + \gamma_{i,j} \leq y_i^- - x_j^+ \leq -m_i - \ux_j - \gamma_{i,j}\Bigg)\\
      & \wedge \Bigg(\bigwedge_{1\leq i_1, i_2 \leq n} y_{i_1}^+ - y_{i_2}^+ \leq \Delta_{i_1,i_2}\Bigg)
      \wedge \Bigg(\bigwedge_{1\leq i_1, i_2 \leq n}  y_{i_1}^- - y_{i_2}^- \leq \Delta_{i_2,i_1} \Bigg) \\
      & \wedge \Bigg(\bigwedge_{1\leq i_1, i_2 \leq n} L_{i_1,i_2} \leq y_{i_1}^+ - y_{i_2}^- \leq \Gamma_{i_1,i_2}\Bigg)
    \end{align*}
    Just like we did not have any non-redundant inequality on $x_i-x_j$, we have no non-redundant inequality on $x_i+x_j$.
%\end{proposition}
Thanks to this reformulation, we can once again use the translation procedure detailed in the proof of Theorem~\ref{thm:intRmtoRn} to get the internal tropical representation in the extended domain which constitutes the result of this proposition. 
\end{proof}

\section{Proof of Theorem \ref{thm:RmtoRn}}

\label{proof:RmtoRn}

\begin{proof}
It can be checked easily that the inequalities %of Theorem \ref{thm:RmtoRn} \ref{eq:RmtoRn1}, \ref{eq:RmtoRn2} and \ref{eq:RmtoRn3} 
are equivalent to the inequalities defining zone $\mathcal{H}_f$ in Proposition \ref{prop:optimal_linear_zone}. For instance, inequality \ref{eq:RmtoRn2} is equivalent  to:
$$\left(x_j-\underline{x}_j \geq 0\right)\wedge \left(\bigwedge\limits_{1 \leq i \leq n} (y_i-M_i+\delta_{i,j} \leq x_j - \underline{x}_j)\right)$$
\noindent which is in turn equivalent to:
$$\left(\underline{x}_j \leq x_j\right)\wedge \left(\bigwedge\limits_{1 \leq i \leq n} (y_i-x_j \leq M_i-\underline{x}_j-\delta_{i,j})\right)$$
\end{proof}

\section{Embedding a tropical polyhedron: external description}

\label{sec:embedext}

%\todo{EG currently here}

In this paragraph, we treat the general case of multi-layer networks with an external description. 
%(which makes it easier to compute intersections). 
%We use the external description of tropical polyhedra, as we will be needing it to compute intersections. 
Tropical polyhedra is thus described here as sets of points $X \in \R_{max}^{n}$ satisfying the following tropical linear inequalities: 
$A_1 \tilde{X} \geq A_2 \tilde{X}$, where 
 $A_1$ and $A_2$ are two matrices of size $m \times (n+1)$, $X=(x_1,\ldots,x_n)$ and $\tilde{X}$ is the augmented vector $(\one, x_1, \ldots, x_n)$. %\todo{Use special character for 0 and 1 in the tropical world}. 
 %
 % , such as $A_1X \geq A_2X$ for each point $X$ in a polyhedron. The first coordinate of $X$ is the tropical multiplication unit $x_0 = 0$, therefore 
 This allows for using the classical homogenization trick, for representing affine inequalities as linear ones in an augmented space: the first column of $A_1$ and $A_2$ represent the constant part of the affine transformation. 

%i.e. $AX = \begin{bmatrix}
%a_{1,0} & a_{1,1} & \hdots & a_{1,n} \\
%\hdotsfor{4} \\
%a_{m,0} & a_{m,1} & \hdots & a_{m,n}
%\end{bmatrix}
%\begin{bmatrix}
%0 \\ x_1 \\ \vdots \\ x_n
%\end{bmatrix}
%= \max\begin{pmatrix}
%\begin{bmatrix}
%a_{1,1} & \hdots & a_{1,n} \\
%\hdotsfor{3} \\
%a_{m,1} & \hdots & a_{m,n}
%\end{bmatrix}
%\begin{bmatrix}
%x_1 \\ \vdots \\ x_n
%\end{bmatrix},
%\begin{bmatrix}
%a_{1,0} \\ \vdots \\ a_{m,0}
%\end{bmatrix}
%\end{pmatrix}
%$\newline

When representing a pair of neural network layers (one input layer and one output layer) with a tropical polyhedron, each node in each layer corresponds to a dimension in the polyhedron and to a column in the matrices of the polyhedron. Therefore, if the input layer has $n_{i}$ nodes and the output layer has $n_{i+1}$ nodes, the resulting matrices will have $n_{i} + n_{i+1} + 1$ columns (and as many rows if there are no subdivisions, one for each inequality): one column for the constants (in orange below), $n_{i}$ columns for the input nodes (numbered from $1$ to $n_{i}$, in red below) and $n_{i+1}$ columns for the output nodes (numbered from $n_{i} + 1$ to $n_{i} + n_{i+1}$, in blue below).

In order to embed a polyhedron to a higher-dimensional space, we need to add columns filled with $-\infty$ corresponding to the new dimensions.

Suppose that we have a polyhedron $\mathcal{P}_i \subset \mathcal{S}_{\{L_i, L_{i+1}\}}$ representing relations between layers $i$ and $i + 1$. This polyhedron is described by two matrices $A_1$ and $A_2$ such as $A_1 \tilde{X} \geq A_2 \tilde{X}$. Below, we colored in orange the first column, that encodes the affine part of the transformation, the part encoding $L_i$ is the set of columns in red, and the part encoding $L_{i+1}$ is the set of columns in blue: 

$$
A_j = \begin{bmatrix}
\color{orange}a_{1,0} & \color{red}a_{1,1} & \color{red}\hdots & \color{red}a_{1,n_i} & \color{blue}a_{1,n_i + 1} & \color{blue}\hdots & \color{blue}a_{1,n_i + n_{i+1}} \\
\hdotsfor{7} \\
\color{orange}a_{m,0} & \color{red}a_{m,1} & \color{red}\hdots & \color{red}a_{m,n_i} & \color{blue}a_{m,n_i + 1} & \color{blue}\hdots & \color{blue}a_{m,n_i + n_{i+1}}
\end{bmatrix}
$$
We get the two matrices corresponding to $Emb(\mathcal{P}_i, \{L_{i-1},L_i,L_{i+1}))$ by inserting a block of $-\infty$, depicted in gray below, corresponding to layer $i-1$, right after the first column (representing the affine part of the transformation): 
$$
Emb(A_j, \{L_{i-1}, L_i, L_{i+1}\}) = \begin{bmatrix}
\color{orange}a_{1,0} & \color{gray}-\infty & \hdots & \color{gray}-\infty & \color{red}a_{1,1} & \color{red}\hdots & \color{red}a_{1,n_i} & \color{blue}a_{1,n_i + 1} & \color{blue}\hdots & \color{blue}a_{1,n_i + n_{i+1}} \\
\hdotsfor{10} \\
\color{orange}a_{m,0} & \color{gray}-\infty & \hdots & \color{gray}-\infty & \color{red}a_{m,1} & \color{red}\hdots & \color{red}a_{m,n_i} & \color{blue}a_{m,n_i + 1} & \color{blue}\hdots & \color{blue}a_{m,n_i + n_{i+1}}
\end{bmatrix}
$$
We get the two matrices corresponding to $Emb(\mathcal{P}_i, \{L_i,L_{i+1},L_{i+2}\})$ by inserting a block of $-\infty$, depicted in gray below, corresponding to layer $i-1$, right after the last column:  
%after the first one for each node in new layers that precede $L_i$ and by inserting one column after the last column for each node in layers that follow $L_{i+1}$.
$$
Emb(A_j, \{L_i, L_{i+1}, L_{i+2}\}) = \begin{bmatrix}
\color{orange}a_{1,0} & \color{red}a_{1,1} & \color{red}\hdots & \color{red}a_{1,n_i} & \color{blue}a_{1,n_i + 1} & \color{blue}\hdots & \color{blue}a_{1,n_i + n_{i+1}} & \color{gray}-\infty & \hdots & \color{gray}-\infty \\
\hdotsfor{10} \\
\color{orange}a_{m,0} & \color{red}a_{m,1} & \color{red}\hdots & \color{red}a_{m,n_i} & \color{blue}a_{m,n_i + 1} & \color{blue}\hdots & \color{blue}a_{m,n_i + n_{i+1}} & \color{gray}-\infty & \hdots & \color{gray}-\infty
\end{bmatrix}
$$

\section{Proof of Theorem \ref{thm:embedint}}

\label{proof:embedint}

    \begin{proof}
    %Suppose first that $p_{\min} \in P$. 
    %Let $X = (x_i, \hdots, x_n) \in H$. 
    %Let $\lambda \in \R_{\max}^m$ such that $X = \bigoplus\limits_{i=1}^m \lambda_i p_i$ with  $\bigoplus\limits_{i=1}^m \lambda_i = 0$ and take any $x_{n+1} \in [a,b]$.
    %We have $Y = (x_1, \hdots, x_n, x_{n+1}) \in H'$ by definition.
    %Let $\lambda_{n+1} = x_{n+1} - b \leq 0$.
    %We have $X = \bigoplus\limits_{i=1}^m \lambda_i p_i \geq \bigoplus\limits_{i=1}^m \lambda_i p_{\min} \geq p_{\min}$ and $\lambda_{m+1} \leq 0$.
    %Therefore $X \oplus \lambda_{m+1} p_{\min} = X$.
    %
    %Let now $p'_i = (p_i, a)$ for $1 \leq i \leq m$ and $p'_{m+1} = (p_{\min}, b)$.
    %We have $\bigoplus\limits_{i=1}^{m+1} \lambda_i p'_i = (X, a \oplus \lambda_{m+1} b) = (X, a \oplus x_{n+1}) = (X, x_{n+1}) = Y$.
    %For $i \in [1, m]$: $p_i \in H$ thus $p'_i \in H'$, and $p_{\min} \in H$ therefore $p'_{m+1} \in H'$.
    %Therefore $P' = (p'_i)_{1\leq i \leq m+1}$ generates $H'$.

Let $I \subset [1,m]$ such that Equations \ref{eq:embed} hold.

First, we prove that this implies that: 
\begin{equation}
    \forall p \in P, \exists j \in I \mbox{ s.t. } p \oplus p_j = p
    \label{thm:I}
    \end{equation}

Let $i \in [1, m]$. Let $S_{p_i} = \{p \in P: p_j < p_i\}$. If $S_{p_i} = \emptyset$, then $i \in I$, thus we have $j = i \in I \mbox{ s.t. } p_j \leq p_i$.
Otherwise, for $p \in S_{p_i}$, we have $S_p \subsetneq S_{p_i}$ from the transitivity and the irreflexivity of $<$. Since $P$ is finite, we can prove the theorem by induction.

Suppose for $k \geq 0$, $\forall p \in P$, if $\#S_p \leq k$, then there exists $j \in I \mbox{ s.t. } p_j \leq p$.

The base case is $S_p = \emptyset$ which we have proven true.

Let $p \in P \mbox{ s.t. } \#S_p = k + 1 > 0$. Let $p' \in S_p$. We have $S_{p'} \subsetneq S_p$, thus $\#S_{p'} \leq k$ and, from the induction hypothesis, there exists $j \in I \mbox{ s.t. } p_j \leq p' < p$. Therefore the induction hypothesis holds for $k+1$.

Therefore, for $p \in P$, there exists $j \in I \mbox{ s.t. } p_j \leq p$.

    %\todo{prove $I$ not empty}
    Now, let $X = (x_i, \hdots, x_n) \in H$.
    Let $\lambda \in \R_{\max}^m$ such that $X = \bigoplus\limits_{i=1}^m \lambda_i p_i$ with  $\bigoplus\limits_{i=1}^m \lambda_i = 0$. Take any $x_{n+1} \in [a,b]$.
    We have $Y = (x_1, \hdots, x_n, x_{n+1}) \in H'$ by definition.
    Note also that $\bigoplus\limits_{i=1}^m \lambda_i = 0$, so there exists $i \in [1,m]$ such that $\lambda_i = 0$.
    If $i \in I$, let $j = i$, otherwise, 
    there exists $j \in I$ such that $p_j \leq p_i$ by the previous Equation \ref{thm:I}.
    
    In both case, we have $j \in I$ such that $p_j \leq p_i \leq X$, thus $p_j \oplus X = X$.
    Let now $\lambda_{n+1} = x_{n+1} - b \leq 0$.
    We have $\bigoplus\limits_{i=1}^m \lambda_i p_i \oplus \lambda_{m+1} p_j = X \oplus \lambda_{m+1} p_j = X$.
    
    Let $p'_i = (p_i, a)$ for $1 \leq i \leq m$ and $p'_{m+1} = (p_j, b)$.
    We have $\bigoplus\limits_{i=1}^{m+1} \lambda_i p'_i = (X, a \oplus \lambda_{m+1} b) = (X, a \oplus x_{n+1}) = (X, x_{n+1}) = Y$.
    For $i \in [1, m]$: $p_i \in H$ thus $p'_i \in H'$, and $p_j \in H$ therefore $p'_{m+1} \in H'$.
    
    Therefore $P' = (p_i, a)_{1 \leq i \leq m}, (p_i, b)_{i \in I}$ generates $H'$.
    \end{proof}

\section{Proof of Theorem \ref{thm:intRmtoRn}}

\label{sec:proof:thm:intRmtoRn}

\begin{proof}
We verify that the internal description matches the external one.
To do so, we ensure that every point which is in the tropical convex hull of $(A, B_1, \hdots, B_m,$ $C_1, \hdots, C_n)$ is inside the polyhedron defined externally and conversely. 
%that every point in that polyhedron is a convex tropical linear combination of $(A, B_1, \hdots, B_m, C_1, \hdots, C_n)$.\newline

Let $H_{ext} = \{(x_1, \hdots, x_m, y_1, \hdots, y_n)\}$ the polyhedron defined externally as in Theorem \ref{thm:RmtoRn}
and let $H_{int}$ % = \{\max(\alpha + A, \hdots \beta_i + B_i, \hdots, \gamma_j + C_j, \hdots), \max(\alpha, \beta_1, \hdots, \beta_m, \gamma_1, \hdots, \gamma_n) \leq 0\}$ 
be the polyhedron defined internally as in Theorem \ref{thm:intRmtoRn}. 

We note the following properties: 
    \begin{equation}
        \forall (i, j) \in [1,n] \times [1,m], \delta_{i,j} \leq \ox_j - \ux_j \mbox{ and } \delta_{i,j} \leq |w_{i,j}|(\ox_j - \ux_j)
        \label{eq:extintequiv1}
        \end{equation}
        
        Moreover, 
    \begin{equation}
         \forall i \in [1,n], M_i - m_i = \sum\limits_{j=1}^m |w_{i,j}| (\ox_j - \ux_j) \geq \sum\limits_{j=1}^m \delta_{i,j}
        \label{eq:extintequiv6}
        \end{equation}
    
    Thus, 
    \begin{equation}
        \forall (i, j) \in [1,n] \times [1,m], M_i - m_i \geq \delta_{i,j}
        \label{eq:extintequiv2}
        \end{equation}

Finally, as $d_{i_1,i_2} = m_{i_2} + w_{i_1,0} - w_{i_2,0} + \sum\limits_{w_{i_1,j} < w_{i_2,j}} \ux_j (w_{i_1,j} - w_{i_2,j}) + \sum\limits_{w_{i_1,j} > w_{i_2,j}} \ox_j$ $(w_{i_1,j} - w_{i_2,j})$, we have, for all 
$i_1, i_2 \in [1,n]$, $i_1 \neq i_2$:
\begin{multline}
        m_{i_1} - d_{i_1,i_2} =  w_{i_1,0}+\sum\limits_{w_{i_1,j} < 0} w_{i_1,j} \ox_j+\sum\limits_{w_{i_1,j} > 0} w_{i_1,j} \ux_j - w_{i_2,0}-\sum\limits_{w_{i_2,j} < 0} w_{i_2,j} \ox_j \\
        -\sum\limits_{w_{i_2,j} > 0} w_{i_2,j} \ux_j 
        -w_{i_1,0} + w_{i_2,0} - \sum\limits_{w_{i_1,j} < w_{i_2,j}} \ux_j (w_{i_1,j} - w_{i_2,j}) \\ 
        - \sum\limits_{w_{i_1,j} > w_{i_2,j}} \ox_j (w_{i_1,j} - w_{i_2,j})
\end{multline}
\noindent Rearranging the terms and separating the sums into the four cases, $w_{i_1,j} < w_{i_2,j} < 0$, $w_{i_1,j} > w_{i_2,j} > 0$, $w_{i_1,j} > 0 \geq w_{i_2,j}$ and $w_{i_1,j} < 0 \leq w_{i_2,j}$, we get:
    \begin{multline}
    m_{i_1} - d_{i_1,i_2} =  
    \sum\limits_{w_{i_1,j} < w_{i_2,j} < 0} (\ox_j - \ux_j) (w_{i_1,j} - w_{i_2,j}) \\
    -
    \sum\limits_{w_{i_1,j} > w_{i_2,j} > 0} (\ox_j - \ux_j) (w_{i_1,j} - w_{i_2,j}) - \sum\limits_{w_{i_1,j} > 0 \geq w_{i_2,j}} (\ox_j - \ux_j) w_{i_1,j} +\\ \sum\limits_{w_{i_1,j} < 0 \leq w_{i_2,j}} (\ox_j - \ux_j) w_{i_1,j} \leq 0
    \label{eq:extintequiv3}
    \end{multline}
    A similar calculation %\todo{Sebastien check please} 
    shows that
        \begin{equation}
    M_{i_1} - d_{i_1,i_2} 
 \geq 0
    \label{eq:extintequiv3bis}
    \end{equation}
    We also have:
    \begin{multline}
    d_{i_1,i_2} - m_{i_1} \geq \sum\limits_{w_{i_1,j} > w_{i_2,j} > 0} (\ox_j - \ux_j) (w_{i_1,j} - w_{i_2,j}) + \sum\limits_{w_{i_1,j} > 0 \geq w_{i_2,j}} (\ox_j - \ux_j) w_{i_1,j} \\
    \geq w_{i_1,j}(\ox_j - \ux_j) \mbox{ \ for all $i_1$ such that $w_{i_1,j} > 0$ and any $i$} \\
    \label{eq:extintequiv7}
    \end{multline}
%    \todo{Sebastien: rajouter la conclusion par rapport aux delta etc. que l'on utilise plus tard, dans l'equation (\ref{eq:extintequiv7}).}
The last inequality is valid since all summands are positive. For the same reasons, 
\begin{multline}
d_{i_1,i_2} - m_{i_1}
    \geq (\ox_j - \ux_j) (w_{i_1,j} - w_{i_2,j}) \mbox{ $\forall {i_1, i_2}$ such that $\lambda_{i_1,j}>\lambda_{i_2,j}$ and any $j$} 
    \label{eq:extintequiv7bis}
\end{multline}
    Similarly, 
    \begin{multline}
    c_{i_1,i_2} - M_{i_2} = 
    \sum\limits_{0 < w_{i_1,j} < w_{i_2,j}} (\ox_j - \ux_j) (w_{i_1,j} - w_{i_2,j}) 
    -
    \sum\limits_{0 > w_{i_1,j} > w_{i_2,j}} (\ox_j - \ux_j) (w_{i_1,j} \\- w_{i_2,j}) -
    \sum\limits_{w_{i_1,j} \leq 0 < w_{i_2,j}} (\ox_j - \ux_j) w_{i_2,j}
    +
    \sum\limits_{w_{i_1,j} \geq 0 > w_{i_2,j}} (\ox_j - \ux_j) w_{i_2,j} \leq 0
    \label{eq:extintequiv4}
    \end{multline}
    \noindent And, by a similar calculation, %\todo{Sebastien check please}, 
    \begin{equation}
    c_{i_1,i_2} - m_{i_2} \geq \delta_{i_2,j}
\label{eq:extintequiv4bis}
\end{equation}

We first prove $H_{int} \subset H_{ext}$, by proving that every extreme point of $H_{int}$ is in $H_{ext}$, i.e. that every extreme point defined in Theorem \ref{thm:intRmtoRn} matches all constraints defined in Theorem \ref{thm:RmtoRn}. 

%\todo{EG 15/01/21, will carry on later}
Consider first generator $A = (x_1,\ldots,x_{m},y_1,\ldots,y_n)=(a_1, \hdots, a_m, m_1,  \hdots,$ $m_n)$ and take any $(i, j) \in [1,n] \times [1,m]$.  First, obviously, $x_j - b_j = \ux_j - \ox_j \leq 0$, so $A$ satisfies the first inequality of Equation (\ref{eq:RmtoRn1}) of Theorem \ref{thm:RmtoRn}. Also, $y_i - M_i = m_i - M_i \leq 0$, which is the second part of Equation (\ref{eq:RmtoRn1}). 
    Similarly, 
     $x_j - a_j = 0 \geq 0$ and $y_i - M_i = m_i - M_i \leq -\delta_{i,j}$ by Equation (\ref{eq:extintequiv2}), which is the first part of Equation (\ref{eq:RmtoRn2}). 
    Finally, 
     $y_i - m_i = 0 \geq 0$ and 
     $x_j - \ox_j = \ux_j - \ox_j \leq -\delta_{i,j}$ by Equation (\ref{eq:extintequiv1}), and 
    for all $i_1, i_2 \in [1,n]$:
$y_{i_1} - d_{i_1,i_2} = m_{i_1} - d_{i_1,i_2} \leq 0$ by Equation (\ref{eq:extintequiv3}), and as $0 = y_{i_2} - m_{i_2}$, we conclude that $y_{i_1} - d_{i_1,i_2}\leq y_{i_2} - m_{i_2}$, which is Equation (\ref{eq:RmtoRn3}) of Theorem \ref{thm:RmtoRn}. 
    Therefore, $A \in H_{ext}$.

Consider now $B_j = (x_1,\ldots,x_{m},y_1,\ldots,y_n)=(\ux_j, \hdots, \ux_{j-1}, \ox_j, \ux_{j+1}, \hdots, \ux_m,$  $m_1 + \delta_{1,j}, \hdots, m_n + \delta_{n,j})$ for some $j \in [1,m]$, and take any 
    $(i, j') \in ([1,n] \times [1,m] \setminus \{j\})$
We have easily 
        $x_j - \ox_j = 0 \leq 0$, 
        $x_{j'} - \ox_{j'} = \ux_{j'} - \ox_{j'} \leq 0$ and 
        $y_i - M_i = m_i - M_i + \delta_{i,j} \leq 0$ %\todo{Explain} 
        which is Equation (\ref{eq:RmtoRn1}) of Theorem \ref{thm:RmtoRn}. Also, 
        $x_j - \ux_j = \ox_j - \ux_j \geq 0$, 
        $x_{j'} - \ux_{j'} = 0 \geq 0$, 
        $y_i - M_i + \delta_{i,j} = m_i - M_i + 2\delta_{i,j} \leq \delta_{i,j} \leq \ox_j - \ux_j = x_j - \ux_j$ by Equations (\ref{eq:extintequiv1}) and (\ref{eq:extintequiv2}), %\todo{EG: complete 16/01/21}
        and 
        $y_i - M_i + \delta_{i,j'} = m_i - M_i + \delta_{i,j} + \delta_{i,j'} \leq 0$ by Equation (\ref{eq:extintequiv6}). %, and then $y_j-M_j+\delta_{i',j} \leq 0 = x_{i'} - a_{i'}$.
        More precisely, this last inequality is obtained as follows: 
        Equation (\ref{eq:extintequiv6}), $M_i - m_i \geq \sum\limits_{j=1}^m \delta_{i,j}$ implies, since all the $\delta_{i,j}$ are positive, that $M_i-m_i\geq \delta_{i,j}+\delta_{i,j'}$, therefore, 
        $m_i - M_i \leq -\delta_{i,j}-\delta_{i,j'}$, i.e. $m_i - M_i +\delta_{i,j}+\delta_{i,j'}\leq 0$.  
%        \textbf{Check, there is a problem here: $\delta_{i',j}\geq 0$ except if $\lambda_{i',j}\leq 0$? equation 8 ou on particularise a somme delta i j avec delta i' j}. 

Finally,         
        $y_i - m_i = \delta_{i,j} \geq 0$, 
        $x_j - \ox_j + \delta_{i,j} = \delta_{i,j} = y_i - m_i$, thus, trivially, $x_j - \ox_j + \delta_{i,j} \leq y_i - m_i$, 
        %\textbf{Check: $\delta_{i,j}\leq M_j-m_j$ but $y_j-m_j \leq M_j-m_j$, not the contrary? OK  car tout est egal}, 
        $x_{j'} - \ox_{j'} + \delta_{i,j'} = \ux_{j'} - \ox_{j'} + \delta_{i,j'} \leq 0$ by Equation (\ref{eq:extintequiv1}), thus, $x_{j'} - \ox_{j'} + \delta_{i,j'} \leq y_i - m_i$ which are Equations (\ref{eq:RmtoRn2}). 

Now, consider any 
    $i_1, i_2 \in [1,n]$. 
        Then $y_{i_1} - d_{i_1,i_2} = \delta_{i_1,j} + m_{i_1} - d_{i_1,i_2}$, therefore $y_{i_1} - d_{i_1,i_2} \leq \delta_{i_1,j}$ by Equation (\ref{eq:extintequiv3}). %\textbf{car y j1 geq min j1}. 
        If $\delta_{i_1,j} \leq \delta_{i_2,j}$, then 
            $y_{i_1} - d_{i_1,i_2} \leq \delta_{i_1,j} \leq \delta_{i_2,j} = y_{i_2} - m_{i_2}$.
            %\textbf{car y j1 geq min j1}
            Otherwise, $\delta_{i_1,j} > \delta_{i_2,j} \geq 0$ implies
            $w_{i_1,j} > 0$ by definition of $\delta_{i_1,j}$. By Equation (\ref{eq:extintequiv7}), since $w_{i_1,j} > 0$, $d_{i_1,i_2} - m_{i_1} \geq w_{i_1,j}(\ox_j - \ux_j)$, which is greater or equal than $\delta_{i_1,j}$ by Equation (\ref{eq:extintequiv1}). If we suppose now $w_{i_2,j} \leq 0$, then, by definition, $\delta_{i_2,j}=0$, and this is in turn greater or equal than $\delta_{i_1,j} = \delta_{i_1,j} - \delta_{i_2,j}$. 
%            \textbf{By equation  7}. 
            Otherwise, if $w_{i_2,j} \geq 1$, then $\delta_{i_1,j} = \delta_{i_2,j} = \ox_j - \ux_j$ and $d_{i_1,i_2} - m_{i_1} \geq \delta_{i_1,j} - \delta_{i_2,j} = 0$. Finally, if $0 < w_{i_2,j} < 1$, then $\delta_{i_2,j} = w_{i_2,j}(\ox_j - \ux_j)$ and, by Equation (\ref{eq:extintequiv7bis}),  $d_{i_1,i_2} - m_{i_1} \geq (\ox_j - \ux_j) (w_{i_1,j} - w_{i_2,j}) = w_{i_1,j}(\ox_j - \ux_j) - \delta_{i_2,j} \geq \delta_{i_1,j} - \delta_{i_2,j}$. Therefore, in every case, $\delta_{i_1,j} - \delta_{i_2,j} \leq d_{i_1,i_2} - m_{i_1}$. %\todo{Explain}. 
            Thus, $y_{i_1} - d_{i_1,i_2} \leq \delta_{i_2,j}=y_{i_2} - m_{i_2}$.
        We conclude that $B_j \in H_{ext}$.

    Consider now $C_i = (x_1,\ldots,x_{m},y_1,\ldots,y_n)=(\ux_1 + \delta_{i,1}, \hdots, \ux_m + \delta_{i,m}, c_{i,1}, \hdots, $ $c_{i,i-1}, M_i, c_{i,i+1}, \hdots, c_{i,n})$ for some $i \in [1,n]$, and take any 
    $(i', j) \in ([1,n]\setminus \{i\}) \times [1,m] $. 

 We have $x_j - \ox_j = \ux_j - \ox_j + \delta_{i,j} \leq 0$ by Equation (\ref{eq:extintequiv1}), $y_i - M_i = 0 \leq 0$ and $y_{i'} - M_{i'} = c_{i,i'} - M_{i'} \leq 0$ by Equation (\ref{eq:extintequiv4}), which shows that $C_i$ satisfies Equation (\ref{eq:RmtoRn1}). 
        Then, $x_j - \ux_j = \delta_{i,j} \geq 0$, by definition of $\delta_{i,j}$, $y_i - M_i + \delta_{i,j} = \delta_{i,j} = x_j - \ux_i$ by definition of $x_j$, $y_{i'} - M_{i'} + \delta_{i',j} = c_{i,i'} - M_{i'} + \delta_{i',j} \leq \delta_{i',j}$ by Equation (\ref{eq:extintequiv4}), which is equal to $x_j - \ux_j$ by definition of $x_1$. These inequalities are exactly Equation (\ref{eq:RmtoRn2}) of Theorem \ref{thm:RmtoRn}. 
        
        Finally, $y_i - m_i = M_i - m_i \geq 0$, $y_{i'} - m_{i'} = c_{i,i'} - m_{i'} \geq 0$ since $c_{i,i'} = M_{i} - d_{i,i'}
+ m_{i'}$ by definition of $c_{i,i'}$, and by Equation (\ref{eq:extintequiv3bis}). 
%\textbf{We need to prove $M_j-d_{j,j'}\geq 0$ similarly to Equation (\ref{eq:extintequiv3}) right?}. 
Also, 
        $x_j - \ox_j + \delta_{i,j} = \ux_j - \ox_j + 2 \delta_{i,j} \leq \delta_{i,j} \leq M_i - m_i$ by Equation (\ref{eq:extintequiv1}) and then by Equation (\ref{eq:extintequiv2}), $x_j - \ox_j + \delta_{i',j} = \ux_j - \ox_j + \delta_{i,j} + \delta_{i',j} \leq \delta_{i',j} \leq c_{i,i'} - m_{i'}$ by Equation (\ref{eq:extintequiv4bis}). 
%        \textbf{Why? Should be proven I guess at the same time as Equation (\ref{eq:extintequiv4})}, 
We also have 
        $y_{i'} - d_{i',i} = c_{i,i'} - d_{i',i} = M_i - m_i = y_i - m_i$ by definition of $c_{i,i'}$ and of $y_i$, and $y_i - d_{i,i'} = M_i - d_{i,i'} = c_{i,i'} - m_{i'} = y_{i'} - m_{i'}$. Finally, for all $(i_1,i_2) \in [1,n]$:
        $y_{i_1} - d_{i_1,i_2} = c_{i,i_1} - d_{i_1,i_2} = M_i + m_{i_1} - d_{i,i_1} - d_{i_1,i_2} \leq M_i - d_{i,i_2} = c_{i,i_2} - m_{i_2}$ %\todo{By equation...avec les lambdas a reecrire juste apres l'equation 9}
        which ends the proof that $C_j$ enjoys Equation (\ref{eq:RmtoRn3}). 
        
    Therefore, $C_j \in H_{ext}$ and $H_{int} \subset H_{ext}$.

We then prove $H_{ext} \subset H_{int}$\footnote{Equivalently, we could have determined the external representation we have is the one deduced from the extremal points, by computing the polar cone, i.e. the dual of the tropical polyhedron defined by its extreme points, and take the extreme points of this dual: this gives the external representation of the tropical polyhedron, see e.g. \cite{SAS2008}.} by proving that every point in $H_{ext}$ is a tropical convex linear combination of the extreme points of $H_{int}$.

Let $P = (x_1,\dots,x_m,y_1,\dots,y_n) \in H_{ext}$ and 
\begin{eqnarray}
    P' & = & (x'_1,\dots,x'_m,y'_1,\dots,y'_n) \\
    & = & \max\left(A, \max\limits_{j=1}^m (B_j + (x_j - \ox_j)),\max\limits_{i=1}^n (C_i + (y_i - M_i))\right)
        \label{eq:Pprime}
    \end{eqnarray}
$P'$ is given as a tropical convex linear combination of generators $A$, $B_j$ and $C_i$ and thus is in $H_{int}$
since 
$$\max\left(0, \ \max\limits_{j=1}^{m} (x_j-\ox_j), \ \max\limits_{i=1}^n (y_i-M_i)\right)=0$$ 
We show that $P=P'$ hence $H_{ext} \subset H_{int}$

For any $j \in [1,m]$, 
$$x'_j = \max\left(\ux_j, \ox_j + x_j - \ox_j, \max\limits_{j' \in [1,m], j \neq j'}(\ux_j + x_{j'} - \ox_{j'}), \max\limits_{i=1}^n(\ux_j + \delta_{i,j} + y_i - M_i)\right)$$
But, for all $j' \in [1,m]$ and $j \neq j'$, $x_{j'} \leq \ox_{j'}$ thus $\max\limits_{j' \in [1,m], j \neq j'}(\ux_j + x_{j'} - \ox_{j'}) \leq \ux_j \leq x_j$.
Similarly, for all $i \in [1,n]$: $y_i - M_i + \delta_{i,j} \leq x_j - \ux_j$ by Equation (\ref{eq:RmtoRn2}), thus $\max\limits_{i=1}^n(\ux_j + \delta_{i,j} + y_i - M_i)) \leq x_j$.
Thus $x'_j = x_j$.

Now for any $i \in [1,n]$,  
$$y'_i = \max\left(m_i, y_i, \max\limits_{j=1}^m (m_i + \delta_{i,j} +  x_j - \ox_j),  \max\limits_{i' \in [1,n], i \neq i'}(c_{i',i} + y_{i'} - M_{i'})\right)$$
We know by Equation (\ref{eq:RmtoRn3}) that for all $j \in [1,m]$: $x_j - \ox_j + \delta_{i,j} \leq y_i - m_i$ thus $\max\limits_{j=1}^m (m_i + \delta_{i,j} +  x_j - \ox_j) \leq y_i$.
Similarly, for all $i' \in [1,n]$, $i \neq i'$: $y_{i'} + c_{i',i} - M_i=y_{i'} - d_{i',i} + m_i$ by definition of $c_{i,i'}$, but by Equation (\ref{eq:RmtoRn3}), this is less or equal than $y_i$ thus $\max\limits_{i' \in [1,n], i \neq i'}(c_{i',i} + y_{i'} - M_{i'})) \leq y_i$.
Therefore $y'_i = y_i$ and $P=P'$. 

%Therefore $P = P' = (0 + A) + \max_{i=1}^m (\beta_i + B_i) + \max_{j=1}^n (\gamma_j + C_j)$ 
%and $max(0, \beta_1, \dots, \beta_m, \gamma_1, \dots, \gamma_n) = 0$.\newline
We conclude that $P \in H_{int}$ and $(A, B_1, \dots, B_m, C_1, \dots, C_n)$ generates $H_{ext}$.
Therefore $H_{ext} \subset H_{int}$, {and thus $H_{ext} = H_{int}$.}

{Finally, we prove that every generator $A$, $B_j$ and $C_i$ is an extreme generator of the polyhedron}. This ensures minimal presentation for $H_{int}$. %(We could also try to prove that every point is distinct from all the others but that would require additional conditions on the parameters, such as $a_i \neq b_i$ and $\lambda_{i,j} \neq 0$, which go beyond the scope of this report)\newline

We know from \cite{vertices} that a point $g$ is extreme in a tropical polyhedron $C \subset \R_{max}^d$ if there exists $1 \leq t \leq d$ such that $g$ is a minimal element of the set $\{ x \in C, x_t = g_t \}$. In that case, $g$ is said to be an extreme of type $t$.
Consider now any $P = (x_1,\dots,x_m,y_1,\dots,y_n) \in H$.

We see first that for all $(i, j) \in [1,n] \times [1,m]$, $\ux_j \leq x_j$ and $m_i \leq y_i$, meaning precisely that $A$ is an extreme generator of $H$.

%%% SYLVIE J4ENSUISI ICI ! 
    Fix now $j \in [1,m]$. Take $P$ as above, such that 
    %Let $P = (x_1,\dots,x_m,y_1,\dots,y_n) \in H$ such as 
    $x_j = \ox_j$. Then 
    for all $j' \in [1,m]$, $j \neq j'$: $\ux_{j'} \leq x_{j'}$. We also have  
    for all $i \in [1,n]$: $m_i + \delta_{i,j} \leq y_i$ since by Equation (\ref{eq:RmtoRn3}), $y_i-m_i\geq xj-\ox_j+\delta_{i,j}=\delta_{i,j}$ because we suppose $x_j=\ox_j$.  %(inequality on $x_i - a_i$). 
    This means that $B_j$ is an extreme generator of type $j$ of $H$. % (see Proposition 1 of \cite{vertices}).
%    \todo{Ca vaudrait le coup meme si j'ai rajoute la ref, de donner la def. dans le rapport.}
    
    Finally, fix $i \in [1,n]$ and take $P$ as above such that 
    %Let $P = (x_1,\dots,x_m,y_1,\dots,y_n) \in H$ such as 
    $y_i = M_i$. Then, 
    for all $j \in [1,m]$, by Equation (\ref{eq:RmtoRn2}), $x_j-\ux_j \geq y_i-M_i+\delta_{i,j}$, but as we supposed $y_i=M_i$, this implies $\ux_j + \delta_{i,j} \leq x_j$. %since (inequality on $y_j - m_j$) 
    We also have that 
    for all $i' \in [1,n]$, $i \neq i'$, by Equation (\ref{eq:RmtoRn3}), $y_{i'}-m_{i'}\geq y_i-d_{i,i'}$
    so $y_{i'}\geq y_i-d_{i,i'}+m_{i'}$, but as we supposed that $y_i=M_i$, we have 
    $c_{i,i'} = M_i - d_{i,i'} + m_{i'} \leq y_{i'}$. %(inequality on $y_{j'} - m_{j'})$. 
    This shows that 
    $C_i$ is an extreme of type $i + m$ of $H$.

Therefore, all points in $(A, B_1, \dots, B_m, C_1, \dots, C_n)$ are extreme points of $H_{int}$.
\end{proof}

%\bibliography{...}

%\end{document}

\section{Proof of Theorems \ref{thm:subd} and \ref{thm:subdgen}}
\label{sec:proofsubd}

\begin{theoremn} (Theorem \ref{thm:subd})
A sound abstraction as a tropical polyhedron $\mathcal{P}$ of the graph of $f$ over $[a, b]$ given by subdividing the domain in $N$ sub-intervals has the following external representation of $N+2$ tropical inequalities: 
%\todo{EG: je suis en train de travailler ca 21/12/2020}:

%The polyhedron can also be given by its external description, in which case each subdivision adds a new inequality, for a total of $N + 2$ inequalities
$\mathcal{P}$ is defined by the two constraints $x \leq b$, $a \leq x$ and the following $N$ constraints, for all $k$ from $0$ to $N-1$, depending on the value of $\lambda$:
\begin{itemize}
    \item If $\lambda \leq 0$, $0 \leq \max(x - c_k, y - f(c_k))$.
    \item If $0 \leq \lambda \leq 1$, $y - f(c_k) \leq \max(0, x - c_k)$.
    \item If $\lambda \geq 1$, $x - c_k \leq \max(0, y - f(c_k))$.
\end{itemize}
$\mathcal{P}$ can also be internally represented as the tropical convex hull of at most $N + 2$ extreme points $A$, $B$ and $C_i$, $i \in [1, N]$ with 
%\begin{center}
%\begin{tabular}{|l|c|c|c|}
%\hline
%\begin{itemize}
%\item 
$A=(a,f(a))$, 
$B=(b,f(b))$, 
%& $A$ & $B$ & $C_i$, $i \in [1,N]$ \\
%\item 
and $C$ is $(c_{i-1}, f(c_i))$ if $\lambda \leq 0$,
$(c_{i-1} + f(c_i) - f(c_{i-1}),  f(c_i))$ if $0 \leq \lambda \leq 1$ 
and $(c_i, f(c_{i-1}) + c_i - c_{i-1})$ if $\lambda \geq 1$.
%\end{itemize}
\end{theoremn}

\begin{theoremn} (Theorem \ref{thm:subdgen})
A sound and tighter over-approximation of $\mathcal{G}_f$ than the one of Theorem \ref{thm:RmtoRn} is given externally by the tropical constraints of Theorem \ref{thm:RmtoRn} plus the following constraints, for any subdivision $c^0_i=a_i,\ldots c^N_i=b_i$ of intervals $[a_i, b_i]$, $i=1,\ldots,m$ in $N$ subintervals:
\begin{itemize}
\item 
If $\lambda_{i,j} \leq 0$, then we add the constraint $0 \leq \max(x_i -c^k_i, y_j - m_j + \lambda_{i,j} (b_i - c^k_i))$. Otherwise     
if $0 \leq \lambda_{i,j} \leq 1$, we add the constraint $y_j - M_j + \lambda_{i,j} (b_i - c^k_i) \leq \max(0, x_i - c^k_i)$.
And finally, we add $x_i - c^k_i \leq \max(0, y_j - m_j - \lambda_{i,j} (c^k_i - a_i))$
if $\lambda_{i,j} \geq 1$. 
\item 
  Let $I_{j-} \subset [1,m]$, maximal, such as for all $i \in I_{j-}$, $\lambda_{i,j} \leq 0$.
    We write $\sigma_{j-} = \sum\limits_{i \in I_{j-}} \lambda_{i,j}(b_i - c^k_i)$. Then, we add the constraint $0 \leq \max(y_j - m_j + \sigma_{j-}, \max\limits_{i \in I_{j-}}(x_i - c^k_i))$.
    \item 
    Let $I_{j0} \subset [1,m]$, maximal, such as for all $i \in I_{j0}$, $0 \leq \lambda_{i,j} \leq 1$ and $\sum\limits_{i \in I_{j0}} \lambda_{i,j} \leq 1$ and let $\sigma_{j0} = \sum\limits_{i \in I_{j0}} \lambda_{i,j}(b_i - c^k_i)$. Then we add the constraint 
    $y_j - M_j + \sigma_{j0} \leq \max(0, \max\limits_{i \in I_{j0}}(x_i - c^k_i))$.
\item Finally, 
for any $J$ subset of $[1,n]$, let $\sigma_{i,J} = \sum\limits_{j \in J} \lambda_{i,j}$ and 
$m_J=\sigma_{0,J} + \sum\limits_{\sigma_{i,J} < 0} \sigma_{i,J} b_i + \sum\limits_{\sigma_{i,J} > 0} \sigma_{i,J} a_i$.
For $j \in J$, let $u_j \in [m_j, M_j]$ such as $\sum\limits_{j \in J} u_j = m_J$: 
we add the constraint 
$0 \leq \max_{j \in J}(y_j - u_j)$. 
%todo{How many constraints added then, quite a lot?}.
\end{itemize}
\end{theoremn}

\begin{proof}
Let $(i,j) \in [1,m] \times [1,n]$ and $c^k_i \in [a_i, b_i]$ be any of the $c^k_i$, $k=0,\ldots, N-1$.
We have: %\todo{To be checked}: 
\begin{itemize}
    \item Suppose $\lambda_{i,j} \leq 0$. Then 
        if $x_i \leq c^k_i$, $y_j \geq m_j - \lambda_{i,j} (b_i - c^k_i)$, otherwise $x_i - c \geq 0$.
   This can be summarized tropically as $0 \leq \max(x_i -c^k_i, y_j - m_j + \lambda_{i,j} (b_i - c^k_i)$.
    \item If $0 \leq \lambda_{i,j} \leq 1$, then suppose first that 
        $x_i \leq c^k_i$. Then $y_j \leq M_j - \lambda_{i,j} (b_i - c^k_i)$. Otherwise 
        $y_j - M_j \leq x_i - c^k_i - \lambda_{i,j} (b_i - c^k_i)$.
    Overall: $y_j - M_j + \lambda_{i,j} (b_i - c^k_i) \leq \max(0, x_i - c^k_i)$.
    \item Finally, if $\lambda_{i,j} \geq 1$, then suppose first that 
        $x_i \geq c^k_i$. Then $y_j - m_j \geq x_i - c^k_i + \lambda_{i,j} (c^k_i - a_i)$.
        Otherwise, $x_i - c^k_i \leq 0$.
    To summarize, in this case: $x_i - c^k_i \leq \max(0, y_j - m_j - \lambda_{i,j} (c^k_i - a_i))$
\end{itemize}
Now, there are also extra relations between the 
%We can also find relations between multiple 
$x_i$s and any of the $y_j$s.

Consider any $c^k=(c^k_1,\ldots, c^k_m \in [a_1,b_1] \times \hdots \times [a_m,b_m]$ and $j \in [1,n]$.

\begin{itemize}
    \item Let $I_{j-} \subset [1,m]$, maximal, such as for all $i \in I_{j-}$, $\lambda_{i,j} \leq 0$.
    We write $\sigma_{j-} = \sum\limits_{i \in I_{j-}} \lambda_{i,j}(b_i - c^k_i)$.Then, 
        if for all $i \in I_{j-}$, $x_i \leq c^k_i$, then $y_j \geq m_j - \sigma_{j-}$. Otherwise, 
        $\max_{i \in I_{j-}}(x_i - c^k_i) \geq 0$. Overall: 
    $0 \leq \max(y_j - m_j + \sigma_{j-}, \max_{i \in I_{j-}}(x_i - c^k_i))$.
    \item Let $I_{j0} \subset [1,m]$, maximal, such as for all $i \in I_{j0}$, $0 \leq \lambda_{i,j} \leq 1$ and $\sum\limits_{i \in I_{j0}} \lambda_{i,j} \leq 1$. 
    Let $\sigma_{j0} = \sum\limits_{i \in I_{j0}} \lambda_{i,j}(b_i - c^k_i)$.
          Now, if for all $i \in I_{j0}$, $x_i \leq c^k_i$, then $y_j \leq M_j - \sigma_{j0}$.
       Otherwise, let $i_{max} \in I_{j0}$ such that $x_{i_{max}} - c^k_{i_{max}}$ is maximized.
        In this case we have $y_j - M_j + \sigma_{j0} \leq \sum\limits_{i \in I_{j0}} \lambda_{i,j} (x_i - c^k_i) \leq \sum\limits_{i \in I_{j0}} \lambda_{i,j} (x_{i_{max}} - c^k_{i_{max}}) \leq x_{i_{max}} - c^k_{i_{max}}$. Overall, we have the affine tropical constraint: 
   $y_j - M_j + \sigma_{j0} \leq \max(0, \max_{i \in I_{j0}}(x_i - c^k_i))$. %\todo{Notation $j0$ to be replaced by $j_0$}.
\end{itemize}

There are also relations between the $y_j$s.
Consider any subset $J$ of $[1,n]$ and $\sigma_{i,J} = \sum\limits_{j \in J} \lambda_{i,j}$.
Then, 
$\sum\limits_{j \in J} y_j = \sum\limits_{j \in J} \lambda_{0,j} + \sum\limits_{i = 1}^m \lambda_{i,j} x_i \geq \sigma_{0,J} + \sum\limits_{\sigma_{i,J} < 0} \sigma_{i,J} b_i + \sum\limits_{\sigma_{i,J} > 0} \sigma_{i,J} a_i = m_J$.
For $j \in J$, let $u_j \in [m_j, M_j]$ such as $\sum\limits_{j \in J} u_j = m_J$. %\todo{Not very clear}.

Suppose that for all $j \in J$, $y_j \leq c^k_j$. Then $\sum\limits_{j \in J} y_j \leq \sum\limits_{j \in J} u_j = m_J$, which is absurd.
Therefore $0 \leq \max_{j \in J}(y_j - u_j)$.
\end{proof}

%\newpage 
%\listoftodos{}

%\end{document}
\end{document}